\documentclass{article}

\usepackage[preprint]{neurips_2025}

\usepackage[utf8]{inputenc} 
\usepackage[T1]{fontenc}    
\usepackage[pagebackref=true,breaklinks=true,colorlinks,bookmarks=false,citecolor=blue,linkcolor=blue]{hyperref}
\usepackage{url}            
\usepackage{booktabs}       
\usepackage{amsfonts}       
\usepackage{nicefrac}       
\usepackage{microtype}      
\usepackage{xcolor}         
\usepackage{multicol}
\usepackage{multirow}
\usepackage{ulem}

\usepackage{newfloat}
\usepackage{listings}
\usepackage{tcolorbox}
\usepackage{algorithm}
\usepackage{algorithmic}
\usepackage{colortbl}
\usepackage{lscape}
\usepackage{graphicx}
\usepackage{pifont}
\usepackage{subcaption} 

\usepackage{amsmath}
\usepackage{amssymb}
\usepackage{mathtools}
\usepackage{amsthm}
\usepackage{xcolor} 
\usepackage{wrapfig}
\usepackage{floatflt}

\title{Unpaired Deblurring via Decoupled Diffusion Model}

\author{
\vspace{2mm}
Junhao Cheng\textsuperscript{1}, 
	  ~Wei-Ting Chen\textsuperscript{2},
        ~Xi Lu\textsuperscript{1},
        ~Ming-Hsuan Yang\textsuperscript{3} \\
\vspace{2mm}
	$^1$Sun Yat-sen University \qquad $^2$ Microsoft \qquad $^3$ University of California, Merced \\
    \url{https://github.com/donahowe/UID-Diff}
}

\begin{document}

\maketitle

\begin{abstract}
Generative diffusion models trained on large-scale datasets have achieved remarkable progress in image synthesis. In favor of their ability to supplement missing details and generate aesthetically pleasing contents, recent works have applied them to image deblurring via training an adapter on blurry-sharp image pairs to provide structural conditions for restoration. However, acquiring substantial amounts of realistic paired data is challenging and costly in real-world scenarios. On the other hand, relying solely on synthetic data often results in overfitting, leading to unsatisfactory performance when confronted with unseen blur patterns. To tackle this issue, we propose UID-Diff, a generative-diffusion-based model designed to enhance deblurring performance on unknown domains by decoupling structural features and blur patterns through joint training on three specially designed tasks. We employ two Q-Formers as structural features and blur patterns extractors separately. The features extracted by them will be used for the supervised deblurring task on synthetic data and the unsupervised blur-transfer task by leveraging unpaired blurred images from the target domain simultaneously. We further introduce a reconstruction task to make the structural features and blur patterns complementary. This blur-decoupled learning process enhances the generalization capabilities of UID-Diff when encountering unknown blur patterns. Experiments on real-world datasets demonstrate that UID-Diff outperforms existing state-of-the-art methods in blur removal and structural preservation in various challenging scenarios. 
\end{abstract}

\section{Introduction}
\label{Introduction}

Dynamic blur occurs when the camera and subject move relative to each other during the exposure time, resulting in a smeared and blurred image. Deblurring, the process of removing the blur pattern while preserving the underlying structure of degraded images, is essential for restoring high-quality images for human perception and low-level computer vision applications.

\begin{figure}[!t]
  \centering
  \setlength{\tabcolsep}{1pt} 
  \begin{tabular}{c}
    \includegraphics[width=.7\textwidth]{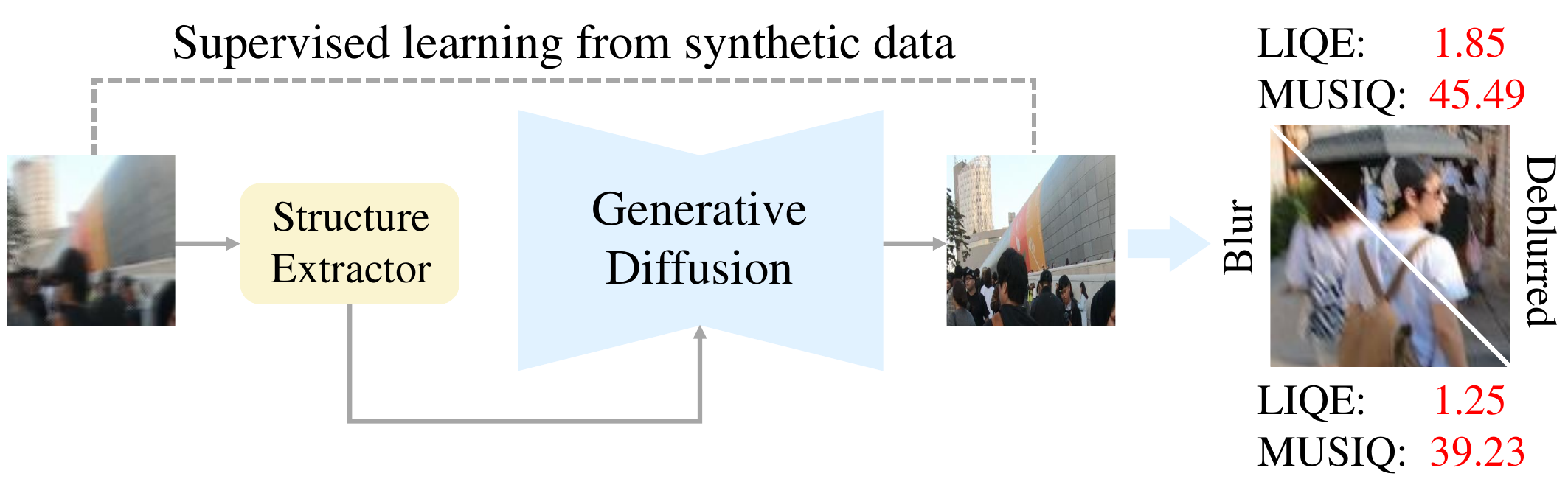} \\
    \multicolumn{1}{c}{\vspace{-14pt}} \\
    \small (a) Existing generative-diffusion-based Method \\
    \multicolumn{1}{c}{\vspace{-8pt}} \\
    \includegraphics[width=.7\textwidth]{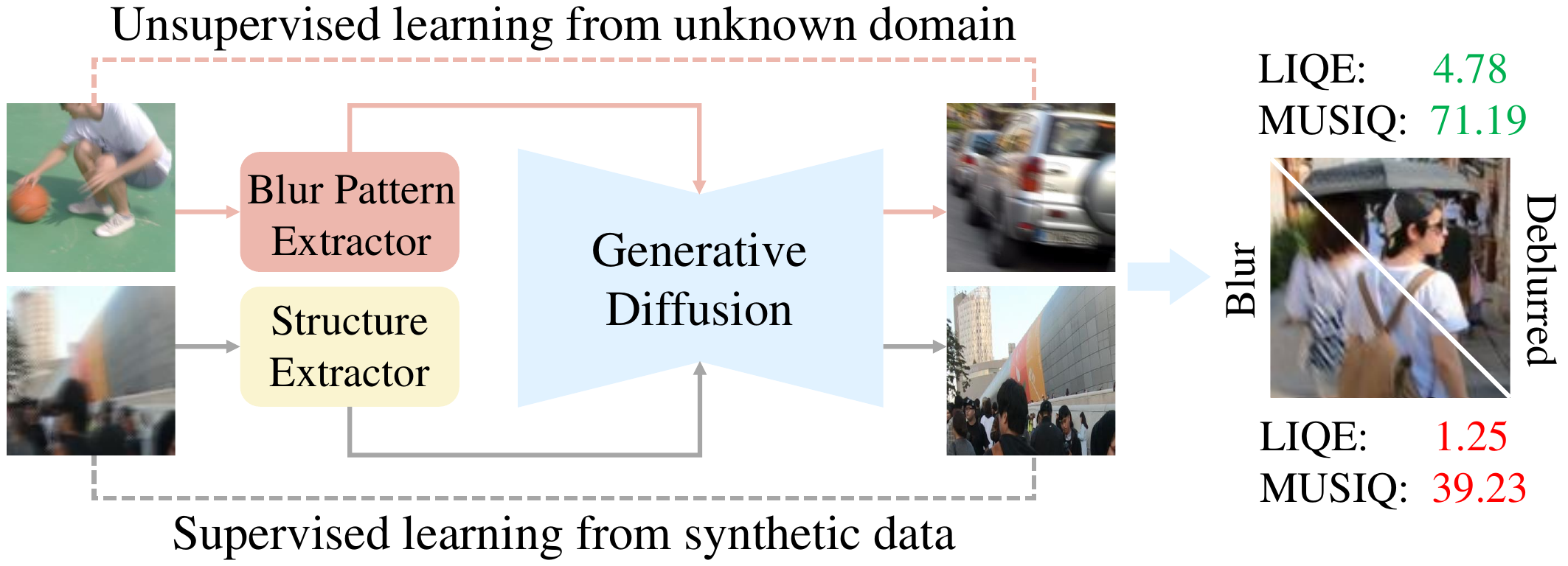} \\
    \multicolumn{1}{c}{\vspace{-12pt}} \\
    \small (b) UID-Diff \\
  \end{tabular}
  \caption{Comparison of UID-Diff with existing generative-diffusion-based deblurring methods. UID-Diff extracts blur patterns from unpaired data to facilitate the supervised structure learning process, thus generalizes well to handle blur images from unknown domains that lack paired data.}
   \label{fig:introduction}
\end{figure}

With the rapid advancement of photographic technology, a wide range of imaging devices are now employed to capture images in real-world scenarios. Due to their diverse lenses and structural designs, these devices may produce distinct blur patterns~\cite{zhang2023neural,pham2024blur2blur,zhang2020deblurring}. This diversity makes it challenging to develop an all-in-one method for deblurring images from arbitrary and varied sources. Consequently, focusing on deblurring algorithms tailored to specific domains has become increasingly significant.

As deep learning has advanced in recent years, existing deblurring models predominantly build on data-driven approaches that employ neural networks trained via supervised learning on synthetic paired data. Existing works have made efforts to develop deblurring models upon CNN~\cite{tao2018scale,nah2017deep}, Transformer~\cite{potlapalli2024promptir,chee2018airnet}, and GAN~\cite{kupyn2018deblurgan,zhang2019gan}. Recently, a new wave of research~\cite{zhang2024diff,lin2025diffbir,liu2024diff} has begun to investigate the integration of pre-trained generative diffusion models~\cite{ho2020denoising}, such as Stable Diffusion (SD)~\cite{rombach2022high}, with an adapter designed to provide structural guidance for deblurring. These approaches aim to harness the generative capabilities of diffusion models to supplement missing details and generate aesthetically pleasing outputs. However, since paired blurry-sharp training data is limited in practical scenarios, these supervised methods often encounter overfitting issues~\cite{pham2024blur2blur}, particularly when dealing with new blur patterns in specific scenarios not captured in the training datasets.

When relying solely on synthetic data is unsuitable, a promising alternative is to develop an unsupervised method that utilizes the unpaired data from a specific domain to perform deblurring directly or serve as an auxiliary for the supervised models. Existing GAN based methods attempt to reproduce the missing fine details~\cite{yi2017dualgan,zhao2022fcl} or estimate prior knowledge from unpaired data~\cite{zhang2023neural,jiangxin2021learning,ren2020neural,jiang2023uncertainty}. They are prone to overfitting to a single blur template. Furthermore, they require specific adversarial training~\cite{goodfellow2014generative}, which limits their application to diffusion-based models. Other methods~\cite{pham2024blur2blur,wu2024id} try to transfer unseen blur to a certain blur pattern. Unfortunately, these approaches entail additional computational costs and are limited by their inability to conduct large-scale training with a wide variety of synthetic blurs. Consequently, only a few unsupervised approaches for generative-diffusion-based methods have been developed to handle blur images from unknown domains.

In this work, we propose UID-Diff, a generative-diffusion-based model for unpaired image deblurring via blur-decoupled learning. As illustrated in Figure~\ref{fig:introduction}, the core concept of UID-Diff lies in its ability to decouple structural features and blur patterns via joint training on supervised structural learning tasks and unsupervised blur pattern learning tasks. 
Specifically, we employ two Q-Formers~\cite{li2023blip} to extract the structure and blur pattern separately. 
The structure extractor is trained on synthetic data to capture structural features from blurry images as conditioning for SD to perform deblurring. Meanwhile, the blur pattern extractor is trained to identify the blur representation of a specific domain by performing unsupervised blur-transfer tasks. 
In addition, we use a reconstruction task to ensure that the extracted structural features and blur patterns are complementary. As such, UID-Diff generalizes well to handle images from blur domains lacking paired data.

We make the following contributions in this work:

\begin{itemize}
    \item We present UID-Diff, a generative-diffusion-based model for image deblurring on unknown domains. To the best of our knowledge, this is the first work that integrates the generative diffusion model into unpaired deblurring tasks.
    \item We introduce an unsupervised learning strategy in the form of the blur-transfer task, which aims to extract blur patterns by leveraging unpaired data from a specific domain, thereby facilitating structure decoupling.
    \item We propose a joint training strategy to better disentangle the blur pattern and structure features. We prove its effectiveness via experiments on real-world datasets.
\end{itemize}

\section{Related Work}
\label{Related Works}

{\bf Diffusion Models for Image Restoration.}
Recent years have witnessed the success of diffusion models in image synthesis~\cite{sohl2015deep,song2019generative,Song_Meng_Ermon_2020,ho2020denoising,rombach2022high,ye2023ip,cheng2024autostudio,luo2025object}. These methods are pre-trained from large-scale text-image pair data~\cite{schuhmann2022laion,krylov2021open} and possess strong generative capabilities in producing realistic images and aesthetically pleasing contents~\cite{dhariwal2021diffusion,ho2022classifier}. Consequently, some recent works have leveraged these models for image restoration. These approaches can be broadly classified into three paradigms: zero-shot, training from scratch, and training from pre-trained models.

Zero-shot methods~\cite{pan2021exploiting,chung2022improving,wang2022zero,chung2022diffusion,kawar2022denoising} leverage pre-trained diffusion models as generative priors. During sampling, they incorporate degraded images as condition to tackle image restoration tasks. However, these methods frequently yield suboptimal results with unpredictable artifacts when applied to real-world data. Other approaches train a conditional diffusion model from scratch~\cite{li2022srdiff,ozdenizci2023restoring,ren2023multiscale,murata2023gibbsddrm,chen2020pre,chung2023parallel}. However, they do not possess the advantages of pre-trained Text-to-Image (T2I) diffusion models. In contrast, some recent works have attempted to utilize T2I diffusion models such as Stable Diffusion~\cite{rombach2022high} for image super-resolution~\cite{wang2024exploiting,lin2025diffbir} and deblurring~\cite{luo2023controlling,zhang2024diff,liu2024diff,kong2025deblurdiffrealworldimagedeblurring}. These methods train a lightweight adapter to provide structural conditions for restoration. However, when faced with the unpaired deblurring task, limited realistic blurry-sharp image pairs are available for supervised learning. On the other hand, relying solely on synthetic data often leads to overfitting, resulting in unsatisfactory performance when encountering unseen blur patterns. In our work, UID-Diff is developed based on a pre-trained T2I diffusion model and leverages unpaired data for unsupervised blur pattern learning to facilitate the structure extraction process in real-world blur patterns.

{\bf Image Deblurring.}
Image deblurring methods based on deep learning can be broadly categorized into supervised and unsupervised approaches.

\textit{Supervised Deblurring with Paired Data.}
With the advances of deep learning and large-scale synthetic blurry-sharp image pairs~\cite{hendrycks2019benchmarking,rim2022realistic}, supervised learning based approaches attempt to learn the transfer function from the blurry domain to the sharp domain. These methods can be generally categorized into two main types: prior-free and prior-related models. The former attempt to directly develop a robust model for blur removal, leveraging CNNs~\cite{chen2022simple,cho2021rethinking,li2022learning,mao2023intriguing} or Transformers~\cite{tsai2022stripformer,kong2023efficient,liang2024image,liu2024deblurdinat}. However, the performance of these methods deteriorates significantly when they encounter unseen blur patterns in real-world scenarios. 
On the other hand, the latter approaches aim to learn the blur prior to guide the deblurring network. They predict the prior representations through flow based~\cite{fang2023self,liu2024motion} or diffusion based models~\cite{chen2024hierarchical,laroche2024fast}. Although these approaches achieve improved performance, they cannot learn the prior from unpaired data. This limitation hinders their application in real-world deblurring tasks, where only unpaired data is available.

\textit{Unsupervised Deblurring with Unpaired Data.} Supervised methods trained on synthetic data often underperform on real-world blur patterns due to overfitting. A promising approach to tackle this issue is by leveraging unpaired data from a specific domain. Some works use unpaired data to perform deblurring directly~\cite{zhao2022fcl,zhang2023neural,jiangxin2021learning,ren2020neural,jiang2023uncertainty}, while use auxiliary tasks to enhance the performance of supervised models~\cite{pham2024blur2blur,wu2024id}. 
For the first category, some methods restore details in blurry input images by blur-sharp conversion~\cite{yi2017dualgan,zhao2022fcl}. On the other hand, some models estimate prior knowledge from unpaired data ~\cite{zhang2023neural,jiangxin2021learning,ren2020neural,jiang2023uncertainty,dong2021learning}. These approaches usually underestimate the diversity of blur patterns. Furthermore, they require specific adversarial training~\cite{goodfellow2014generative}, which limits their application to diffusion based models. While the methods in the second category~\cite{pham2024blur2blur,wu2024id} aim to transfer unknown blur into a certain known blur pattern. Unfortunately, these approaches entail additional computational costs and are limited by their inability to conduct large-scale training with various synthetic blur patterns. In our work, UID-Diff tends to decouple the learning process of structural features and blur patterns to integrate the generative diffusion model into the unpaired deblurring task.

\section{Method}

\subsection{Preliminaries}

SD is a type of latent diffusion model~\cite{rombach2022high} that performs a sequence of gradual denoising operations within the latent space and subsequently remaps the denoised latent code into the pixel space to generate the final output image. During the training process, SD initially encodes an input image $x$ into a latent code $z$ using a Variational Auto-Encoder (VAE)~\cite{kingma2013auto}. In the subsequent stages, the noisy latent code $z_t$ at timestep $t$ serves as the input for the denoising U-Net $\epsilon_\theta$, which interacts with text condition $c$ via cross-attention. The training objective for this process is defined as follows:
\begin{equation}
\label{eq:diffusion loss}
\mathcal{L} = \mathbb{E}_{z,c,\epsilon \sim \mathcal{N}(0,1),t} \left[ \| \epsilon - \epsilon_\theta (z_t, t, c) \|_2^2 \right],
\end{equation}
where $\epsilon$ represents random noise sampled from a standard Gaussian distribution.

\subsection{Problem Formulation}

Following~\cite{pham2024blur2blur}, we formulate a blurry image $y$ as a function of the corresponding sharp image $x$ with underlying structural features $f_{s}$ through a blur operator $\mathcal{F}_C(\cdot, k)$, which is associated with a device-dependent blur domain $C$ and a blur kernel $k$:
\begin{equation}
    y = \mathcal{F}_C(x, f_{s}, k) + \eta,
\end{equation}
where $\eta$ is a noise term. Our objective is to develop a function $\mathcal{G}_C$ that can extract structural representations from a blurry image $y \in C$. The extracted features will serve as conditions for a pre-trained generative diffusion model $SD$ with an adapter $\mathcal{A}$ to recover the sharp image, i.e.,
\begin{equation}
\begin{aligned}
    x &= SD\left[y, \mathcal{A}(f_{s}), z\right], \\
    f_{s} &= \mathcal{G}_C(y),
\end{aligned}
\end{equation}
where $z$ is random Gaussian noise. $f_{s}$ is the structural representation shared by $x$ and $y$.

\begin{figure}[t]
	\centering
	\includegraphics[width=0.8\textwidth]{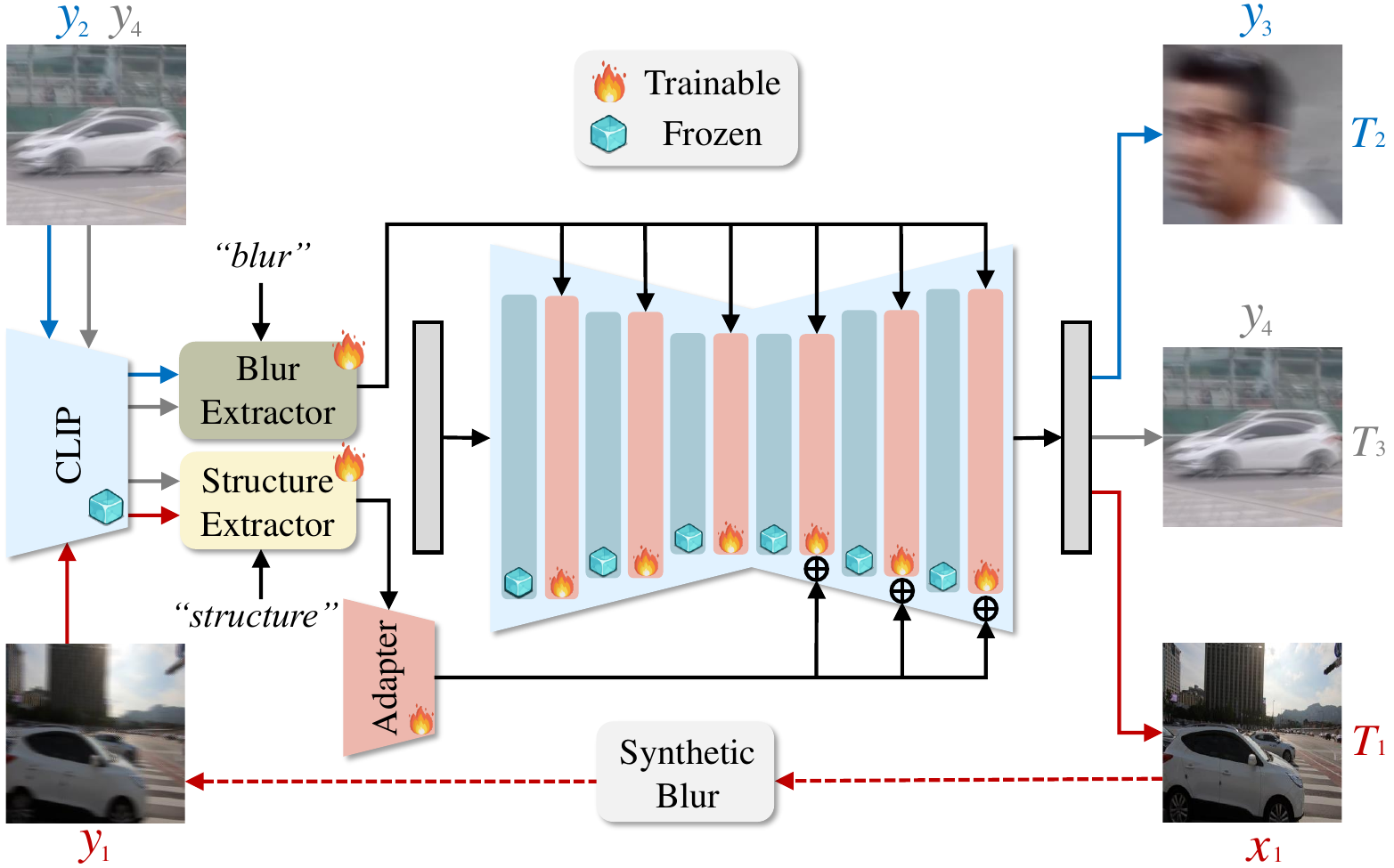}
\caption{Training pipeline of UID-Diff. UID-Diff achieves blur pattern-structure representation decoupling by joint training on three specific tasks (distinguished by arrows of different colors): 1) The deblurring task, denoted as \textcolor[HTML]{C00000}{$T_1$}, enables the structure extractor to learn structural information from synthetic data; 2) The blur-transfer task, denoted as \textcolor[HTML]{0070C0}{$T_2$}, facilitates the blur extractor in learning blur patterns from unpaired blurry images; 3) The reconstruction task, denoted as \textcolor[HTML]{7F7F7F}{$T_3$}, ensures that the extracted structural features and blur patterns are complementary.}
	\label{fig:train}
\end{figure}

\subsection{UID-Diff} 

Q-Former often acts as a representation extractor in conditional image generation~\cite{li2023blip, li2024blip}. Inspired by this, UID-Diff employs two Q-Formers to separately extract structural features and blur patterns from blurry images as illustrated in Figure~\ref{fig:train}. These extracted features serve as generation conditions for three specifically designed training tasks, thereby achieving decoupling: 1) The deblurring task (Section~\ref{sec:task1}) equips a Q-Former $Q_s$ with the ability to extract structural representation from synthetic data pairs. These features are then sent to an adapter $\mathcal{A}$ to provide restoration condition for SD; 2) The blur-transfer task (Section~\ref{sec:task2}) enables another Q-Former $Q_b$ to learn blur patterns from unpaired blurry images within a specific domain; 3) The reconstruction task (Section~\ref{sec:task3}) activates both $Q_s$ and $Q_b$ to ensure that the extracted structural and blur features are complementary, thus enabling $Q_s$ to effectively handle images from specific blur domains that lack paired data. For inference (Section~\ref{sec:inference}), the $Q_s$ and $\mathcal{A}$ are retained to restore sharp images, as illustrated in Figure~\ref{fig:infer}.

\subsection{Deblurring}
\label{sec:task1}

In SD, additional generation conditions related to image structure, such as sketches and depth maps~\cite{cheng2024theatergen,peng2024controlnext}, are often incorporated by adding additional residuals through a ControlNet-like adapter~\cite{zhang2023adding,mou2024adapter}. However, the entanglement of the blur pattern and structural information in the degraded image $y$ makes directly using it as the control signal for $\mathcal{A}$ problematic, as it often leads to instability and produces artifacts~\cite{lin2025diffbir}. To mitigate these issues, we employ a Q-Former $Q_s$ as a feature extractor to capture the underlying structural features from $y$. 

Denoted as $T_1$ in Figure~\ref{fig:train}, we synthetic a blur image $y_1$ from $x_1$ and feed $y_1$ into the CLIP image encoder~\cite{radford2021learning}, whose output interacts with the learnable query tokens of $Q_s$ through cross-attention. In this process, we set the word "structure" as the input text for $Q_s$ in anticipation that it can extract features $f_s$ that capture the low-level structure of the image. We follow previous works~\cite{liu2024diff,lin2025diffbir} to employ an adapter $\mathcal{A}$ for UID-Diff. As illustrated in Figure~\ref{fig:adapter}, $f_s$ which contains the structural features of $y_1$, interacts with the down-sample blocks of $\mathcal{A}$ via cross-attention and is incorporated into the SD U-Net as additional residuals to serve as a structure condition.

The deblurring task is trained non-reconstructive, thus requiring paired blurry-sharp image data. We utilize the method from~\cite{hendrycks2019benchmarking} to synthesize blur to sharp images, thereby creating the necessary paired data for training. The loss function of the deblurring task can be depicted as:
\begin{equation}
\label{eq:loss1}
\mathcal{L}_1 = 
\mathbb{E}_{z^{x_1},c_s,f_s,\epsilon \sim \mathcal{N}(0,1),t} \left[ \| \epsilon - \epsilon_\theta (z_t^{x_1}, c_s, f_s, t) \|_2^2 \right], 
\end{equation}
where $z_t^{x_1}$ is obtained by adding noise to the latent code of $x_1$. The text condition $c_s$ is set as "sharp and clean image'' for the original text cross-attention of SD.

\subsection{Blur-tansfer}
\label{sec:task2}

When the deblurring task is trained solely on synthetic data, it becomes challenging for $Q_s$ to handle images with unseen blur domains $C$. To address this issue, we employ the blur-transfer task that leverages unpaired blurred images from $C$ to help the model better extract structural features. 

In SD, additional generation conditions related to style are typically considered through additional cross attention mechanisms~\cite{ye2023ip,chen2024artadapter}. Inspired by this, we use blur patterns as a special style form. Denoted as $T_2$ in Figure~\ref{fig:train}, we employ another Q-Former $Q_b$ to extract $f_b$ that captures the blur pattern of an image $y_2 \in C$. We set the word "blur'' as the input text of $Q_b$ to extract representations related to the blur pattern. These representations will be fed into SD through additional cross-attention operations. Specifically, given the query features $Z$ and the text features $c$, the output of cross-attention $Z'$ from UID-Diff can be defined by the following equation:
\begin{equation}
Z' = \text{Attention}(Q, K, V) + \text{Attention}(Q, K', V'),
\end{equation}
where $Q = ZW_q$, $K = cW_k$, and $V = cW_v$ are the query, key, and value matrices of the text attention operation, respectively. Here, $W_q$, $W_k$, and $W_v$ are the frozen projection matrices. Additionally, $K' = c_{b}W'_k$ and $V' = c_{b}W'_v$ are the newly added key and value matrices for the blur-pattern attention operation, respectively, with $W'_k$ and $W'_v$ being the new trainable projection matrices.

The optimization goal of the blur transfer task is to utilize $f_b$ as a blur condition to generate another image $y_3 \in C$, with a different structure but in the same blur domain. The loss function can be depicted as follows:
\begin{equation}
\label{eq:loss2}
\mathcal{L}_2 = 
\mathbb{E}_{z^{y_3},c_b,f_b,\epsilon \sim \mathcal{N}(0,1),t} \left[ \| \epsilon - \epsilon_\theta (z_t^{y_3}, c_b, f_b, t) \|_2^2 \right], 
\end{equation}
where the text condition $c_b$ is set as "blurry image'', $f_b$ is the extracted blur features of $y_2$, and $z_t^{y_3}$ is obtained by adding noise to the latent code of $y_3$. In this way, we anticipate that $Q_b$ can learn to extract unknown blur patterns in an unsupervised paradigm.

\subsection{Reconstruction}
\label{sec:task3}

In the aforementioned two tasks, we use text prompts and a decoupled conditioning mechanism, which allows structure and blur patterns to independently serve as conditions for the diffusion model, to guide the two Q-Formers to focus on their specific tasks. However, there may still be optimization biases. Specifically, $Q_s$ might learn the inverse function of synthetic blur, causing it to still perform poorly in unknown domains. To circumvent this, we employ the reconstruction task, which is illustrated as $T_3$ in Figure~\ref{fig:train}. In this task, $Q_s$ and $Q_b$ are used to extract structure and blur patterns from the same image $y_4 \in C$. These features are simultaneously considered as generation conditions to reconstruct $y_4$, with the loss function depicted as:
\begin{equation}
\label{eq:loss3}
\mathcal{L}_3 = 
\mathbb{E}_{z^{y_4},c_b,c_s,f_b,f_c,\epsilon \sim \mathcal{N}(0,1),t} \left[ \| \epsilon - \epsilon_\theta (z_t^{y_4}, c_b, c_s, f_b,f_c, t) \|_2^2 \right],
\end{equation}
where $z_t^{y_4}$ is obtained by adding noise to the latent code of a blur image from $C$. This process enables the structure and blur patterns to complement each other, enhancing the generalization capability of $Q_s$ when facing realistic blur patterns.

\subsection{Training and Inference}
\label{sec:inference}

For fast convergence, we first train the deblurring and blur-transfer tasks separately, then incorporate the reconstruction task for joint training. The loss function of the joint training process can be formulated as follows:
\begin{equation}
\label{eq:our loss}
\mathcal{L} = \alpha  \cdot \mathcal{L}_1 + \beta \cdot \mathcal{L}_2 + \gamma \cdot \mathcal{L}_3.
\end{equation}
Here, $\mathcal{L}_1$, $\mathcal{L}_2$, and $\mathcal{L}_3$ represent the loss functions for the deblurring task, the blur-transfer task, and the reconstruction task, respectively; $\alpha + \beta + \gamma = 1$ stands for the sample weights for these three tasks, where we set them to 1:1:1. During the training process, only the Q-Formers, the adapter and the newly added projection metrics are optimized. 

\begin{figure}[!t]
    \centering
    \begin{minipage}[t]{0.64\textwidth} 
        \centering
        \includegraphics[width=\textwidth]{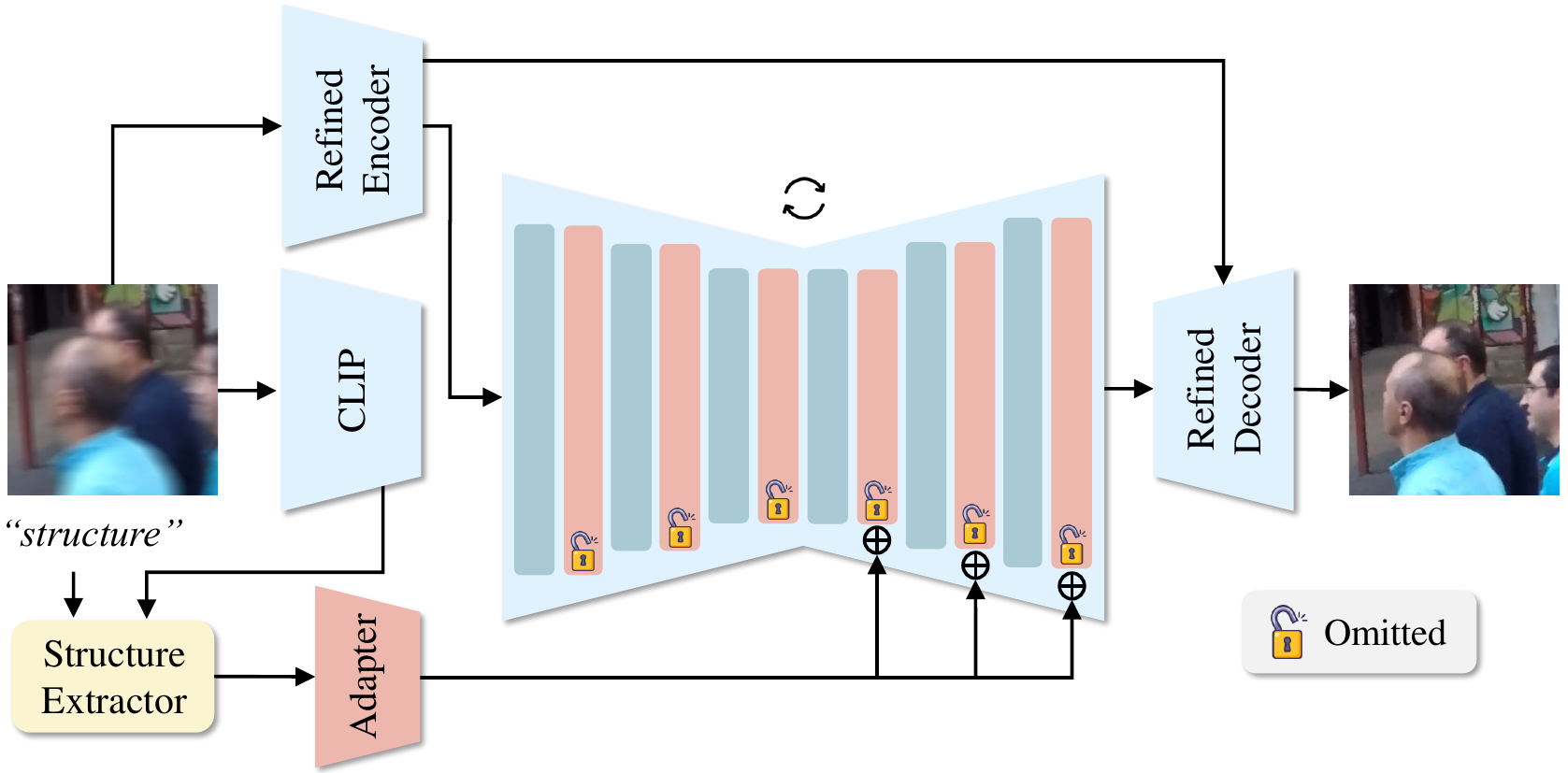}
        \caption{Inference process of UID-Diff. We preserve the $Q_s$ and $\mathcal{A}$ to generate structure guidance, while the newly added cross-attention layer for blur-transfer is omitted.}
        \label{fig:infer}
    \end{minipage}
    \hfill 
    \begin{minipage}[t]{0.34\textwidth}
        \centering
        \includegraphics[width=\textwidth]{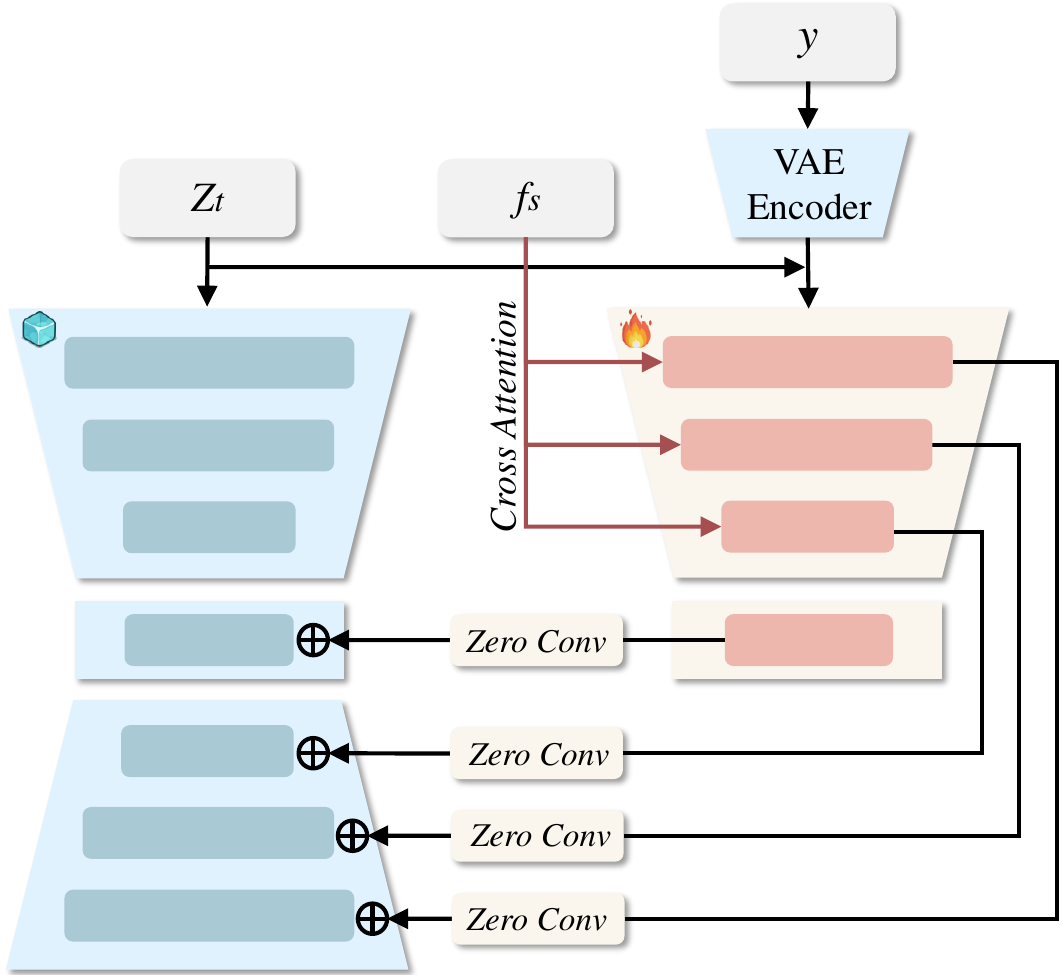}
        \caption{Schematic illustration of the adapter of UID-Diff.}
        \label{fig:adapter}
    \end{minipage}
\end{figure}

The inference process of UID-Diff is illustrated in Figure~\ref{fig:infer}. We retain $Q_s$ and $\mathcal{A}$ to extract structural features, while the newly added cross-attention for the blur-transfer is omitted. We note that SD tends to excessively embellish details, resulting in deviations from the original image. To alleviate this phenomenon, we replace the original SD VAE with a refined-VAE\footnote{See Appendix A for more details.}, based on previous works~\cite{chen2025unirestore,zhang2024diff}.

\section{Experiments}
\label{sec:exp}
\subsection{Datasets and Metrics}

We evaluate our UID-Diff on four widely-used real-world datasets: GoPro~\cite{nah2017deep}, RealBlur-J\&R~\cite{rim2020real}, and REDS~\cite{nah2019ntire}. To emulate real-world deblurring scenarios without paired data, we adhere to existing works~\cite{pham2024blur2blur} to split the training set into distinct blurry subset $\mathcal{B}$ and sharp subset $\mathcal{S}$. Since generative-diffusion-based methods require large-scale paired data for training, we employ the synthetic blur technique proposed by~\cite{rim2020real} to generate synthetic blur data $\mathcal{B'}$ from $\mathcal{S}$, thereby providing adequate blurry-sharp image pairs for supervised training. The statistics of source image sets are reported in Table~\ref{tab:Data description}. To effectively evaluate image quality, we include several no-reference image quality assessment (NR-IQA) metrics: MANIQA~\cite{yang2022maniqa}, LIQE~\cite{zhang2023blind}, MUSIQ~\cite{ke2021musiq}, and CLIP-IQA~\cite{wang2023exploring}, which align with prior works~\cite{lin2025diffbir,zhang2024diff} for fair comparison\footnote{See Appendix B for additional results.}.

\begin{table}[!t]
  \caption{Quantitative comparison with SOTA methods on real-world datasets. \textbf{Bold} and \uline{underlined} indicate the best and the second-best performance, respectively.}
  \label{tab:Model Performance on No-Reference Metrics}
  \centering
\resizebox{\linewidth}{!}{
\begin{tabular}{clcccccccccc}
\toprule
Data & Metrics & AirNet &PromptIR &FFTformer &AdaRevD &HI-Diff &Blur2Blur &Diff-Plugin &DiffBir &DA-CLIP &\cellcolor[HTML]{E6F0E8}\textbf{Ours} \\
\midrule \midrule
\multirow{4}{*}{GoPro}
& MANIQA$\uparrow$  & 0.1927 & 0.2248 & 0.2736 & 0.2938 & 0.2532 &  0.2321  & 0.2933 &  \uline{0.3012} & 0.2873 & \cellcolor[HTML]{E6F0E8}\textbf{0.3246}  \\
& LIQE$\uparrow$ & 1.08 & 1.06 & 1.18 & \uline{1.31} & 1.13  &  1.23 &   1.29 &   1.24  & 1.03 & \cellcolor[HTML]{E6F0E8}\textbf{1.97}   \\
& MSUIQ$\uparrow$ &   36.86  &  36.19 & 43.11 & 47.53  & 39.09 & 40.31 & 41.09 & \uline{50.33} & 32.06 & \cellcolor[HTML]{E6F0E8}\textbf{54.94} \\
& CLIP-IQA$\uparrow$ & 0.1904  & 0.1396 & 0.1999 & 0.2831 & 0.1824 & 0.2342  &  0.2458  &  \uline{0.2839} &  0.1184 & \cellcolor[HTML]{E6F0E8}\textbf{0.2978}    \\
\midrule
\multirow{4}{*}{REDS}
& MANIQA$\uparrow$  &  0.2652  & 0.3072 & 0.3602 & 0.3233 & 0.2893 &  0.3342 & 0.3622 & \uline{0.3697} & 0.3526 & \cellcolor[HTML]{E6F0E8}\textbf{0.3824} \\
& LIQE$\uparrow$  & 1.44  & 1.86 & 2.33 & \uline{2.51}  & 2.43 &  1.93 & 2.03 & \uline{2.51} &  1.87 & \cellcolor[HTML]{E6F0E8}\textbf{2.85}           \\
& MSUIQ$\uparrow$ & 44.27 & 54.39 & 48.34& 57.42 & 58.99 &   50.32& 56.49& \uline{59.95}  &  52.57 & \cellcolor[HTML]{E6F0E8}\textbf{63.91}   \\
& CLIP-IQA$\uparrow$  & 0.1980& 0.2982 & 0.4094 &  \uline{0.5903} &  0.5542 & 0.2392 & 0.3968 & \textbf{0.6086}  &  0.2889 & \cellcolor[HTML]{E6F0E8}\textbf{0.6086}  \\
\midrule
\multirow{4}{*}{RealBlur-J}
& MANIQA$\uparrow$  &  0.1896 & 0.1933 & 0.2699 & 0.2603 & \uline{0.2754} &  0.2102 & 0.2513 &  0.2688  & 0.2642 & \cellcolor[HTML]{E6F0E8}\textbf{0.3092}    \\
& LIQE$\uparrow$ & 1.07 & 1.14 & 1.78 & \uline{2.07} & 1.24 & 1.12  & 1.17  & 1.37  & 1.33 &\cellcolor[HTML]{E6F0E8}\textbf{2.11}       \\
& MSUIQ$\uparrow$ & 41.11 & 45.23 &  45.03 & 41.41 & 44.42 &    43.98 & \uline{45.31} & 44.85 & 41.72 & \cellcolor[HTML]{E6F0E8}\textbf{47.06}  \\
& CLIP-IQA$\uparrow$ & 0.1305 & 0.1749 & 0.1976 & \uline{0.2378} & 0.2217 &  0.1623 & 0.1546 & 0.2098 & 0.1363 & \cellcolor[HTML]{E6F0E8}\textbf{0.2392} \\
\midrule
\multirow{4}{*}{RealBlur-R}
& MANIQA$\uparrow$ & 0.2598 & 0.2731 & 0.2786 & 0.3092 & 0.3113 & 0.2873 &  0.3026 & \uline{0.3198}  &0.2945 & \cellcolor[HTML]{E6F0E8}\textbf{0.3577}  \\
& LIQE$\uparrow$  & 1.25 & 1.69 &  1.88 & 2.64 & 2.03 &  1.67 & 1.89 & \uline{2.71} & 1.12 & \cellcolor[HTML]{E6F0E8}\textbf{2.81}            \\
& MSUIQ$\uparrow$ & 38.19 & 46.22 & 42.34 & 46.63 & 49.39 &  40.32  & 48.43 & \uline{51.72}& 21.72 & \cellcolor[HTML]{E6F0E8}\textbf{57.19}   \\
& CLIP-IQA$\uparrow$ & 0.2131 & 0.2987 & 0.3984 & \uline{0.5132} & 0.4987 & 0.2451 & 0.2803 & 0.4885 & 0.2761 & \cellcolor[HTML]{E6F0E8}\textbf{0.5213}  \\
\bottomrule
\end{tabular}}
\end{table}

\subsection{Implementation Details}
\label{sec:details}

\begin{wraptable}{r}{0.5\textwidth}
 \captionsetup{font=small}
\vspace{-1.25em}
\caption{\small Statistics of data used as unknown domains.}
\label{tab:Data description}
\centering
\resizebox{\linewidth}{!}{
\begin{tabular}{lcccc}
\toprule
\multicolumn{1}{c}{\multirow{2}{*}{Dataset}} & \multicolumn{4}{c}{Number of data samples}    \\
\multicolumn{1}{c}{}                         & Sharp ($\mathcal{S})$  & Blur ($\mathcal{B})$ & Synthetic blur ($\mathcal{B'})$ & Test \\
\midrule
GoPro                                        & 1261  & 842   & 4210           & 1111 \\
REDS                                         & 14400 & 9600  & 48000          & 3000 \\
RealBlur-J                                   & 2064  & 1375  & 6875           & 1474 \\
RealBlur-R                                   & 2064  & 1375  & 6875           & 1474 \\
\bottomrule
\end{tabular}}
\end{wraptable}

We leverage the stable diffusion v1.5 as the pre-trained generative diffusion model. 
For the image encoder, we employ the ViT-L/14 model from CLIP~\cite{radford2021learning} and set the number of learnable query tokens in the Q-Former to 16, consistent with BLIP-Diffusion~\cite{li2024blip}. The structural and blur pattern extractors are initialized using the pre-trained weights provided by BLIP-Diffusion for fast convergence. The training process of UID-Diff includes two phases. First, we train the deblurring and blur-transfer tasks for 50,000 steps as a warm-up. Then, we conduct joint training across all tasks with a 1:1:1 sampling ratio for an additional 500,000 steps. All experiments are executed on NVIDIA A100 GPUs. We utilize the AdamW~\cite{loshchilov2017decoupled} optimizer with a uniform learning rate of $5 \times 10^{-5}$ across all training tasks. For the inference phase, we adapt the DDIM sampler~\cite{Song_Meng_Ermon_2020} with 30 steps to generate outputs. The guidance scale for classifier-free guidance~\cite{ho2022classifier} is set to 7.5. Completing all training stages of UID-Diff takes approximately 2 days. During inference, it requires around 20GB of memory on a single A100 GPU and takes about 25 seconds per image.

\subsection{Comparisons with State-of-the-art Methods}
\label{sec:Comparison}

In this section, we compare UID-Diff with several SOTA methods, including the generative-diffusion-based approaches DiffBir~\cite{lin2025diffbir}, Diff-Plugin~\cite{liu2024diff}, and DA-CLIP~\cite{luo2023controlling}, as well as other advanced methods: AdaRevD~\cite{xintm2024AdaRevD}, FFTformer~\cite{kong2022efficientfrequencydomainbasedtransformers}, HI-Diff~\cite{chen2023hierarchical}, PromptIR~\cite{potlapalli2024promptir}, AirNet~\cite{chee2018airnet}, and Blur2Blur (unpaired training)~\cite{pham2024blur2blur}.

\textbf{Quantitative Comparisons.} The performance comparisons on the test sets of our selected datasets are presented in Table~\ref{tab:Model Performance on No-Reference Metrics}. UID-Diff outperforms all baseline models, including the latest unpaired training approach (Blur2Blur) and supervised training generative-diffusion-based method (Diff-Plugin). The results indicate that paired training models exhibit suboptimal performance when faced with unknown domain blur patterns in these real-world datasets. Non-generative-diffusion-based unpaired training method Blur2Blur demonstrates some effectiveness in these scenarios, but it still falls short compared to generative-diffusion-based approaches in most cases. This can be attributed to the capacity of generative diffusion models to generate intricate textures and details through extensive pre-training and aesthetic alignment. Our UID-Diff harnesses the advantages of generative diffusion models and enhances the generalization capability of structure extraction through decoupled training, which results in favorable performance against existing SOTA methods on unpaired deblurring tasks.

\begin{figure*}[!t]
  \centering
  \setlength{\tabcolsep}{0.5pt} 
  \begin{tabular}{cccccccccccc}
    &  \scriptsize Ground Truth & \scriptsize AdaRevD & \scriptsize PromptIR & \scriptsize Blur2Blur & \scriptsize HI-Diff & \scriptsize DA-CLIP & \scriptsize DiffBir & \scriptsize Diff-Plugin & \scriptsize \textbf{Ours}  & \scriptsize Blur \\
    \raisebox{1.5\height}{\rotatebox[origin=c]{90}{\scriptsize GoPro}} &
    \includegraphics[width=.0965\textwidth]{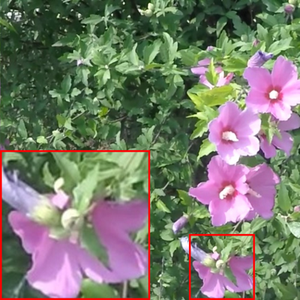} & 
    \includegraphics[width=.0965\textwidth]{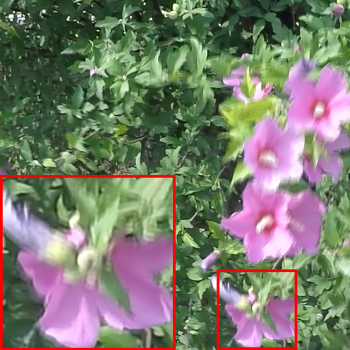} & 
    \includegraphics[width=.0965\textwidth]{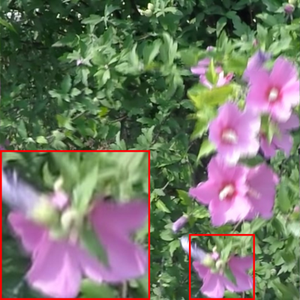} & 
    \includegraphics[width=.0965\textwidth]{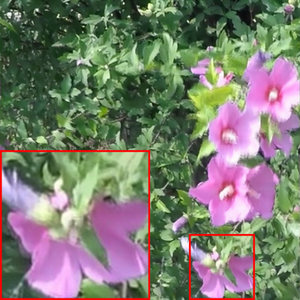} & 
    \includegraphics[width=.0965\textwidth]{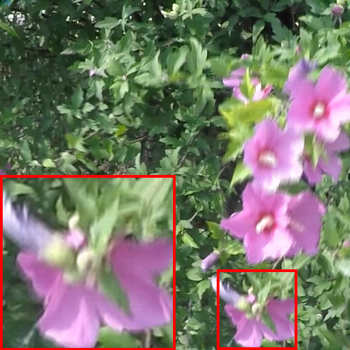} & 
    \includegraphics[width=.0965\textwidth]{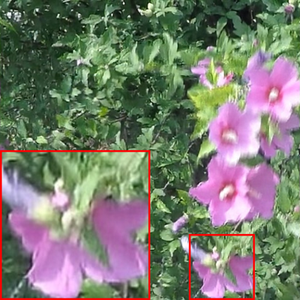} & 
    \includegraphics[width=.0965\textwidth]{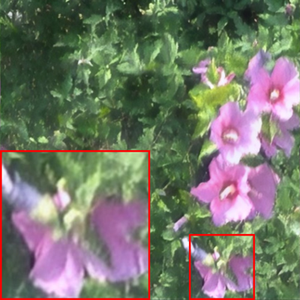} & 
    \includegraphics[width=.0965\textwidth]{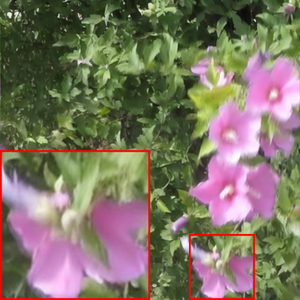} & 
    \includegraphics[width=.0965\textwidth]{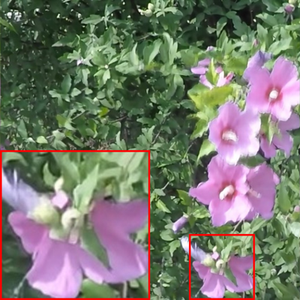} &
    \includegraphics[width=.0965\textwidth]{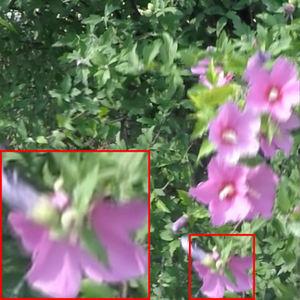} \\
   \multicolumn{10}{c}{\vspace{-13.5pt}} \\
    \raisebox{1.55\height}{\rotatebox[origin=c]{90}{\scriptsize REDS}} &
    \includegraphics[width=.0965\textwidth]{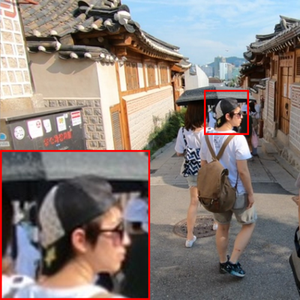} & 
    \includegraphics[width=.0965\textwidth]{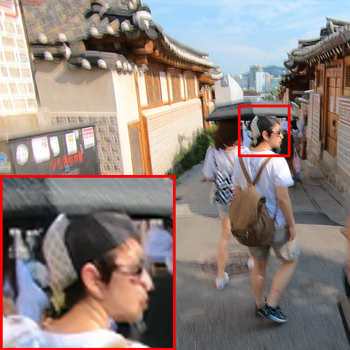} & 
    \includegraphics[width=.0965\textwidth]{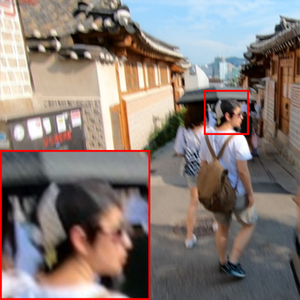} & 
    \includegraphics[width=.0965\textwidth]{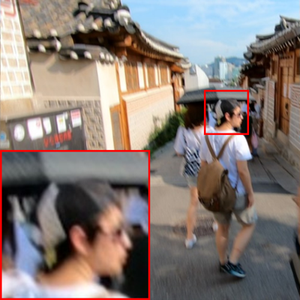} & 
    \includegraphics[width=.0965\textwidth]{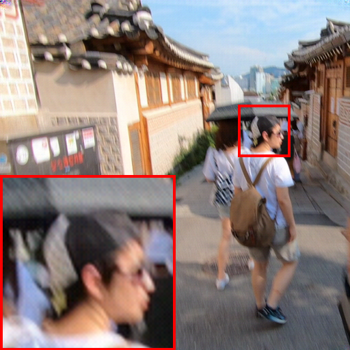} & 
    \includegraphics[width=.0965\textwidth]{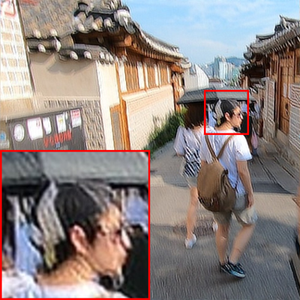} & 
    \includegraphics[width=.0965\textwidth]{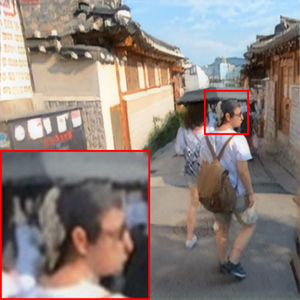} & 
    \includegraphics[width=.0965\textwidth]{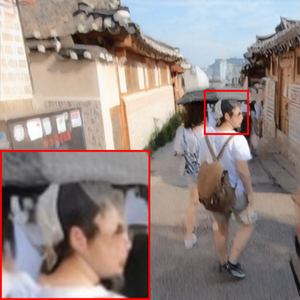} & 
    \includegraphics[width=.0965\textwidth]{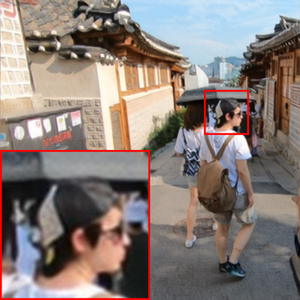} &
    \includegraphics[width=.0965\textwidth]{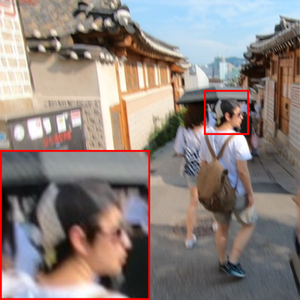} \\
   \multicolumn{10}{c}{\vspace{-13.5pt}} \\
    \raisebox{1\height}{\rotatebox[origin=c]{90}{\scriptsize RealBlur-J}} &
    \includegraphics[width=.0965\textwidth]{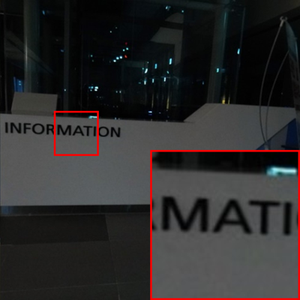} & 
    \includegraphics[width=.0965\textwidth]{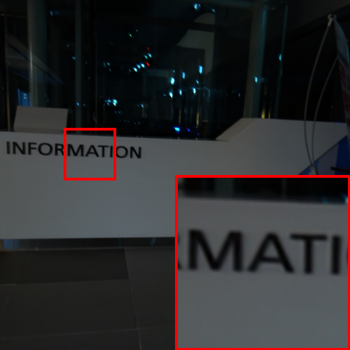} & 
    \includegraphics[width=.0965\textwidth]{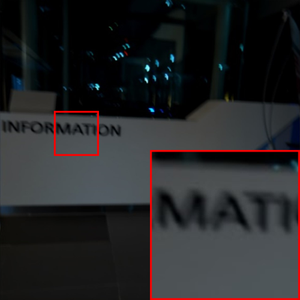} & 
    \includegraphics[width=.0965\textwidth]{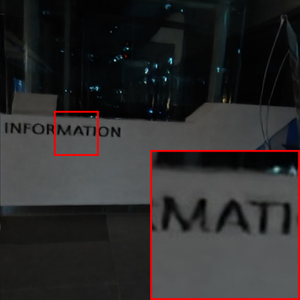} & 
    \includegraphics[width=.0965\textwidth]{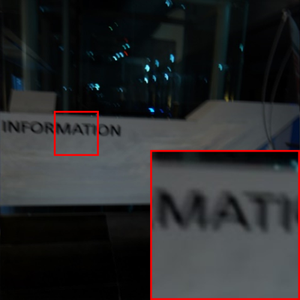} & 
    \includegraphics[width=.0965\textwidth]{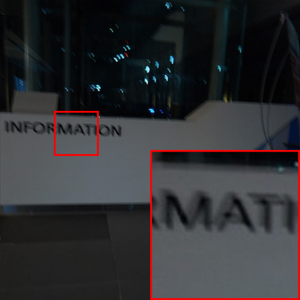} & 
    \includegraphics[width=.0965\textwidth]{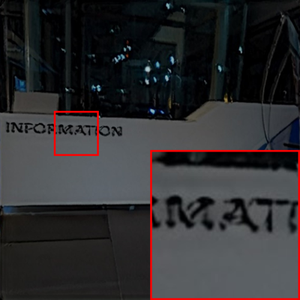} & 
    \includegraphics[width=.0965\textwidth]{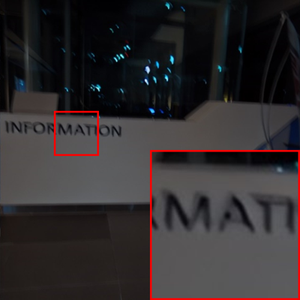} & 
    \includegraphics[width=.0965\textwidth]{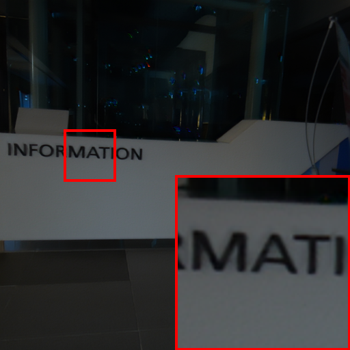} &
    \includegraphics[width=.0965\textwidth]{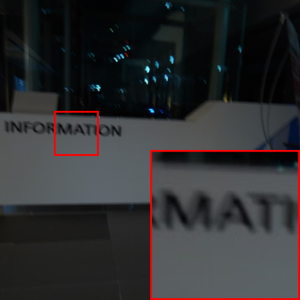} \\
   \multicolumn{10}{c}{\vspace{-13.5pt}} \\
    \raisebox{1\height}{\rotatebox[origin=c]{90}{\scriptsize RealBlur-R}} &
    \includegraphics[width=.0965\textwidth]{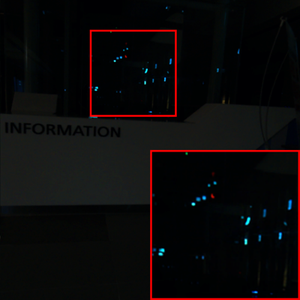} & 
    \includegraphics[width=.0965\textwidth]{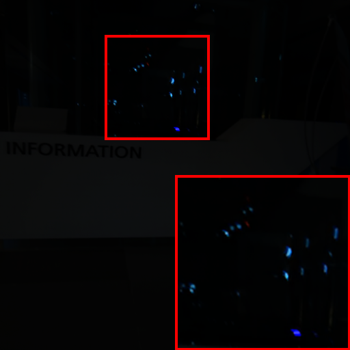} & 
    \includegraphics[width=.0965\textwidth]{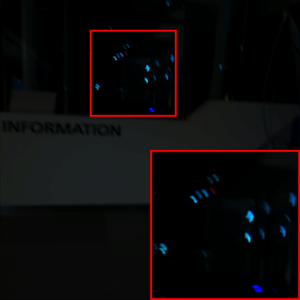} & 
    \includegraphics[width=.0965\textwidth]{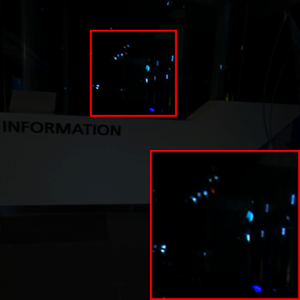} & 
    \includegraphics[width=.0965\textwidth]{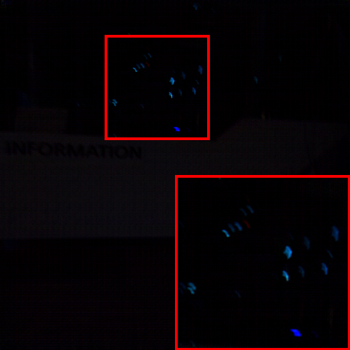} & 
    \includegraphics[width=.0965\textwidth]{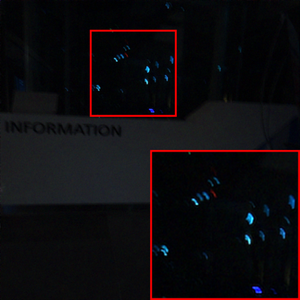} & 
    \includegraphics[width=.0965\textwidth]{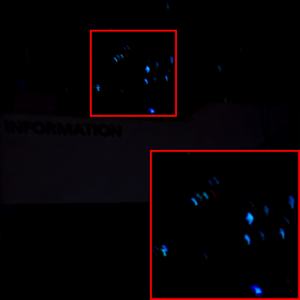} & 
    \includegraphics[width=.0965\textwidth]{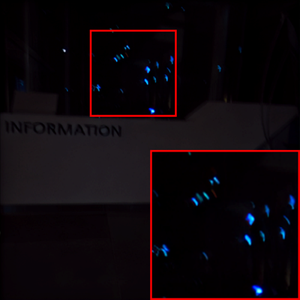} & 
    \includegraphics[width=.0965\textwidth]{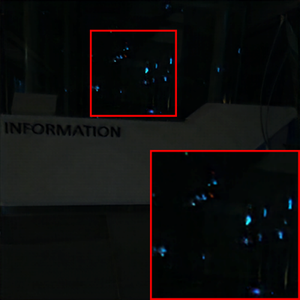} &
    \includegraphics[width=.0965\textwidth]{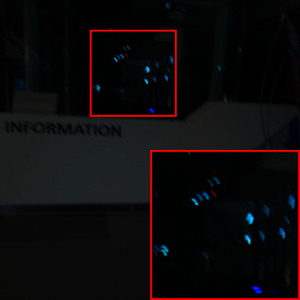} \\
  \end{tabular}
  \vspace{-0.6em}
\caption{Visual comparison of deblurring results. UID-Diff outperforms other SOTA methods in real-world deblurring scenarios.}
\vspace{-1em}
  \label{fig:visualization of deblur}
\end{figure*}

\begin{figure*}[!t]
  \centering
  \setlength{\tabcolsep}{0.5pt} 
  \begin{tabular}{ccccccccc}
    &  \scriptsize Ground Truth & \scriptsize Diff-Plugin & \scriptsize DiffBir & \scriptsize DA-CLIP & \scriptsize Ours w/o $Q_s$ & \scriptsize Ours w/o $T_{2}T_{3}$ & \scriptsize  Ours w/o $T_{3}$ & \scriptsize \textbf{Ours} \\
    \raisebox{1.8\height}{\rotatebox[origin=c]{90}{\scriptsize GoPro}} &
    \includegraphics[width=.1210\textwidth]{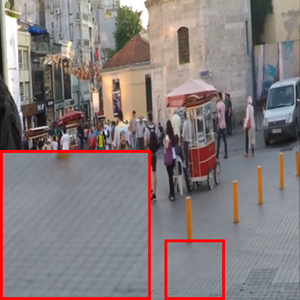} & 
    \includegraphics[width=.1210\textwidth]{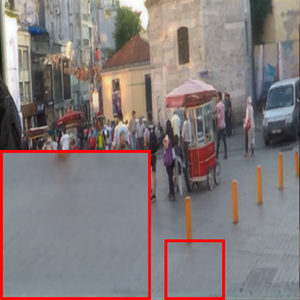} & 
    \includegraphics[width=.1210\textwidth]{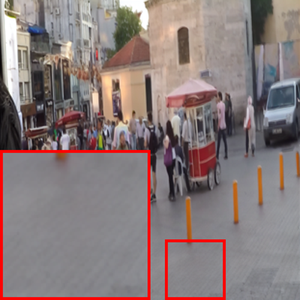} & 
    \includegraphics[width=.1210\textwidth]{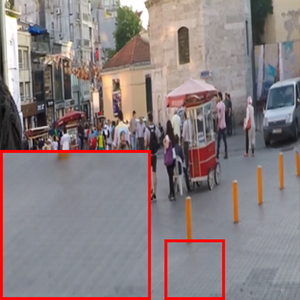} & 
    \includegraphics[width=.1210\textwidth]{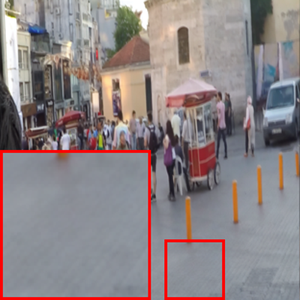} & 
    \includegraphics[width=.1210\textwidth]{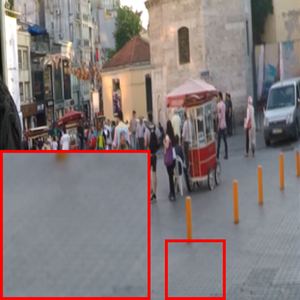} & 
    \includegraphics[width=.1210\textwidth]{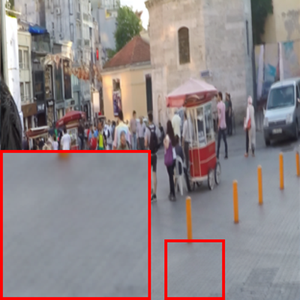} & 
    \includegraphics[width=.1210\textwidth]{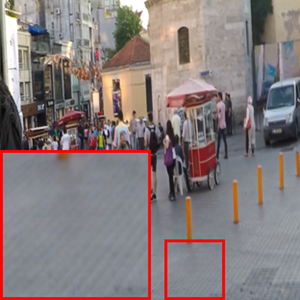} \\
   \multicolumn{9}{c}{\vspace{-13.5pt}} \\
    \raisebox{2\height}{\rotatebox[origin=c]{90}{\scriptsize REDS}} &
    \includegraphics[width=.1210\textwidth]{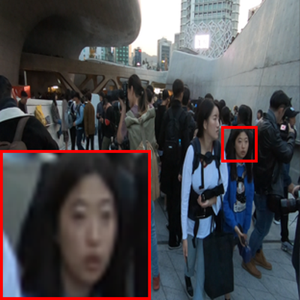} & 
    \includegraphics[width=.1210\textwidth]{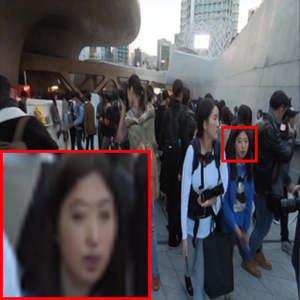} & 
    \includegraphics[width=.1210\textwidth]{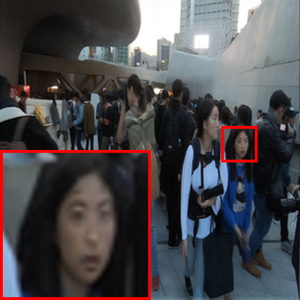} & 
    \includegraphics[width=.1210\textwidth]{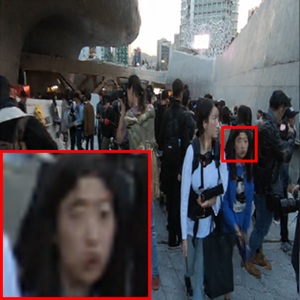} & 
    \includegraphics[width=.1210\textwidth]{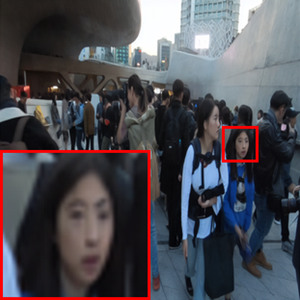} & 
    \includegraphics[width=.1210\textwidth]{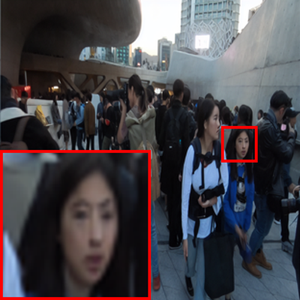} & 
    \includegraphics[width=.1210\textwidth]{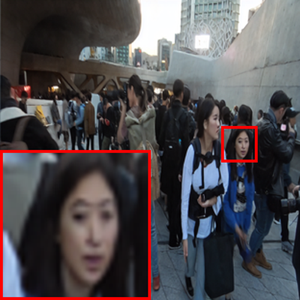} & 
    \includegraphics[width=.1210\textwidth]{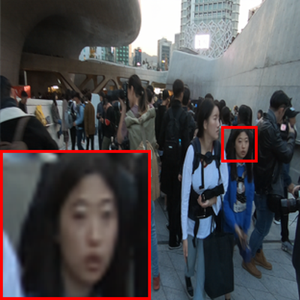} \\
   \multicolumn{9}{c}{\vspace{-13.5pt}} \\
    \raisebox{1.2\height}{\rotatebox[origin=c]{90}{\scriptsize RealBlur-J}} &
    \includegraphics[width=.1210\textwidth]{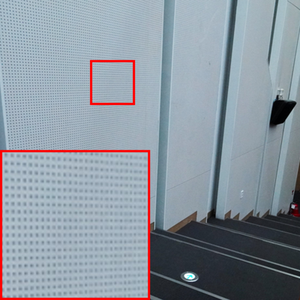} & 
    \includegraphics[width=.1210\textwidth]{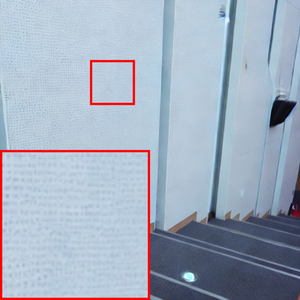} & 
    \includegraphics[width=.1210\textwidth]{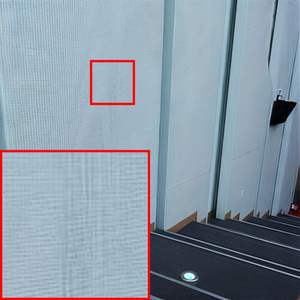} & 
    \includegraphics[width=.1210\textwidth]{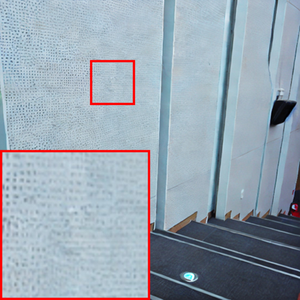} & 
    \includegraphics[width=.1210\textwidth]{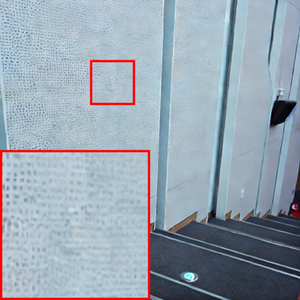} & 
    \includegraphics[width=.1210\textwidth]{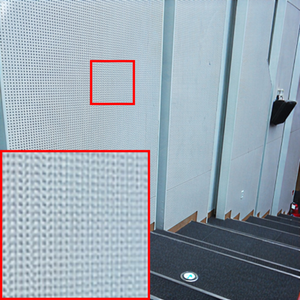} & 
    \includegraphics[width=.1210\textwidth]{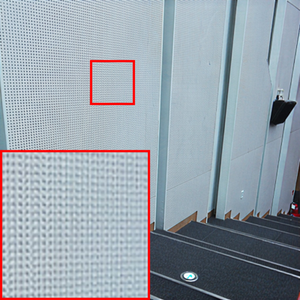} & 
    \includegraphics[width=.1210\textwidth]{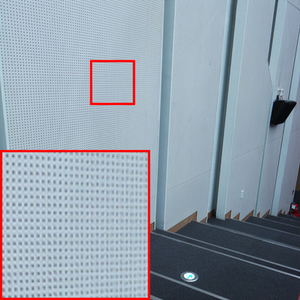} \\
   \multicolumn{9}{c}{\vspace{-13.5pt}} \\
    \raisebox{1.2\height}{\rotatebox[origin=c]{90}{\scriptsize RealBlur-R}} &
    \includegraphics[width=.1210\textwidth]{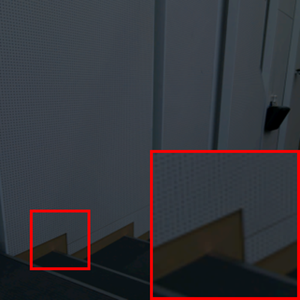} & 
    \includegraphics[width=.1210\textwidth]{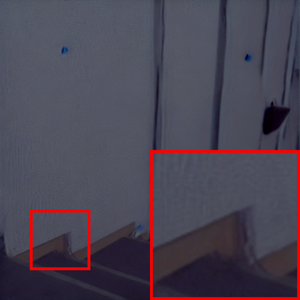} & 
    \includegraphics[width=.1210\textwidth]{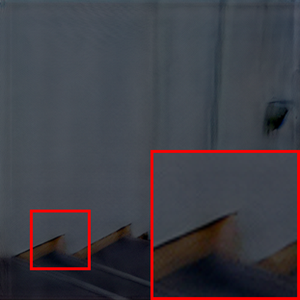} & 
    \includegraphics[width=.1210\textwidth]{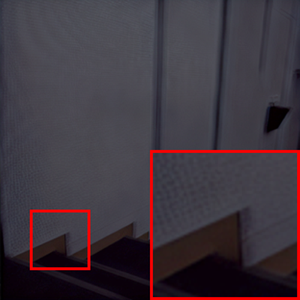} & 
    \includegraphics[width=.1210\textwidth]{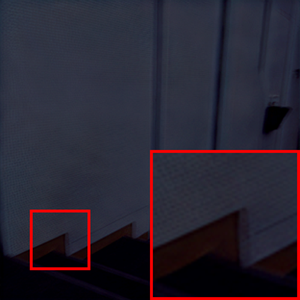} & 
    \includegraphics[width=.1210\textwidth]{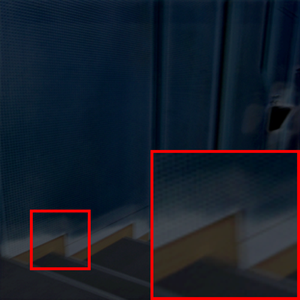} & 
    \includegraphics[width=.1210\textwidth]{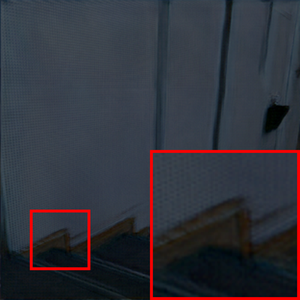} & 
    \includegraphics[width=.1210\textwidth]{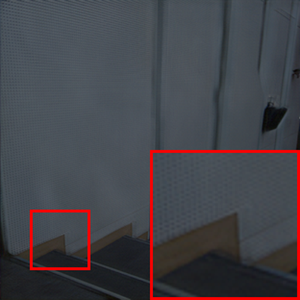} \\
  \end{tabular}
   \vspace{-0.6em}
\caption{Visual comparison on structural reconstruction. UID-Diff outperforms other generative-diffusion-based methods in preserving fine details.}
\vspace{-1.3em}
\label{fig:visualization of structural}
\end{figure*}

\textbf{Qualitative Comparisons.} To assess the visual quality of the results generated by different models, we present their deblurred outputs on the test set of our selected datasets. Figure~\ref{fig:visualization of deblur} illustrates the effectiveness of UID-Diff in deblurring across various scenarios, including nighttime, indoor, and wild real-world scenes. On the other hand, existing supervised training methods struggle to remove real blur patterns. In contrast, the performance of the deblurring module limits unpaired training methods and fails to achieve satisfactory visual results. We observe that using inappropriate features as input for generative-diffusion-based methods may lead to artifacts and distortions. Based on this, we compare the structure reconstruction performance of UID-Diff with existing supervised generative-diffusion-based methods and with three variants\footnote{Detailed in Section~\ref{sec:ablation}.} of our model that only processes supervised training. As demonstrated in Figure~\ref{fig:visualization of structural}, UID-Diff performs favorably in reconstructing detailed structures such as facial features, text, and intricate patterns. In contrast, other supervised diffusion-based methods struggle to effectively decouple structure and blur patterns in real-world scenarios, resulting in distortions and artifacts in the restored images. Please refer to Appendix C for more visual results.

\begin{figure}[!t]
  \centering
  \setlength{\tabcolsep}{1pt} 
  \begin{tabular}{ccccccc}
   \scriptsize  Ground Truth & \scriptsize  Ours w/o $Q_s$ &\scriptsize  Ours w/o $T_{2}T_{3}$ & \scriptsize Ours w/o $T_{3}$ & \scriptsize Ours w/o r-VAE & \scriptsize \textbf{Ours} & \scriptsize   Blur \\
    \includegraphics[width=.138\textwidth]{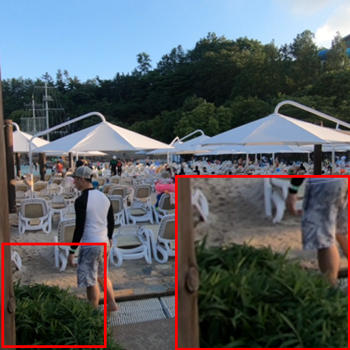} & 
    \includegraphics[width=.138\textwidth]{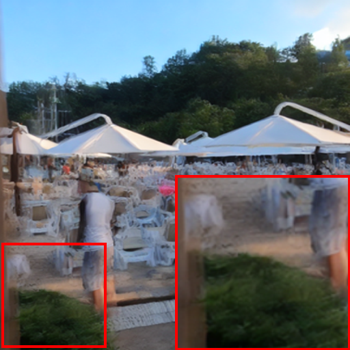} & 
    \includegraphics[width=.138\textwidth]{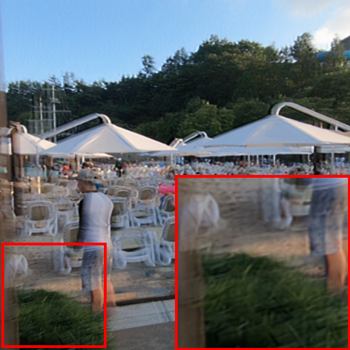} & 
    \includegraphics[width=.138\textwidth]{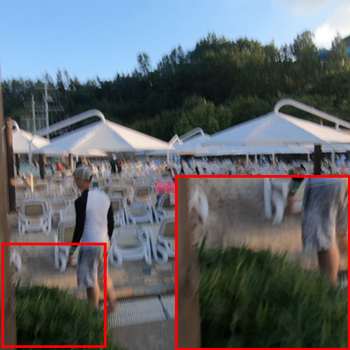} & 
    \includegraphics[width=.138\textwidth]{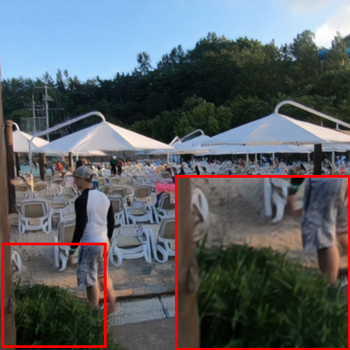} &
    \includegraphics[width=.138\textwidth]{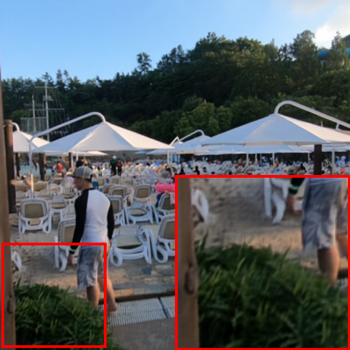} &
    \includegraphics[width=.138\textwidth]{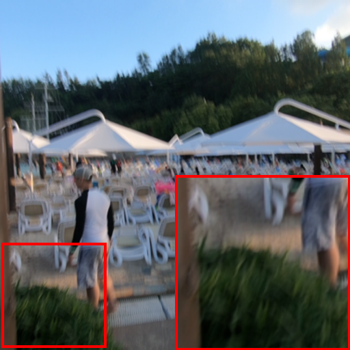} \\
  \end{tabular}
  \vspace{-0.7em}
	\caption{Deblurring results of ablation studies. Key regions are marked with red boxes and magnified.}
   \vspace{-1em}
	\label{fig:ablation}
\end{figure}

\subsection{Ablation Study}
\label{sec:ablation}

\begin{table}[!t]
  \caption{Results of ablation study.}
  \label{tab:No-Reference Metrics Results of Ablation Studies}
  \centering
\resizebox{.83\linewidth}{!}{
\begin{tabular}{cl|ccc|ccc|cccc}
\toprule
& \multicolumn{1}{c}{} & \multicolumn{3}{c}{Components} & \multicolumn{3}{c}{Training Tasks} & \multicolumn{4}{c}{Metrics} \\
\multirow{-2}{*}{Dataset}    & \multicolumn{1}{c|}{\multirow{-2}{*}{Model}} & \multicolumn{1}{c}{$Q_s$} & \multicolumn{1}{c}{$Q_b$} & \multicolumn{1}{c|}{r-VAE} & $T_{1}$ & $T_{2}$ & $T_{3}$ & MANIQA↑ & LIQE↑ & MUSIQ↑ & CLIP-IQA↑ \\
\midrule \midrule
& w/o $Q_s$  & \ding{55} & \ding{55} & \checkmark & \checkmark  & \ding{55} & \ding{55} & 0.2797& 1.29     & 41.22     & 0.2469                                 \\
& w/o $T_{2}T_{3}$ & \checkmark  & \ding{55} & \checkmark  & \checkmark    & \ding{55}    & \ding{55}      & 0.2933     & 1.91     & 52.33     & 0.2599                                  \\
& w/o $T_{3}$   & \checkmark  & \checkmark & \checkmark  & \checkmark    & \checkmark    & \ding{55}        & 0.3124 & 1.95    & 50.88    & 0.2659\\
& w/o r-VAE & \checkmark  & \checkmark & \ding{55} & \checkmark   & \checkmark      & \checkmark & 0.3240 & 1.86& 54.69	& 0.2966 \\
\multirow{-5}{*}{GoPro} & \cellcolor[HTML]{E6F0E8}\textbf{Ours} & \cellcolor[HTML]{E6F0E8}\checkmark & \cellcolor[HTML]{E6F0E8}\checkmark & \cellcolor[HTML]{E6F0E8}\checkmark & \cellcolor[HTML]{E6F0E8}\checkmark & \cellcolor[HTML]{E6F0E8}\checkmark & \cellcolor[HTML]{E6F0E8}\textbf{\checkmark}& \cellcolor[HTML]{E6F0E8}\textbf{0.3246} & \cellcolor[HTML]{E6F0E8}\textbf{1.97} & \cellcolor[HTML]{E6F0E8}\textbf{54.94} & \cellcolor[HTML]{E6F0E8}\textbf{0.2978}                \\
\midrule

& w/o $Q_s$  & \ding{55} & \ding{55} & \checkmark & \checkmark  & \ding{55} & \ding{55} & 0.3698        & 2.33     & 57.05     & 0.4003                                 \\
& w/o $T_{2}T_{3}$ & \checkmark  & \ding{55} & \checkmark  & \checkmark    & \ding{55}    & \ding{55}      & 0.3703      & 1.91     & 54.15     & 0.3433                                  \\
& w/o $T_{3}$   & \checkmark  & \checkmark & \checkmark  & \checkmark    & \checkmark    & \ding{55}        & 0.3729 & 2.23     & 61.89     & 0.5402 \\
& w/o r-VAE & \checkmark  & \checkmark & \ding{55} & \checkmark   & \checkmark      & \checkmark & 0.3817 & 2.81 & 63.52 & 0.6066 \\
\multirow{-5}{*}{REDS} & \cellcolor[HTML]{E6F0E8}\textbf{Ours} & \cellcolor[HTML]{E6F0E8}\checkmark & \cellcolor[HTML]{E6F0E8}\checkmark & \cellcolor[HTML]{E6F0E8}\checkmark & \cellcolor[HTML]{E6F0E8}\checkmark & \cellcolor[HTML]{E6F0E8}\checkmark & \cellcolor[HTML]{E6F0E8}\textbf{\checkmark}& \cellcolor[HTML]{E6F0E8}\textbf{0.3824} & \cellcolor[HTML]{E6F0E8}\textbf{2.85} & \cellcolor[HTML]{E6F0E8}\textbf{63.91} & \cellcolor[HTML]{E6F0E8}\textbf{0.6086}                \\
\midrule

& w/o $Q_s$  & \ding{55} & \ding{55} & \checkmark & \checkmark  & \ding{55} & \ding{55} & 0.2705        & 1.16     & 25.58     & 0.1633                                  \\
& w/o $T_{2}T_{3}$ & \checkmark  & \ding{55} & \checkmark  & \checkmark    & \ding{55}    & \ding{55}      & 0.2823 & 1.40      & 20.98     & 0.1993                                  \\
& w/o $T_{3}$   & \checkmark  & \checkmark & \checkmark  & \checkmark    & \checkmark    & \ding{55}        &  0.2905     & 1.54     & 24.71     & 0.2092 \\
& w/o r-VAE & \checkmark  & \checkmark & \ding{55} & \checkmark   & \checkmark      & \checkmark & 0.3024 & 2.05 & 26.96 & 0.2390 \\
\multirow{-5}{*}{RealBlur-J} & \cellcolor[HTML]{E6F0E8}\textbf{Ours} & \cellcolor[HTML]{E6F0E8}\checkmark & \cellcolor[HTML]{E6F0E8}\checkmark & \cellcolor[HTML]{E6F0E8}\checkmark & \cellcolor[HTML]{E6F0E8}\checkmark & \cellcolor[HTML]{E6F0E8}\checkmark & \cellcolor[HTML]{E6F0E8}\textbf{\checkmark}& \cellcolor[HTML]{E6F0E8}\textbf{0.3092} & \cellcolor[HTML]{E6F0E8}\textbf{2.11} & \cellcolor[HTML]{E6F0E8}\textbf{27.06} & \cellcolor[HTML]{E6F0E8}\textbf{0.2392}                \\
\midrule

& w/o $Q_s$  & \ding{55} & \ding{55} & \checkmark & \checkmark  & \ding{55} & \ding{55} & 0.3289        & 1.94     & 48.41     & 0.2925                                  \\
& w/o $T_{2}T_{3}$ & \checkmark  & \ding{55} & \checkmark  & \checkmark    & \ding{55}    & \ding{55}      & 0.3401    & 1.44     & 40.98     & 0.2392                                  \\
& w/o $T_{3}$   & \checkmark  & \checkmark & \checkmark  & \checkmark    & \checkmark    & \ding{55}        &  0.3485     & 2.11     & 47.19     & 0.4013 \\
& w/o r-VAE & \checkmark  & \checkmark & \ding{55} & \checkmark   & \checkmark      & \checkmark & 0.3574 & 2.73 & 57.11 & 0.5210 \\
\multirow{-5}{*}{RealBlur-R} & \cellcolor[HTML]{E6F0E8}\textbf{Ours} & \cellcolor[HTML]{E6F0E8}\checkmark & \cellcolor[HTML]{E6F0E8}\checkmark & \cellcolor[HTML]{E6F0E8}\checkmark & \cellcolor[HTML]{E6F0E8}\checkmark & \cellcolor[HTML]{E6F0E8}\checkmark & \cellcolor[HTML]{E6F0E8}\textbf{\checkmark}& \cellcolor[HTML]{E6F0E8}\textbf{0.3577} & \cellcolor[HTML]{E6F0E8}\textbf{2.81} & \cellcolor[HTML]{E6F0E8}\textbf{57.19} & \cellcolor[HTML]{E6F0E8}\textbf{0.5213}      \\
\bottomrule
\end{tabular}
}
\end{table}

In this section, we explore the effectiveness of each key component and training phase of UID-Diff. All ablation studies are conducted under the same settings as described in Section~\ref{sec:details}.

\textbf{Effectiveness of Reconstruction.} We perform ablation by removing the reconstruction task during the joint training phase, denoted as "w/o $T_3$". The quantitative results presented in Table~\ref{tab:No-Reference Metrics Results of Ablation Studies} and qualitative results shown in Figure~\ref{fig:visualization of structural} and Figure~\ref{fig:ablation} indicate that omitting the reconstruction task results in decline in both quantitative metrics and visual quality. This underscores the importance of the reconstruction task in enhancing the complementarity between structural features and blur patterns, thus improve the generalization ability of $Q_s$.

\textbf{Effectiveness of Blur-transfer.} We conduct ablation studies to evaluate the effectiveness of the blur-transfer task by removing the joint-training phase, denoted as "w/o $T_{2}T_{3}$". As illustrated in Table~\ref{tab:No-Reference Metrics Results of Ablation Studies}, Figure~\ref{fig:visualization of structural} and Figure~\ref{fig:ablation}, training the deblurring task solely on synthetic data is insufficient to handle unknown domain blurs, resulting in decline in both quantitative results and visual quality. This is because the structure extractor overfits synthetic blur patterns and fails to decouple unseen blur patterns in real-world scenarios effectively. We further validate the effectiveness of the blur pattern extractor $Q_b$ by visualizing the blur-transfer results in Appendix C.

\textbf{Effectiveness of the Structure Extractor.} We investigate the impact of the structure extractor $Q_s$ by removing it in the deblurring task, denoted as "w/o $Q_s$". As demonstrated in Table~\ref{tab:No-Reference Metrics Results of Ablation Studies}, the absence of $Q_s$ results in a performance decline across all datasets. Figure~\ref{fig:visualization of structural} and  Figure~\ref{fig:ablation} reveal the emergence of artifacts and distortions when $Q_s$ is removed, underscoring its importance in eliminating features unrelated to structural information to enhance the quality of the restored output. 

\textbf{Ablation on the refined-VAE.} We conduct an ablation study by replacing the refined-VAE with the original SD VAE, denoted as "w/o r-VAE". As shown in Table~\ref{tab:No-Reference Metrics Results of Ablation Studies}, using the original VAE results in a slight performance drop in no-reference metrics; however, UID-Diff with the original VAE still outperforms all baseline methods. The visual results presented in Figure~\ref{fig:ablation} indicate that UID-Diff with the refined-VAE effectively mitigates deviations from the original image.

\section{Conclusion}
In this work, we propose UID-Diff, a generative-diffusion-based model for unpaired image deblurring via blur-decoupled learning. By joint training on a supervised deblurring task, an unsupervised blur pattern learning task, and a reconstruction task, UID-Diff generalizes well to handle images from unknown domains that lack paired data. Experiments on widely-used real-world datasets demonstrate that UID-Diff outperforms existing SOTA methods in blur removal and structural preservation, showcasing its potential for application in diverse and challenging environments.


\clearpage

\appendix

\section*{Appendix}
The outline of the Appendix is as follows:
 \begin{itemize}
     \item Details of the refined-VAE in UID-Diff.
     \item Additional quantitative results.
     \item Additional qualitative results.
     \item Broader Impact.
 \end{itemize}

\section{Details of the Refined-VAE in UID-Diff}
\label{appendix: vaerefiner}

\begin{figure}[!t]
	\centering
	\includegraphics[width=0.65\textwidth]{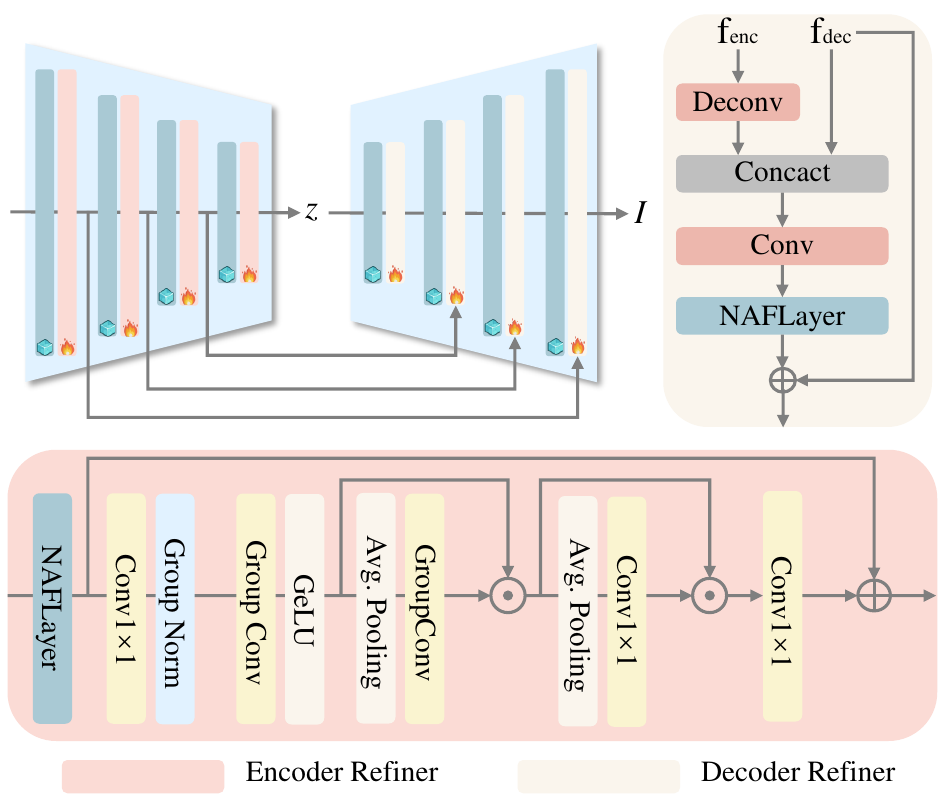}
	\caption{Architecture of the refined-VAE in UID-Diff.}
	\label{fig:refiner}
\end{figure}

In this section, we introduce the refined-VAE in UID-Diff. The pre-trained SD model utilizes a VAE to compress images into latent codes, reducing computational costs during diffusion. However, due to the compression inherent in the VAE and the randomness of the diffusion process, using the original SD VAE in deblurring tasks can lead to issues such as detail distortion and loss, resulting in discrepancies between the restored images and the input ones~\cite{zhu2023designing}. To mitigate this issue, we replace the VAE of SD with a refined-VAE, consistent with previous generative-diffusion-based methods~\cite{zhang2024diff,chen2025unirestore}. As illustrated in Figure~\ref{fig:refiner}, the refined-VAE $\mathcal{V'}$ introduces feature constraints by adding a refiner to both the encoder and decoder of the original VAE $\mathcal{V}$ to align the compressed latent codes and the input images. The encoder refiner performs pre-filtering operations on the degraded features after each layer of the original VAE encoder, thereby reducing content loss during compression. These refined features are then combined with the decoder-refiner via skip connections. The decoder refiner operates after each decoder layer of the original VAE, aiming to combine the features of the original image when decoding the latent codes refined by the diffusion process. This approach helps to minimize the discrepancies between the decoded image and the original one.

During training, we adapt the strategy from~\cite{chen2025unirestore}, which consists of two phases. In the first phase, we train the encoder refiner. The degraded image $y$ is encoded into latent codes using $\mathcal{V'}$, while the corresponding sharp image $x$ is encoded into ground truth latent codes using $\mathcal{V}$. The Mean Squared Error (MSE) loss is used for constraint. The optimization function is as follows:
\begin{equation}
\mathcal{L}_{\text{encoder}} = \text{MSE}(\mathcal{V}(x), \mathcal{V'}{\text{encoder}}(x)).
\end{equation}
In the second phase, we train the decoder refiner. We freeze the trained encoder refiner and utilize $\mathcal{V'}_{\text{encoder}}$ to encode $x$ into latent codes $z$ with intermediate features. The trained UID-Diff then processes $z$ to generate the final latent codes, which will be decoded into images using the refined VAE decoder $\mathcal{V'}_{\text{decoder}}$, considering the intermediate features. The generated image is compared with the corresponding sharp image $x$ using MSE loss. The optimization function is as follows:
\begin{equation}
\begin{aligned}
\mathcal{L}_{\text{decoder}} &= \text{MSE}(x, \mathcal{V'}{\text{decoder}}(\text{UID-Diff}(z))), \\
z &= \mathcal{V'}{\text{encoder}}(y).
\end{aligned}
\end{equation}
For fast convergence, the parameters of the original VAE are frozen during the training process. We use the same method as described in Section 4.2 to construct synthetic data for training the refiner to ensure fair comparison.

\begin{table}[!t]
  \caption{Ablation study results on reference-based metrics.}
  \label{tab:addition ablation}
  \centering
\resizebox{1\textwidth}{!}{
\begin{tabular}{cl|ccc|ccc|cccc}
\toprule
& \multicolumn{1}{c}{} & \multicolumn{3}{c}{Components} & \multicolumn{3}{c}{Training Tasks} & \multicolumn{4}{c}{Metrics} \\
\multirow{-2}{*}{Dataset}    & \multicolumn{1}{c|}{\multirow{-2}{*}{Model}} & \multicolumn{1}{c}{$Q_s$} & \multicolumn{1}{c}{$Q_b$} & \multicolumn{1}{c|}{r-VAE} & $T_{1}$ & $T_{2}$ & $T_{3}$ & FID$\downarrow$ & PSNR$\uparrow$ & SSIM$\uparrow$ & LPIPS$\downarrow$  \\
\midrule
& w/o $Q_s$  & \ding{55} & \ding{55} & \checkmark & \checkmark  & \ding{55} & \ding{55} & 50.52 & 22.67 & 0.6887 & 0.1689 \\
& w/o $T_{2}T_{3}$ & \checkmark  & \ding{55} & \checkmark  & \checkmark & \ding{55} & \ding{55}    & 54.53 & 22.91 & 0.7316 & 0.1541 \\
& w/o $T_{3}$   & \checkmark  & \checkmark & \checkmark  & \checkmark    & \checkmark    & \ding{55}  & 50.45 & 24.34 & 0.7231 & 0.1552  \\
& w/o r-VAE & \checkmark  & \checkmark & \ding{55} & \checkmark   & \checkmark  & \checkmark & 50.37 & 23.96 & 0.6984 & 0.1563 \\
\multirow{-5}{*}{GoPro}       & \cellcolor[HTML]{E6F0E8}\textbf{Ours}       & \cellcolor[HTML]{E6F0E8}\checkmark & \cellcolor[HTML]{E6F0E8}\checkmark & \cellcolor[HTML]{E6F0E8}\checkmark & \cellcolor[HTML]{E6F0E8}\checkmark & \cellcolor[HTML]{E6F0E8}\textbf{\checkmark} & \cellcolor[HTML]{E6F0E8}\textbf{\checkmark} & \cellcolor[HTML]{E6F0E8}\textbf{49.68} & \cellcolor[HTML]{E6F0E8}\textbf{25.08} & \cellcolor[HTML]{E6F0E8}\textbf{0.7403} & \cellcolor[HTML]{E6F0E8}\textbf{0.1310} \\

\midrule
& w/o $Q_s$  & \ding{55} & \ding{55} & \checkmark & \checkmark  & \ding{55} & \ding{55} & 26.86 & 22.03 & 0.6210 & 0.0783 \\
& w/o $T_{2}T_{3}$ & \checkmark  & \ding{55} & \checkmark  & \checkmark    & \ding{55}    & \ding{55}    &  27.22   & 23.53       & 0.6871 & 0.0782   \\
& w/o $T_{3}$   & \checkmark  & \checkmark & \checkmark  & \checkmark    & \checkmark    & \ding{55}  & 30.03& 23.60& 0.7137& 0.0813  \\
& w/o r-VAE & \checkmark  & \checkmark & \ding{55} & \checkmark   & \checkmark  & \checkmark  & 28.88  & 23.36  & 0.7223 & 0.1687\\
\multirow{-5}{*}{REDS}       & \cellcolor[HTML]{E6F0E8}\textbf{Ours}       & \cellcolor[HTML]{E6F0E8}\checkmark & \cellcolor[HTML]{E6F0E8}\checkmark & \cellcolor[HTML]{E6F0E8}\checkmark & \cellcolor[HTML]{E6F0E8}\checkmark & \cellcolor[HTML]{E6F0E8}\checkmark & \cellcolor[HTML]{E6F0E8}\textbf{\checkmark} & \cellcolor[HTML]{E6F0E8}\textbf{26.35} & \cellcolor[HTML]{E6F0E8}\textbf{25.83} & \cellcolor[HTML]{E6F0E8}\textbf{0.7782} & \cellcolor[HTML]{E6F0E8}\textbf{0.0682} \\
\midrule

& w/o $Q_s$  & \ding{55} & \ding{55} & \checkmark & \checkmark  & \ding{55} & \ding{55} & 38.92 & 18.93 & 0.5283 & 0.2090 \\
& w/o $T_{2}T_{3}$ & \checkmark  & \ding{55} & \checkmark  & \checkmark    & \ding{55}    & \ding{55}    & 38.98& 21.27& 0.7283 & 0.2257  \\
& w/o $T_{3}$   & \checkmark  & \checkmark & \checkmark  & \checkmark    & \checkmark    & \ding{55}  & 38.94& 21.33 & 0.7205 & 0.2170 \\
& w/o r-VAE & \checkmark  & \checkmark & \ding{55} & \checkmark   & \checkmark   & \checkmark & 38.78 & 20.63 & 0.7189 & 0.1544 \\
\multirow{-5}{*}{RealBlur-J} & \cellcolor[HTML]{E6F0E8}\textbf{Ours}       & \cellcolor[HTML]{E6F0E8}\checkmark & \cellcolor[HTML]{E6F0E8}\checkmark  & \cellcolor[HTML]{E6F0E8}\checkmark & \cellcolor[HTML]{E6F0E8}\checkmark & \cellcolor[HTML]{E6F0E8}\checkmark & \cellcolor[HTML]{E6F0E8}\textbf{\checkmark} & \cellcolor[HTML]{E6F0E8}\textbf{37.99} & \cellcolor[HTML]{E6F0E8}\textbf{22.76} & \cellcolor[HTML]{E6F0E8}\textbf{0.7293} & \cellcolor[HTML]{E6F0E8}\textbf{0.1379}  \\
\midrule

& w/o $Q_s$  & \ding{55} & \ding{55} & \checkmark & \checkmark  & \ding{55} & \ding{55} & 37.44 & 16.21 & 0.5213 & 0.2501 \\
& w/o $T_{2}T_{3}$ & \checkmark  & \ding{55} & \checkmark  & \checkmark    & \ding{55}    & \ding{55} &37.08& 17.27 & 0.5283 & 0.2457 \\
& w/o $T_{3}$   & \checkmark  & \checkmark & \checkmark  & \checkmark    & \checkmark    & \ding{55}  & 34.49& 19.74&0.5345&0.2458\\
& w/o r-VAE & \checkmark  & \checkmark & \ding{55} & \checkmark   & \checkmark   & \checkmark  & 33.74 & 21.84 & 0.5104 & 0.2563 \\
\multirow{-5}{*}{RealBlur-R} & \cellcolor[HTML]{E6F0E8}\textbf{Ours}       & \cellcolor[HTML]{E6F0E8}\checkmark & \cellcolor[HTML]{E6F0E8}\checkmark & \cellcolor[HTML]{E6F0E8}\checkmark & \cellcolor[HTML]{E6F0E8}\checkmark & \cellcolor[HTML]{E6F0E8}\checkmark & \cellcolor[HTML]{E6F0E8}\textbf{\checkmark} & \cellcolor[HTML]{E6F0E8}\textbf{32.11} & \cellcolor[HTML]{E6F0E8}\textbf{22.47} & \cellcolor[HTML]{E6F0E8}\textbf{0.5387}  & \cellcolor[HTML]{E6F0E8}\textbf{0.2348} \\
\bottomrule
\end{tabular}
}
\end{table}

\section{Additional Quantitative Results}
\label{appendix: additional metrics}
In this section, we present additional quantitative results on reference-based metrics: FID, PSNR, SSIM, and LPIPS~\cite{zhang2018unreasonable}. All experimental settings are aligned with Section 4.

\textbf{Comparison with SOTA Methods.} The performance comparisons with selected baselines on reference-based metrics are presented in Table~\ref{tab:addition ablation}. UID-Diff performs favorably against other generative-diffusion-based methods across all selected datasets. Notably, due to the generative nature of T2I diffusion models and the randomness of the diffusion process, generative-diffusion-based methods tend to supplement details in blurred regions at the expense of consistency with the original image. As a result, they generally underperform on reference-based metrics compared to non-diffusion-based models. However, the decoupled learning strategy of UID-Diff allows it to effectively extract structural information from blurred images and restore structures that are faithful to the original image, thereby maintaining competitiveness with other SOTA methods.
 
\textbf{Ablation Study.} The results of the ablation study on reference-based metrics are presented in Table~\ref{tab:addition ablation}. Omitting any key component or training phase from UID-Diff leads to a decline across all reference-based metrics, which aligns with the performance trend observed in the NR-IQA metrics shown in Table 3 in the main paper. This further demonstrates the effectiveness of UID-Diff.

\section{Additional Qualitative Results}

\begin{figure}[!b]
  \centering
  \setlength{\tabcolsep}{3pt} 
  \begin{tabular}{cccccc}
    & \small Sample 1 &  \small Sample 2 &\small Sample 3 &\small Sample 4 &\small Sample 5 \\
     \raisebox{3\height}{\rotatebox[origin=c]{90}{\small Blur}} &
    \includegraphics[width=.18\textwidth]{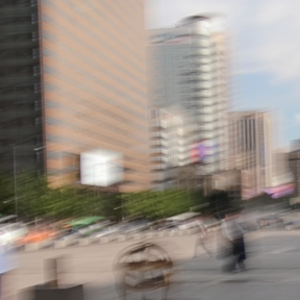} &
    \includegraphics[width=.18\textwidth]{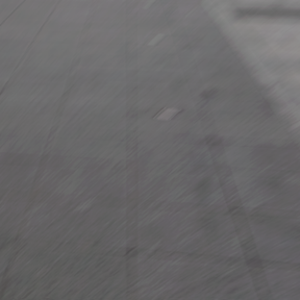} &
    \includegraphics[width=.18\textwidth]{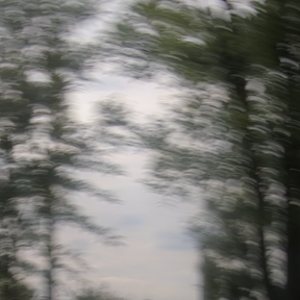} &
    \includegraphics[width=.18\textwidth]{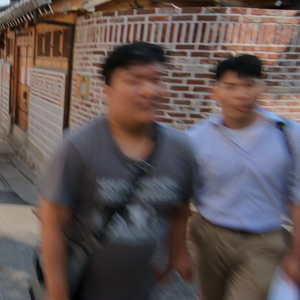} &
    \includegraphics[width=.18\textwidth]{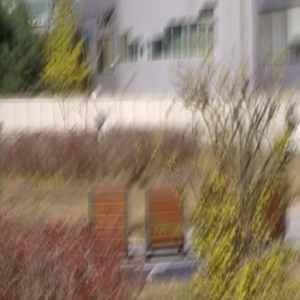} \\
    \multicolumn{4}{c}{\vspace{-9pt}} \\
    \raisebox{3\height}{\rotatebox[origin=c]{90}{\small Sharp}} &
    \includegraphics[width=.18\textwidth]{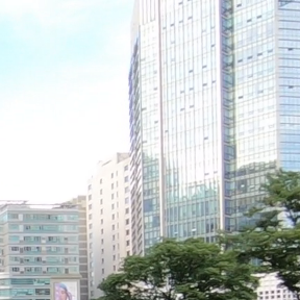} &
    \includegraphics[width=.18\textwidth]{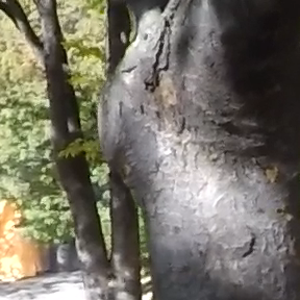} &
    \includegraphics[width=.18\textwidth]{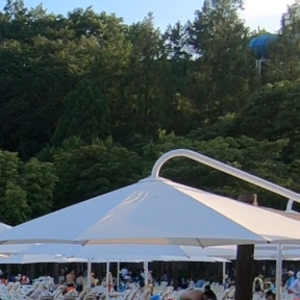} &
    \includegraphics[width=.18\textwidth]{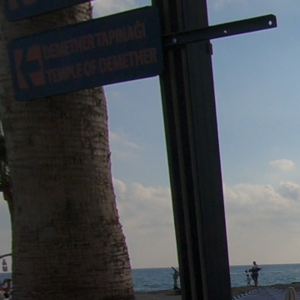} &
    \includegraphics[width=.18\textwidth]{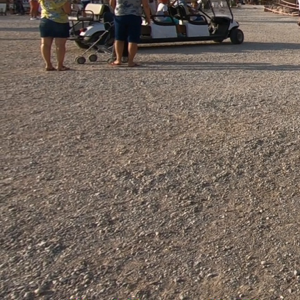} \\
    \multicolumn{4}{c}{\vspace{-9pt}} \\
    \raisebox{1\height}{\rotatebox[origin=c]{90}{\small Blur-transfered}} &
    \includegraphics[width=.18\textwidth]{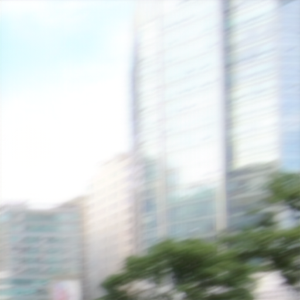} &
    \includegraphics[width=.18\textwidth]{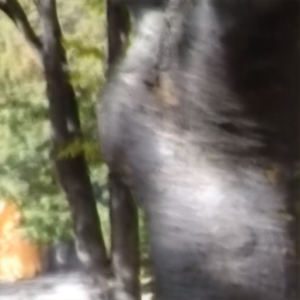} &
    \includegraphics[width=.18\textwidth]{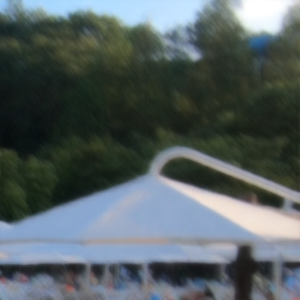} &
    \includegraphics[width=.18\textwidth]{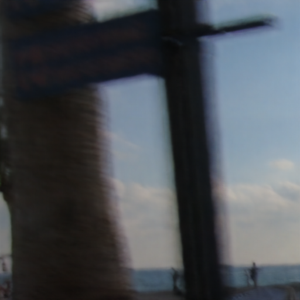} &
    \includegraphics[width=.18\textwidth]{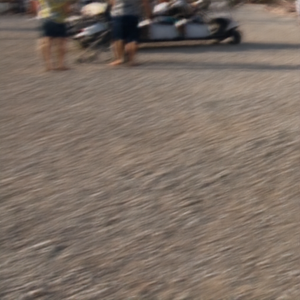} \\
  \end{tabular}
\caption{Visualization of blur transferring. The blur pattern in the images of the first row is transferred to the images in the second row to generate the blurred output in the third row.}
  \label{fig:blur_transfered}
\end{figure}

\begin{figure}[!b]
  \centering
  \setlength{\tabcolsep}{3pt} 
  \begin{tabular}{cccccc}
    & \small Sample 1 &  \small Sample 2 &\small Sample 3 &\small Sample 4 &\small Sample 5 \\
     \raisebox{3\height}{\rotatebox[origin=c]{90}{\small Blur}} &
    \includegraphics[width=.18\textwidth]{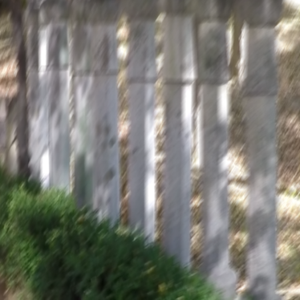} &
    \includegraphics[width=.18\textwidth]{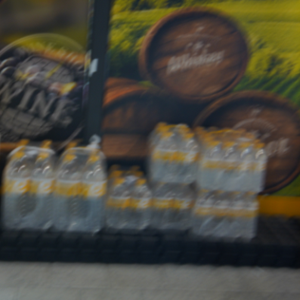} &
    \includegraphics[width=.18\textwidth]{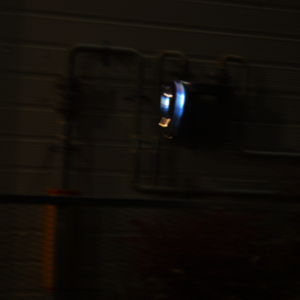} &
    \includegraphics[width=.18\textwidth]{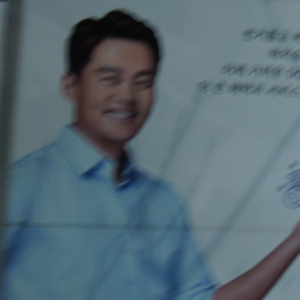} &
    \includegraphics[width=.18\textwidth]{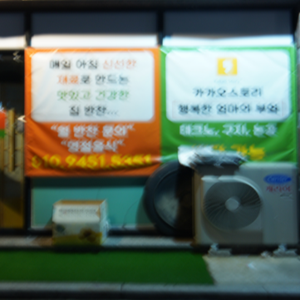} \\
    \multicolumn{4}{c}{\vspace{-9pt}} \\
    \raisebox{3\height}{\rotatebox[origin=c]{90}{\small Sharp}} &
    \includegraphics[width=.18\textwidth]{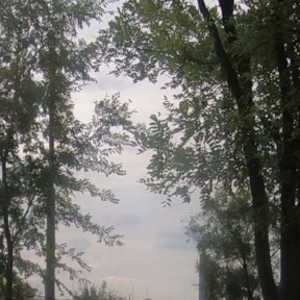} &
    \includegraphics[width=.18\textwidth]{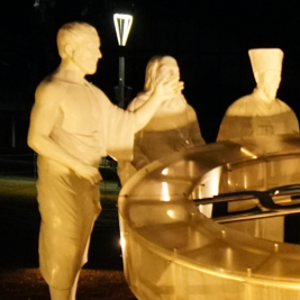} &
    \includegraphics[width=.18\textwidth]{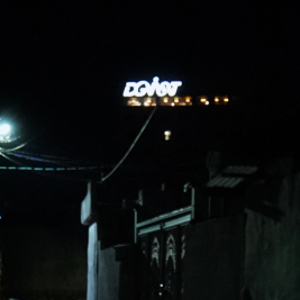} &
    \includegraphics[width=.18\textwidth]{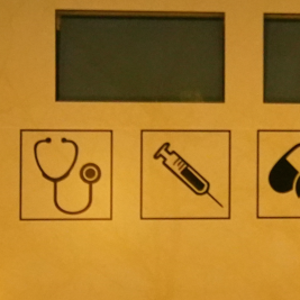} &
    \includegraphics[width=.18\textwidth]{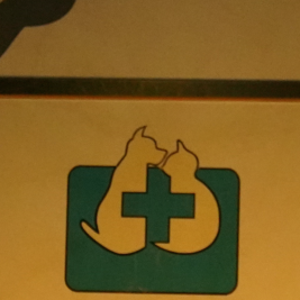} \\
    \multicolumn{4}{c}{\vspace{-9pt}} \\
    \raisebox{1\height}{\rotatebox[origin=c]{90}{\small Blur-transfered}} &
    \includegraphics[width=.18\textwidth]{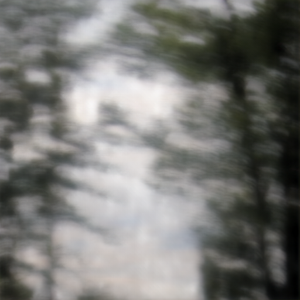} &
    \includegraphics[width=.18\textwidth]{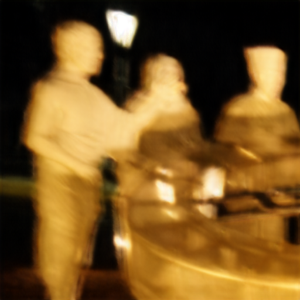} &
    \includegraphics[width=.18\textwidth]{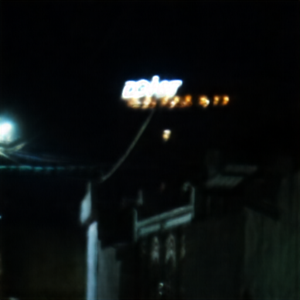} &
    \includegraphics[width=.18\textwidth]{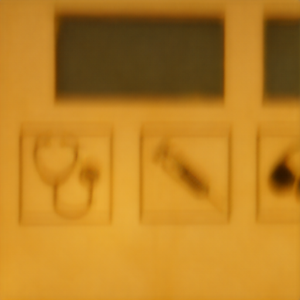} &
    \includegraphics[width=.18\textwidth]{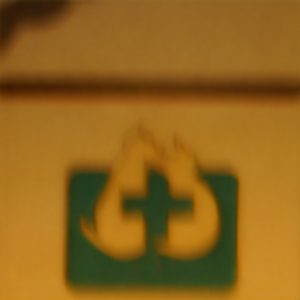} \\
  \end{tabular}
\caption{Visualization of blur transferring. The blur pattern in the images of the first row is transferred to the images in the second row to generate the blurred output in the third row.}
  \label{fig:blur_transfered2}
\end{figure}

\textbf{Blur-transfer Visualization.} In addition to the quantitative results that demonstrate the effectiveness of the blur-transfer task, we further evaluate the performance of the blur pattern extractor $Q_b$ by visualizing the blur-transfer outputs. We randomly select a sharp image $a$ and a blurred image $b$ from the test sets. We encode $b$ into a latent code and perform diffusion sampling conditioned on the extracted blur pattern of $a$. The results in Figure~\ref{fig:blur_transfered} and Figure~\ref{fig:blur_transfered2} demonstrate that $Q_b$ successfully captures and transfers the blur patterns from $a$ to $b$. Although the blur-transfer result $c$ does not achieve pixel-level correspondence with $b$ due to the randomness of the diffusion process, $c$ remains independent of the structure of $a$. This indicates that $Q_b$ has been effectively trained to extract only the blur pattern. On the other hand, $Q_s$ can focus on extracting the complementary structural features.

\begin{figure*}[!t]
  \centering
  \setlength{\tabcolsep}{5pt} 
  \begin{tabular}{ccccc}
    &   Sample 1 &   Sample 2 &   Sample 3 &   Sample 4 \\
    \raisebox{0.9\height}{\rotatebox[origin=c]{90}{ Ground Truth}} &
    \includegraphics[width=.18\textwidth]{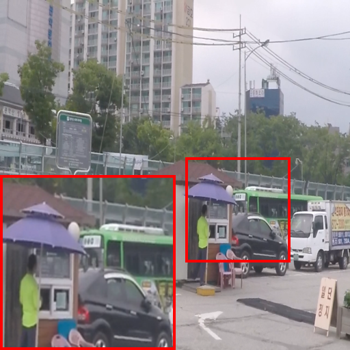} & 
    \includegraphics[width=.18\textwidth]{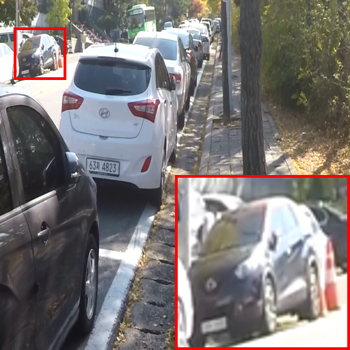} & 
    \includegraphics[width=.18\textwidth]{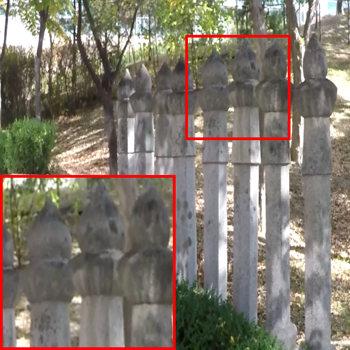} & 
    \includegraphics[width=.18\textwidth]{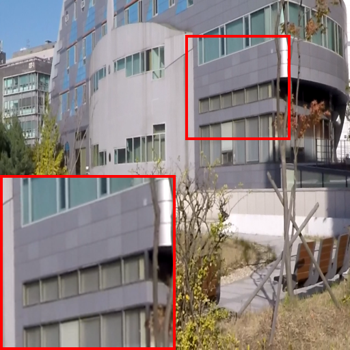}\\
   \multicolumn{5}{c}{\vspace{-6pt}} \\
    \raisebox{1.55\height}{\rotatebox[origin=c]{90}{ AirNet}} &
    \includegraphics[width=.18\textwidth]{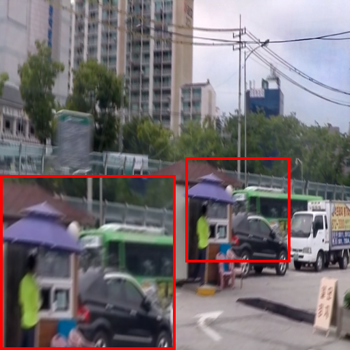} & 
    \includegraphics[width=.18\textwidth]{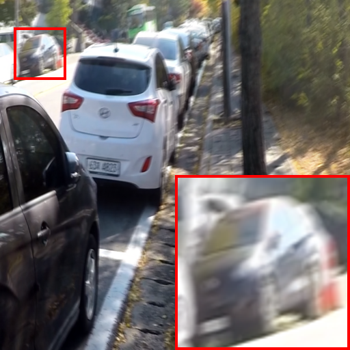} & 
    \includegraphics[width=.18\textwidth]{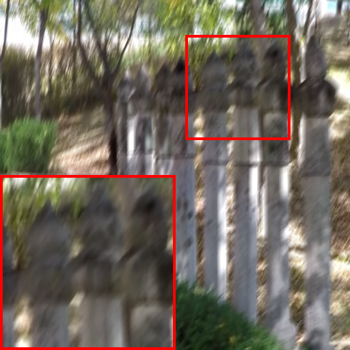} & 
    \includegraphics[width=.18\textwidth]{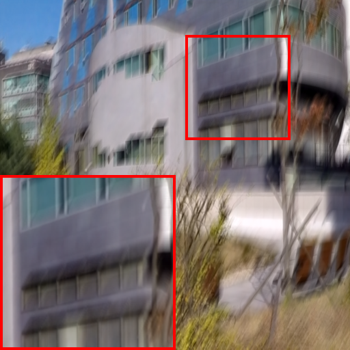}\\
   \multicolumn{5}{c}{\vspace{-6pt}} \\
    \raisebox{1.3\height}{\rotatebox[origin=c]{90}{ PromptIR}} &
    \includegraphics[width=.18\textwidth]{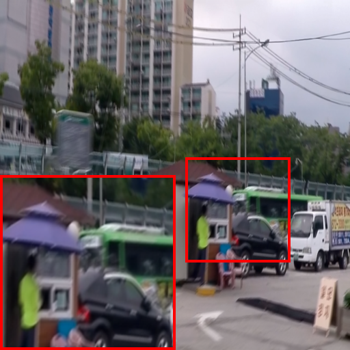} & 
    \includegraphics[width=.18\textwidth]{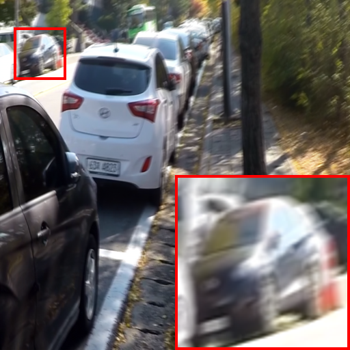} & 
    \includegraphics[width=.18\textwidth]{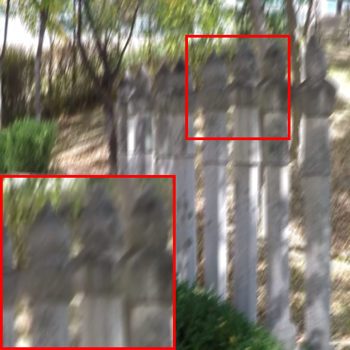} & 
    \includegraphics[width=.18\textwidth]{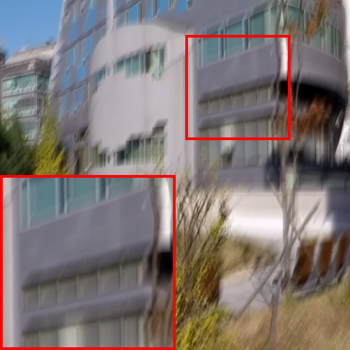}\\
   \multicolumn{5}{c}{\vspace{-6pt}} \\
    \raisebox{1.1\height}{\rotatebox[origin=c]{90}{ FFTformer}} &
    \includegraphics[width=.18\textwidth]{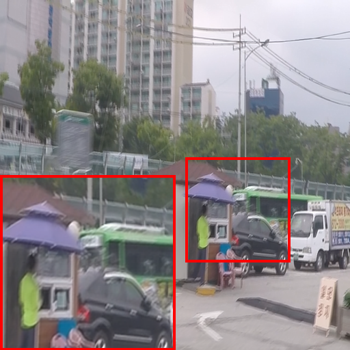} & 
    \includegraphics[width=.18\textwidth]{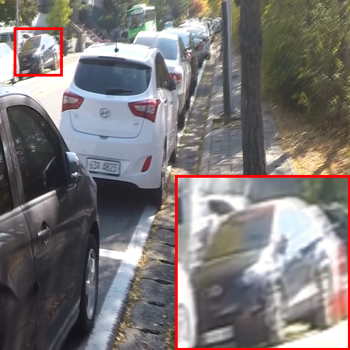} & 
    \includegraphics[width=.18\textwidth]{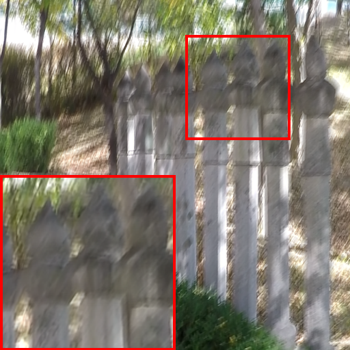} & 
    \includegraphics[width=.18\textwidth]{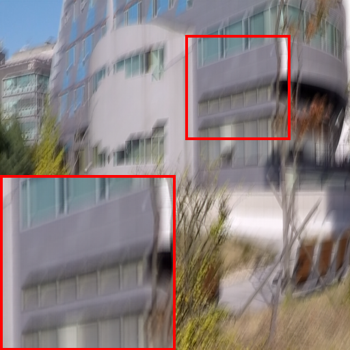}\\
   \multicolumn{5}{c}{\vspace{-6pt}} \\
    \raisebox{1.3\height}{\rotatebox[origin=c]{90}{ AdaRevD}} &
    \includegraphics[width=.18\textwidth]{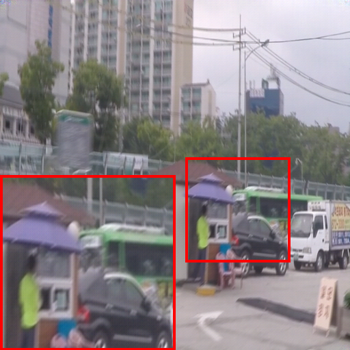} & 
    \includegraphics[width=.18\textwidth]{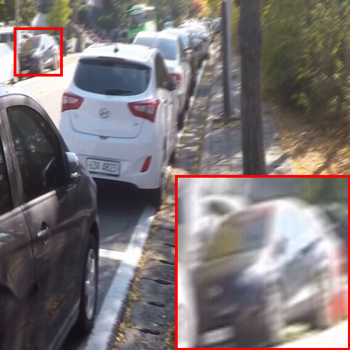} & 
    \includegraphics[width=.18\textwidth]{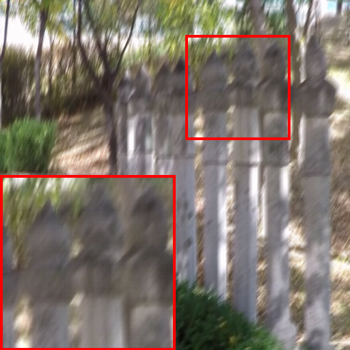} & 
    \includegraphics[width=.18\textwidth]{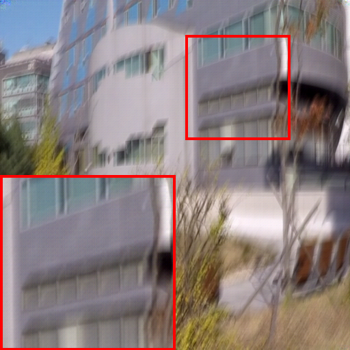}\\
   \multicolumn{5}{c}{\vspace{-6pt}} \\
    \raisebox{1.5\height}{\rotatebox[origin=c]{90}{ HI-diff}} &
    \includegraphics[width=.18\textwidth]{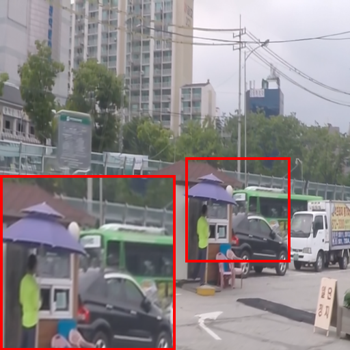} & 
    \includegraphics[width=.18\textwidth]{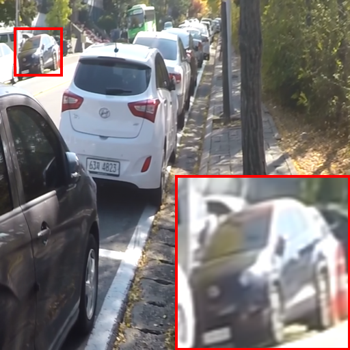} & 
    \includegraphics[width=.18\textwidth]{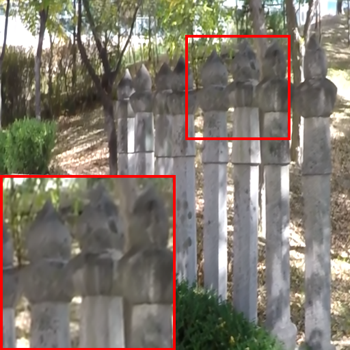} & 
    \includegraphics[width=.18\textwidth]{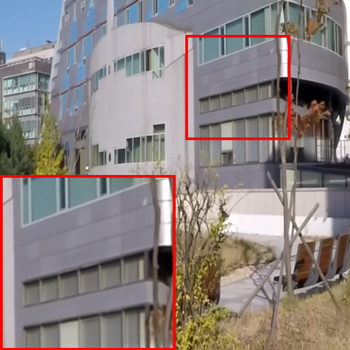}\\
   \multicolumn{5}{c}{\vspace{-6pt}} \\
    \raisebox{1.1\height}{\rotatebox[origin=c]{90}{ Blur2Blur}} &
    \includegraphics[width=.18\textwidth]{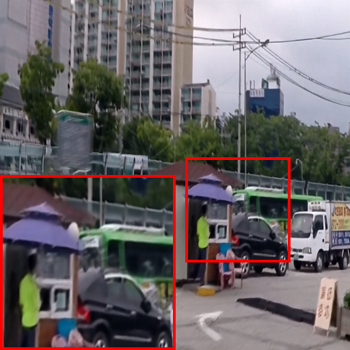} & 
    \includegraphics[width=.18\textwidth]{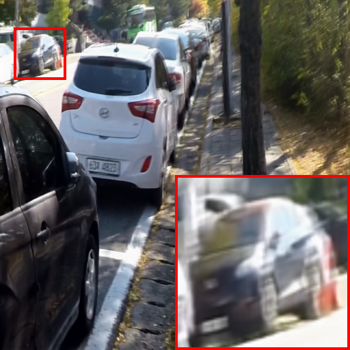} & 
    \includegraphics[width=.18\textwidth]{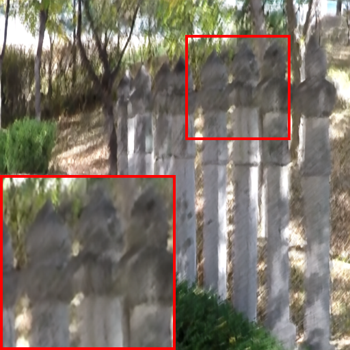} & 
    \includegraphics[width=.18\textwidth]{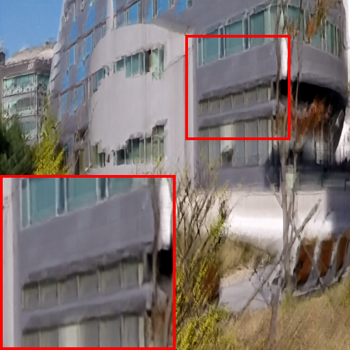} \\
 \multicolumn{5}{c}{\vspace{-6pt}} \\
  \end{tabular}
  \caption{Additional visual comparison on the \textcolor{red}{GoPro} dataset.}
  \label{fig:appendix resultsG1}
\end{figure*}

\begin{figure*}[!t]
  \centering
  \setlength{\tabcolsep}{5pt} 
  \begin{tabular}{ccccc}
    &   Sample 1 &   Sample 2 &   Sample 3 &   Sample 4 \\
    \raisebox{1.4\height}{\rotatebox[origin=c]{90}{ DiffBir}} &
    \includegraphics[width=.18\textwidth]{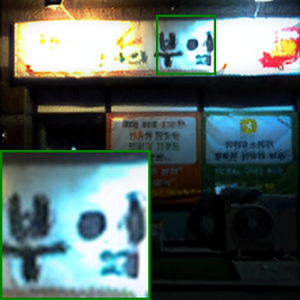} & 
    \includegraphics[width=.18\textwidth]{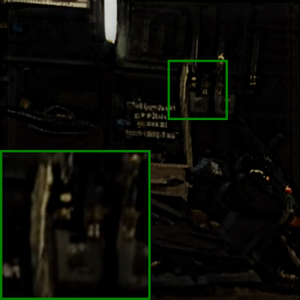} & 
    \includegraphics[width=.18\textwidth]{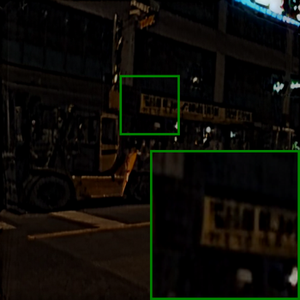} & 
    \includegraphics[width=.18\textwidth]{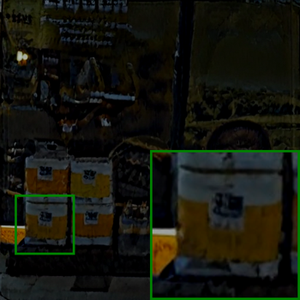} \\
   \multicolumn{5}{c}{\vspace{-6pt}} \\
    \raisebox{1.1\height}{\rotatebox[origin=c]{90}{ Diff-Plugin}} &
    \includegraphics[width=.18\textwidth]{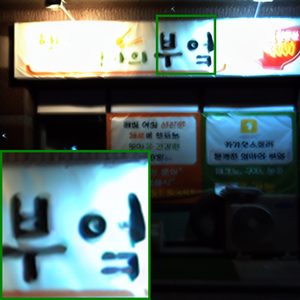} & 
    \includegraphics[width=.18\textwidth]{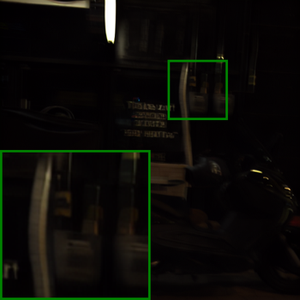} & 
    \includegraphics[width=.18\textwidth]{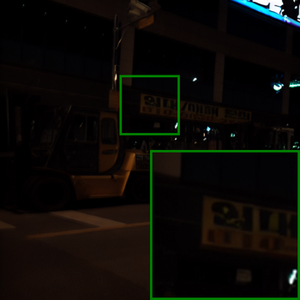} & 
    \includegraphics[width=.18\textwidth]{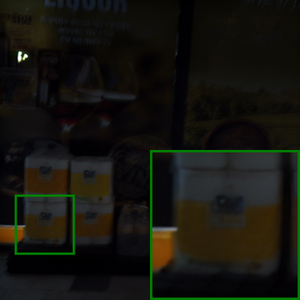}\\
   \multicolumn{5}{c}{\vspace{-6pt}} \\
    \raisebox{1.2\height}{\rotatebox[origin=c]{90}{ DA-CLIP}} &
    \includegraphics[width=.18\textwidth]{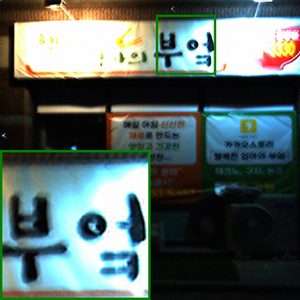} & 
    \includegraphics[width=.18\textwidth]{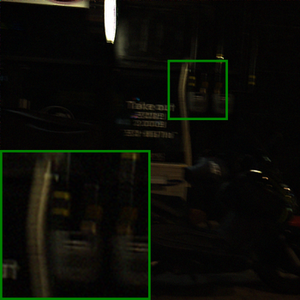} & 
    \includegraphics[width=.18\textwidth]{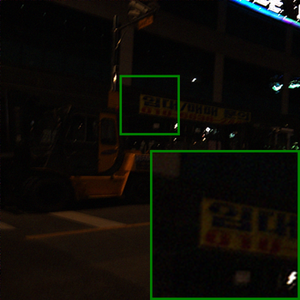} & 
    \includegraphics[width=.18\textwidth]{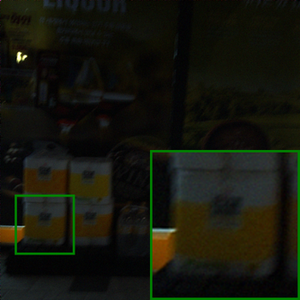}\\
   \multicolumn{5}{c}{\vspace{-6pt}} \\
    \raisebox{1\height}{\rotatebox[origin=c]{90}{ Ours w/o $Q_s$}} &
    \includegraphics[width=.18\textwidth]{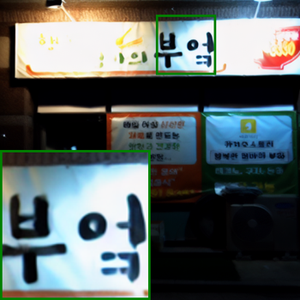} & 
    \includegraphics[width=.18\textwidth]{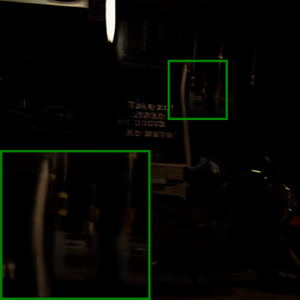} & 
    \includegraphics[width=.18\textwidth]{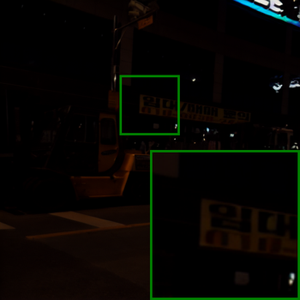} & 
    \includegraphics[width=.18\textwidth]{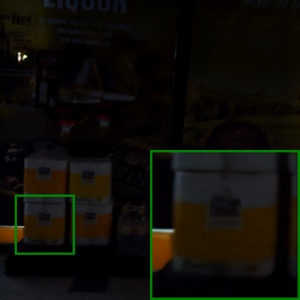} \\
   \multicolumn{5}{c}{\vspace{-6pt}} \\
    \raisebox{0.9\height}{\rotatebox[origin=c]{90}{ Ours w/o $T_{2}T_{3}$}} &
    \includegraphics[width=.18\textwidth]{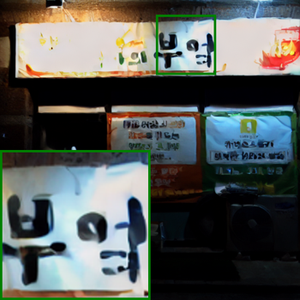} & 
    \includegraphics[width=.18\textwidth]{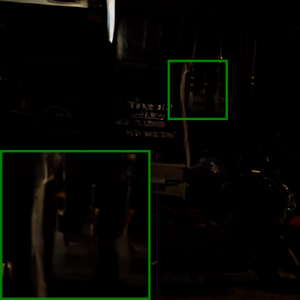} & 
    \includegraphics[width=.18\textwidth]{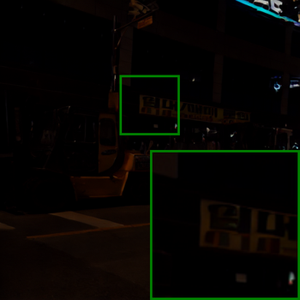} & 
    \includegraphics[width=.18\textwidth]{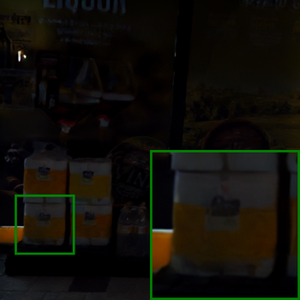}\\
   \multicolumn{5}{c}{\vspace{-6pt}} \\
    \raisebox{1.1\height}{\rotatebox[origin=c]{90}{ Ours w/o $T_{3}$}} &
    \includegraphics[width=.18\textwidth]{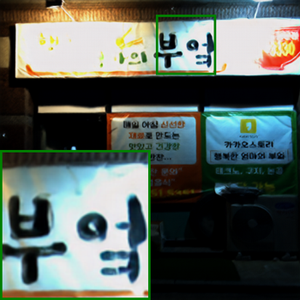} & 
    \includegraphics[width=.18\textwidth]{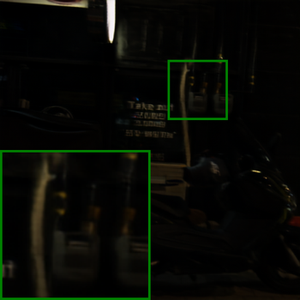} & 
    \includegraphics[width=.18\textwidth]{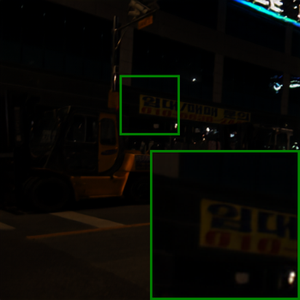} & 
    \includegraphics[width=.18\textwidth]{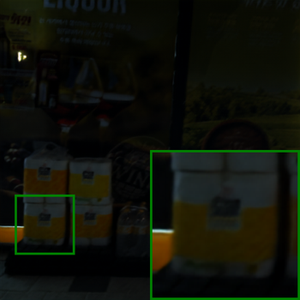} \\
   \multicolumn{5}{c}{\vspace{-6pt}} \\
    \raisebox{1.7\height}{\rotatebox[origin=c]{90}{ \textbf{Ours}}} &
    \includegraphics[width=.18\textwidth]{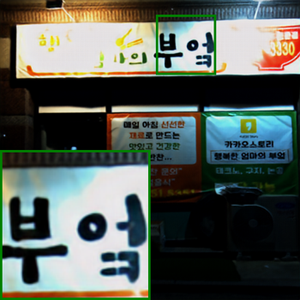} & 
    \includegraphics[width=.18\textwidth]{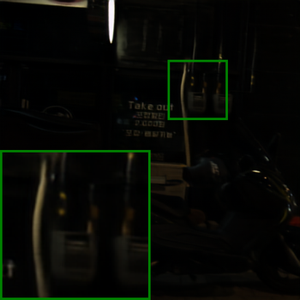} & 
    \includegraphics[width=.18\textwidth]{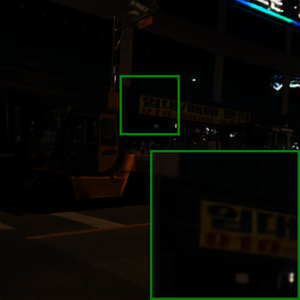} & 
    \includegraphics[width=.18\textwidth]{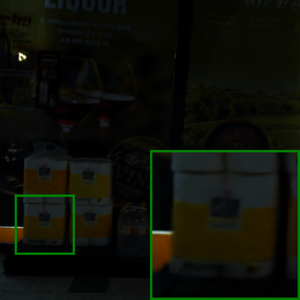} \\
   \multicolumn{5}{c}{\vspace{-6pt}} \\
    \raisebox{2.2\height}{\rotatebox[origin=c]{90}{ Blur}} &
    \includegraphics[width=.18\textwidth]{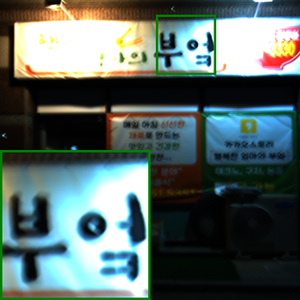} & 
    \includegraphics[width=.18\textwidth]{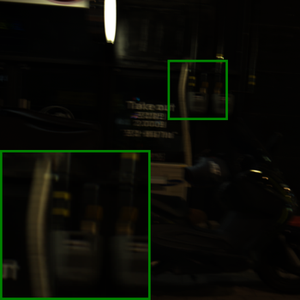} & 
    \includegraphics[width=.18\textwidth]{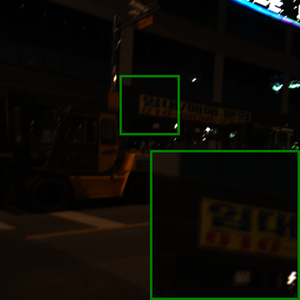} & 
    \includegraphics[width=.18\textwidth]{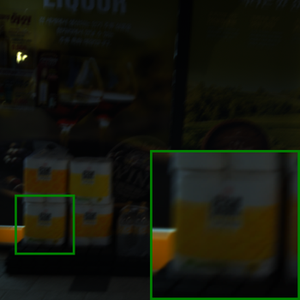} \\
   \multicolumn{5}{c}{\vspace{-6pt}} \\
  \end{tabular}
  \caption{Additional visual comparison on the \textcolor{green}{RealBlur-R} dataset (Continue).}
  \label{fig:appendix resultsR2}
\end{figure*}

\textbf{Debluring Visualization.}
In this section, we present additional visualization results on the test sets of our selected datasets: GoPro, REDS, RealBlur-J \& R. As shown in Figure~\ref{fig:appendix resultsG1} and Figure~\ref{fig:appendix resultsR2}, UID-Diff performs favorably against other SOTA methods across various real-world deblurring scenarios.

\section{Broader Impact}

The proposed UID-Diff contributes to advancing the field of image deblurring by introducing generative-diffusion-model into unpaired deblurring task. Its ability to generalize across diverse domains has potential applications in areas that lack paired data for supervised training.

However, the deployment of generative diffusion models in image deblurring also raises certain ethical considerations. The ability to manipulate images could be misused for deceptive purposes, such as altering visual evidence or creating misleading content. Researchers and practitioners should evaluate the societal implications and ensure that their deployment aligns with ethical guidelines.

\begin{table}[!h]
  \caption{Quantitative comparison with SOTA methods on reference-based metrics. \textbf{Bold} and \uline{underlined} indicate the best and the second-best performance, respectively.}
  \label{tab:addition results}
  \centering
\resizebox{\linewidth}{!}{
\begin{tabular}{c|lcccccc|cccc}
\toprule
Data & Metrics & AirNet &PromptIR &FFTformer &AdaRevD &HI-Diff &Blur2Blur &Diff-Plugin &DiffBir &DA-CLIP &\cellcolor[HTML]{E6F0E8}\textbf{Ours} \\
\midrule \midrule
\multirow{4}{*}{GoPro}
& FID$\downarrow$ & 56.09 & 49.65 & 48.74 & 31.92 & 59.32  & 46.31 & \uline{50.55} & 58.45 & 54.79 & \cellcolor[HTML]{E6F0E8}\textbf{49.68} \\
& PSNR$\uparrow$ & 21.82 & 21.97 &  22.21 & 26.93 &  23.46  & 24.33 & 22.87 & 23.42 & \uline{24.24} & \cellcolor[HTML]{E6F0E8}\textbf{25.08} \\
& SSIM$\uparrow$ & 0.7348 & 0.7361 & 0.6843 & 0.7492 & 0.7304  & 0.7392 & 0.6902 &\uline{0.7043} & 0.6442 & \cellcolor[HTML]{E6F0E8}\textbf{0.7403} \\
& LPIPS$\downarrow$ & 0.1446 & 0.1390 & 0.1187 & 0.0893 & 0.1293 & 0.1317 & 0.1388 & 0.1382 & \textbf{0.1299} & \cellcolor[HTML]{E6F0E8}\uline{0.1310} \\
\midrule
\multirow{4}{*}{REDS}
& FID$\downarrow$ & 33.36 & 16.57 & 12.84 & 14.74 & 13.75 & 15.49 & \uline{26.80} & 37.19 & 62.14 & \cellcolor[HTML]{E6F0E8}\textbf{15.35} \\
& PSNR$\uparrow$ & 23.94 & 27.03 & 27.53 & 28.78 & 24.77 & 28.94 & 21.95 & \uline{22.67} & 21.43 & \cellcolor[HTML]{E6F0E8}\textbf{25.48} \\
& SSIM$\uparrow$ & 0.7419 & 0.8149 & 0.8320 & 0.8623 & 0.8132  & 0.8522 & \uline{0.6215} & 0.5714 & 0.5953 & \cellcolor[HTML]{E6F0E8}\textbf{0.7822} \\
& LPIPS$\downarrow$  & 0.0975 & 0.0504 & 0.0498  & 0.0231 & 0.0510 & 0.0431 & \uline{0.0773} & 0.0817 & 0.1402 & \cellcolor[HTML]{E6F0E8}\textbf{0.0498} \\
\midrule
\multirow{4}{*}{RealBlur-J}
& FID$\downarrow$  & 38.52 & 17.02 & 15.43 & 16.53 & 16.23 & 16.64 & \uline{38.68} & 48.29 & 52.61 & \cellcolor[HTML]{E6F0E8}\textbf{17.99} \\
& PSNR$\uparrow$ & 22.51 & 22.51 & 25.52 & 27.95 & 21.53 & 22.71 & \uline{22.61} & \textbf{22.76} & 22.35 & \cellcolor[HTML]{E6F0E8}\textbf{22.76} \\
& SSIM$\uparrow$ & 0.7347 & 0.7598 & 0.7622 & 0.8531 & 0.7523 & 0.7642 & 0.7035 &\uline{0.7220} & 0.7129 & \cellcolor[HTML]{E6F0E8}\textbf{0.7693} \\
& LPIPS$\downarrow$ & 0.1483 & 0.1382 & 0.1135 & 0.1352 & 0.1346 & 0.1314 & 0.1568 & \uline{0.1471} & 0.1794 & \cellcolor[HTML]{E6F0E8}\textbf{0.1379} \\
\midrule
\multirow{4}{*}{RealBlur-R}
& FID$\downarrow$ & 40.93 & 36.82 & 35.94 & 40.35 & 28.87  & 21.43 & \uline{35.91} & 38.33 & 92.61 & \cellcolor[HTML]{E6F0E8}\textbf{32.11} \\
& PSNR$\uparrow$  & 16.82 & 17.03 & 26.94 & 27.85 & 21.02 & 19.82 & 16.56 & 17.24 & \uline{21.35} & \cellcolor[HTML]{E6F0E8}\textbf{22.47}  \\
& SSIM$\uparrow$ & 0.5093 & 0.5290 & 0.7542 & 0.8034 & 0.7099 & 0.5432 & 0.5206 & 0.4883 & \uline{0.7129} & \cellcolor[HTML]{E6F0E8}\textbf{0.7384} \\
& LPIPS$\downarrow$ & 0.2295 & 0.2011 & 0.1790 & 0.1872 & 0.2093 & 0.2231 & \uline{0.2147} & \textbf{0.2107} & 0.2794 & \cellcolor[HTML]{E6F0E8}0.2348 \\
\bottomrule
\end{tabular}}
\end{table}
\clearpage

\clearpage

\begin{figure*}[!t]
  \centering
  \setlength{\tabcolsep}{5pt} 
  \begin{tabular}{ccccc}
    &   Sample 1 &   Sample 2 &   Sample 3 &   Sample 4 \\
    \raisebox{1.4\height}{\rotatebox[origin=c]{90}{ DiffBir}} &
    \includegraphics[width=.18\textwidth]{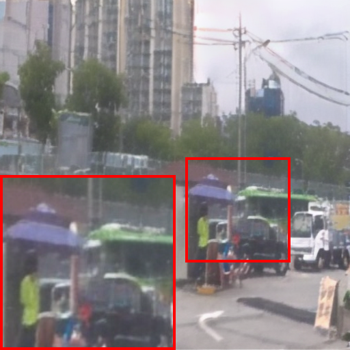} & 
    \includegraphics[width=.18\textwidth]{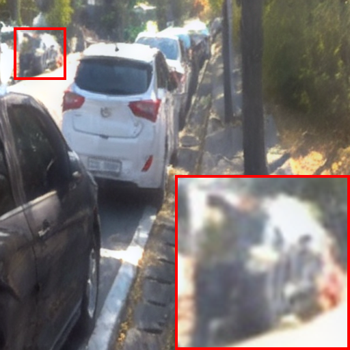} & 
    \includegraphics[width=.18\textwidth]{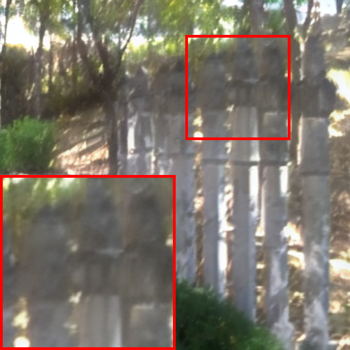} & 
    \includegraphics[width=.18\textwidth]{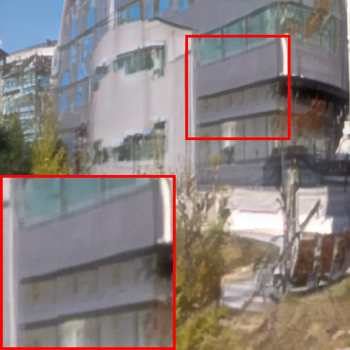} \\
   \multicolumn{5}{c}{\vspace{-6pt}} \\
    \raisebox{1.1\height}{\rotatebox[origin=c]{90}{ Diff-Plugin}} &
    \includegraphics[width=.18\textwidth]{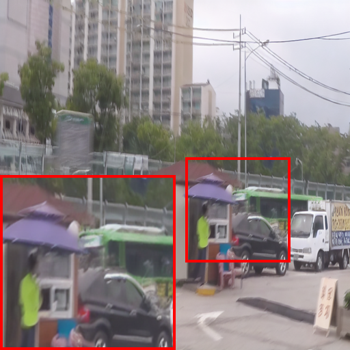} & 
    \includegraphics[width=.18\textwidth]{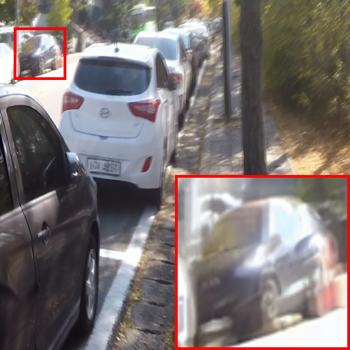} & 
    \includegraphics[width=.18\textwidth]{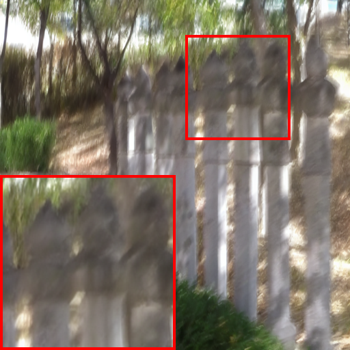} & 
    \includegraphics[width=.18\textwidth]{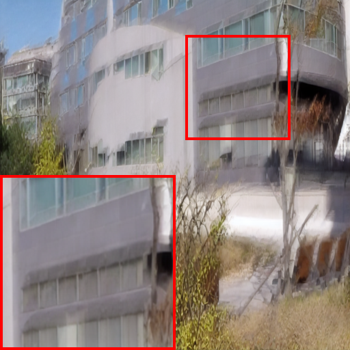}\\
   \multicolumn{5}{c}{\vspace{-6pt}} \\
    \raisebox{1.2\height}{\rotatebox[origin=c]{90}{ DA-CLIP}} &
    \includegraphics[width=.18\textwidth]{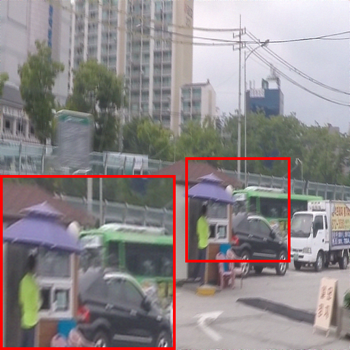} & 
    \includegraphics[width=.18\textwidth]{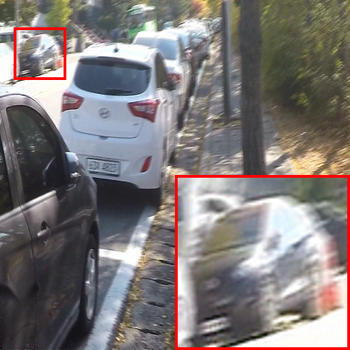} & 
    \includegraphics[width=.18\textwidth]{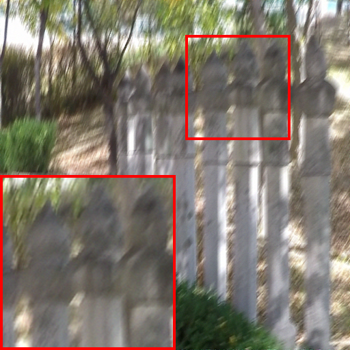} & 
    \includegraphics[width=.18\textwidth]{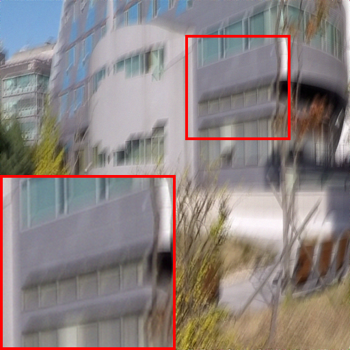}\\
   \multicolumn{5}{c}{\vspace{-6pt}} \\
    \raisebox{1\height}{\rotatebox[origin=c]{90}{ Ours w/o $Q_s$}} &
    \includegraphics[width=.18\textwidth]{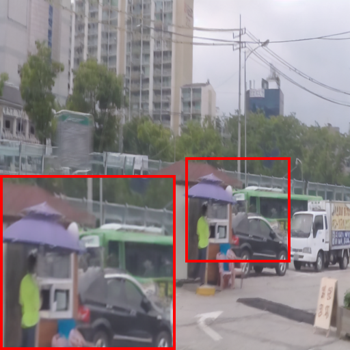} & 
    \includegraphics[width=.18\textwidth]{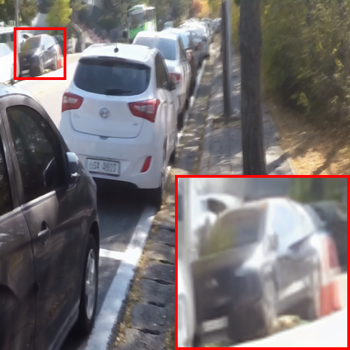} & 
    \includegraphics[width=.18\textwidth]{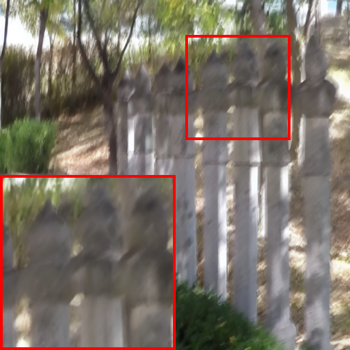} & 
    \includegraphics[width=.18\textwidth]{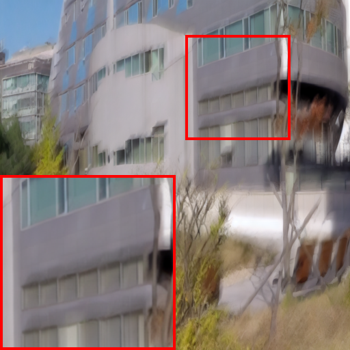} \\
   \multicolumn{5}{c}{\vspace{-6pt}} \\
    \raisebox{0.9\height}{\rotatebox[origin=c]{90}{ Ours w/o $T_{2}T_{3}$}} &
    \includegraphics[width=.18\textwidth]{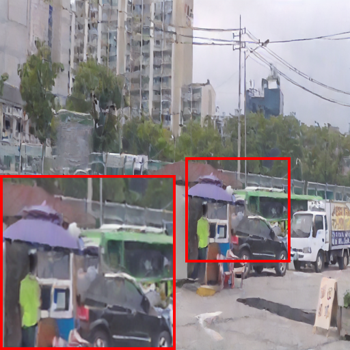} & 
    \includegraphics[width=.18\textwidth]{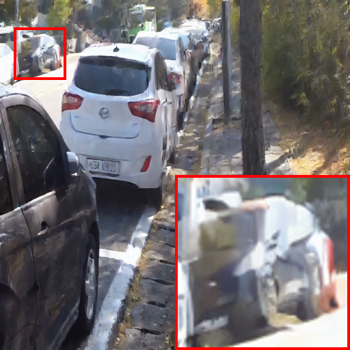} & 
    \includegraphics[width=.18\textwidth]{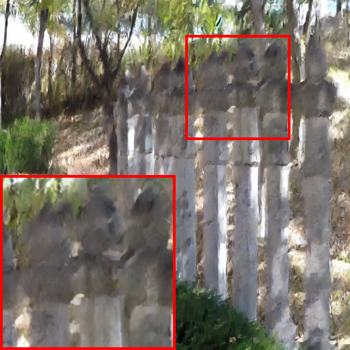} & 
    \includegraphics[width=.18\textwidth]{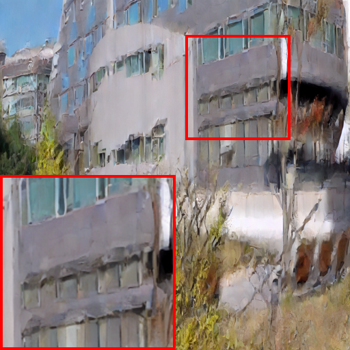}\\
   \multicolumn{5}{c}{\vspace{-6pt}} \\
    \raisebox{1.1\height}{\rotatebox[origin=c]{90}{ Ours w/o $T_{3}$}} &
    \includegraphics[width=.18\textwidth]{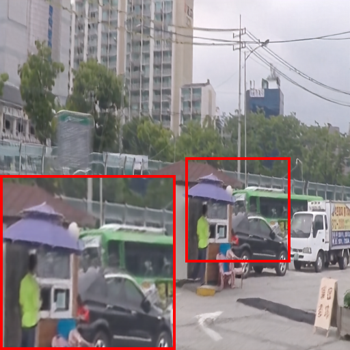} & 
    \includegraphics[width=.18\textwidth]{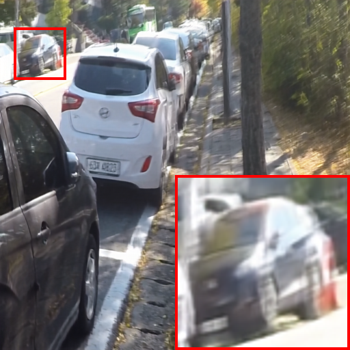} & 
    \includegraphics[width=.18\textwidth]{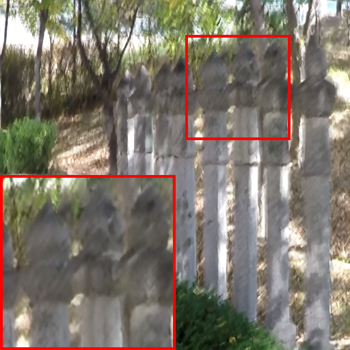} & 
    \includegraphics[width=.18\textwidth]{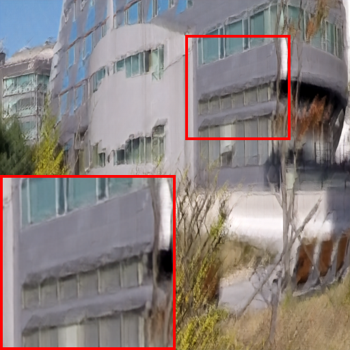} \\
   \multicolumn{5}{c}{\vspace{-6pt}} \\
    \raisebox{1.7\height}{\rotatebox[origin=c]{90}{ \textbf{Ours}}} &
    \includegraphics[width=.18\textwidth]{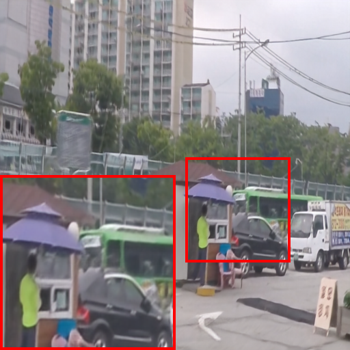} & 
    \includegraphics[width=.18\textwidth]{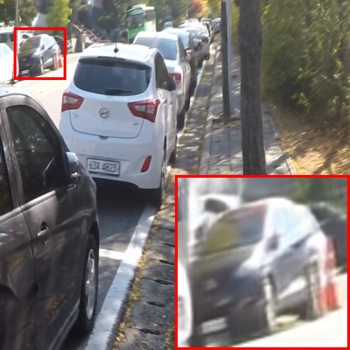} & 
    \includegraphics[width=.18\textwidth]{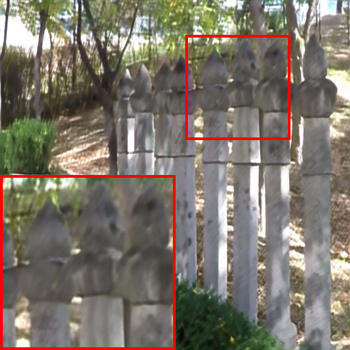} & 
    \includegraphics[width=.18\textwidth]{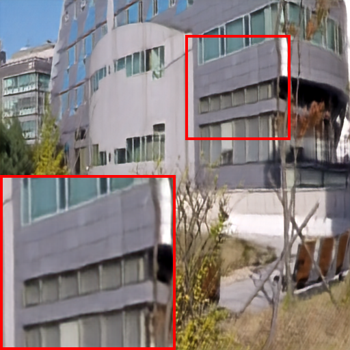} \\
   \multicolumn{5}{c}{\vspace{-6pt}} \\
    \raisebox{2.2\height}{\rotatebox[origin=c]{90}{ Blur}} &
    \includegraphics[width=.18\textwidth]{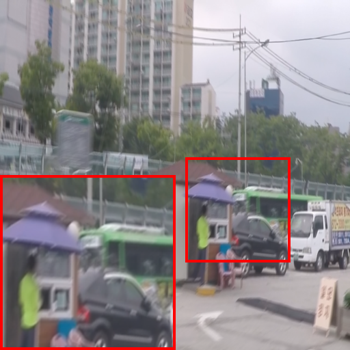} & 
    \includegraphics[width=.18\textwidth]{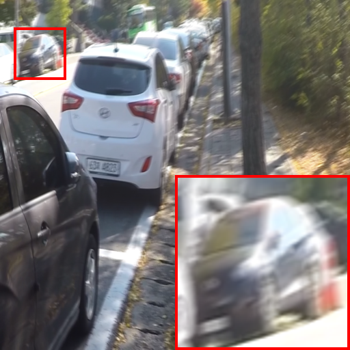} & 
    \includegraphics[width=.18\textwidth]{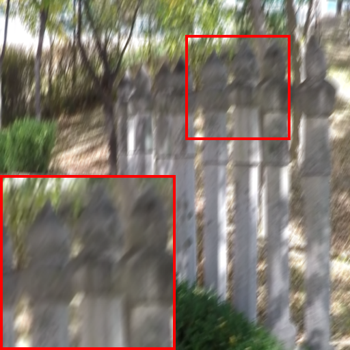} & 
    \includegraphics[width=.18\textwidth]{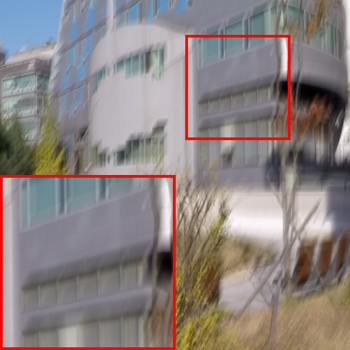} \\
   \multicolumn{5}{c}{\vspace{-6pt}} \\
  \end{tabular}
  \caption{Additional visual comparison on the \textcolor{red}{GoPro} dataset (Continue).}
  \label{fig:appendix resultsG2}
\end{figure*}

\begin{figure*}[!t]
  \centering
  \setlength{\tabcolsep}{5pt} 
  \begin{tabular}{ccccc}
    &   Sample 1 &   Sample 2 &   Sample 3 &   Sample 4 \\
    \raisebox{0.9\height}{\rotatebox[origin=c]{90}{ Ground Truth}} &
    \includegraphics[width=.18\textwidth]{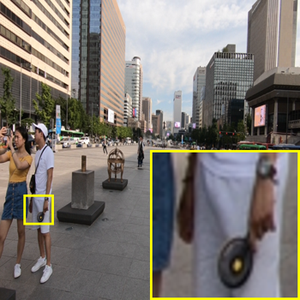} & 
    \includegraphics[width=.18\textwidth]{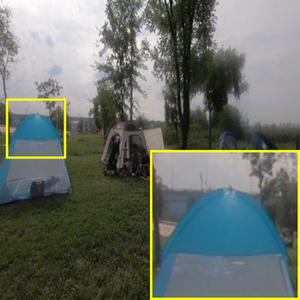} & 
    \includegraphics[width=.18\textwidth]{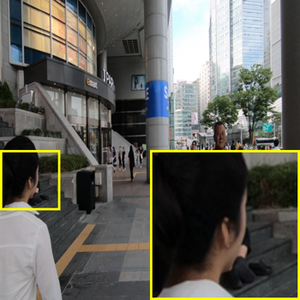} & 
    \includegraphics[width=.18\textwidth]{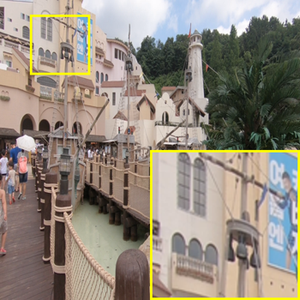}\\
   \multicolumn{5}{c}{\vspace{-6pt}} \\
    \raisebox{1.55\height}{\rotatebox[origin=c]{90}{ AirNet}} &
    \includegraphics[width=.18\textwidth]{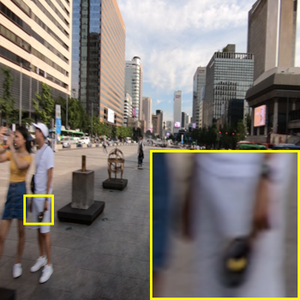} & 
    \includegraphics[width=.18\textwidth]{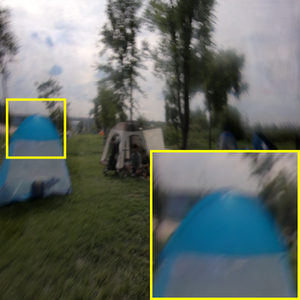} & 
    \includegraphics[width=.18\textwidth]{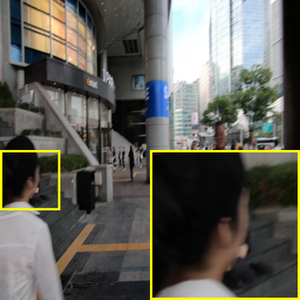} & 
    \includegraphics[width=.18\textwidth]{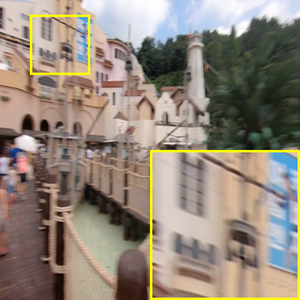}\\
   \multicolumn{5}{c}{\vspace{-6pt}} \\
    \raisebox{1.3\height}{\rotatebox[origin=c]{90}{ PromptIR}} &
    \includegraphics[width=.18\textwidth]{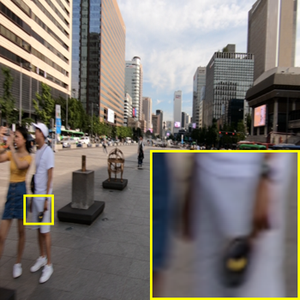} & 
    \includegraphics[width=.18\textwidth]{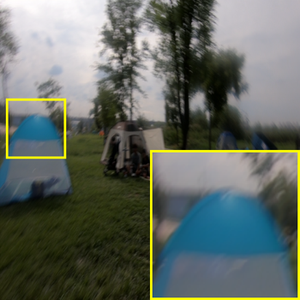} & 
    \includegraphics[width=.18\textwidth]{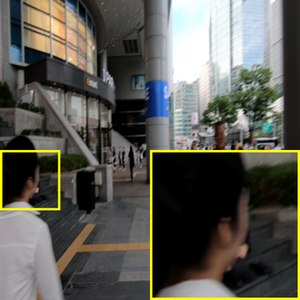} & 
    \includegraphics[width=.18\textwidth]{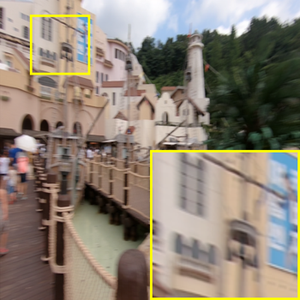}\\
   \multicolumn{5}{c}{\vspace{-6pt}} \\
    \raisebox{1.1\height}{\rotatebox[origin=c]{90}{ FFTformer}} &
    \includegraphics[width=.18\textwidth]{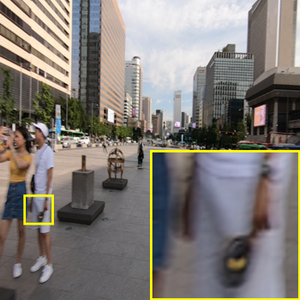} & 
    \includegraphics[width=.18\textwidth]{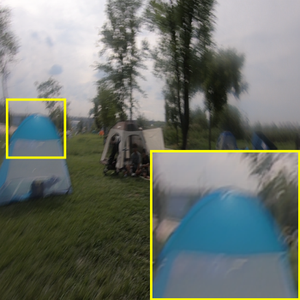} & 
    \includegraphics[width=.18\textwidth]{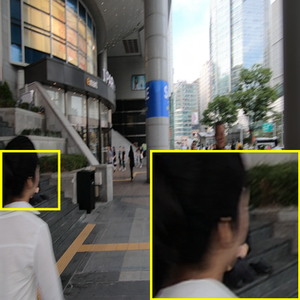} & 
    \includegraphics[width=.18\textwidth]{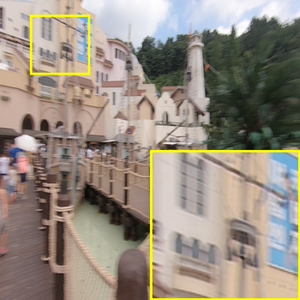}\\
   \multicolumn{5}{c}{\vspace{-6pt}} \\
    \raisebox{1.3\height}{\rotatebox[origin=c]{90}{ AdaRevD}} &
    \includegraphics[width=.18\textwidth]{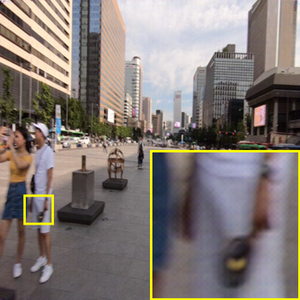} & 
    \includegraphics[width=.18\textwidth]{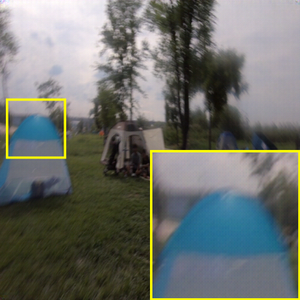} & 
    \includegraphics[width=.18\textwidth]{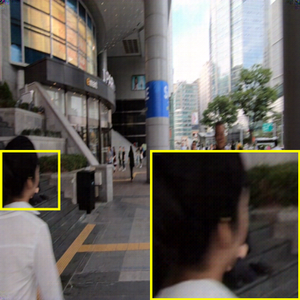} & 
    \includegraphics[width=.18\textwidth]{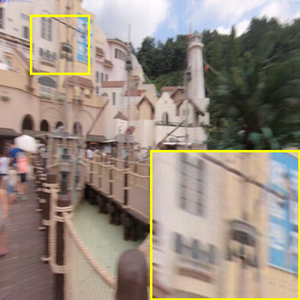}\\
   \multicolumn{5}{c}{\vspace{-6pt}} \\
    \raisebox{1.5\height}{\rotatebox[origin=c]{90}{ HI-diff}} &
    \includegraphics[width=.18\textwidth]{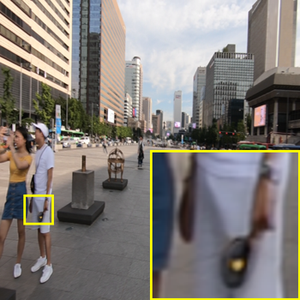} & 
    \includegraphics[width=.18\textwidth]{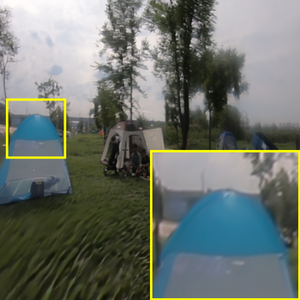} & 
    \includegraphics[width=.18\textwidth]{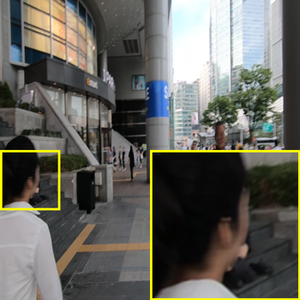} & 
    \includegraphics[width=.18\textwidth]{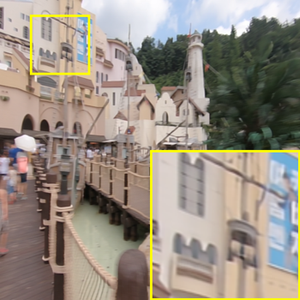}\\
   \multicolumn{5}{c}{\vspace{-6pt}} \\
    \raisebox{1.1\height}{\rotatebox[origin=c]{90}{ Blur2Blur}} &
    \includegraphics[width=.18\textwidth]{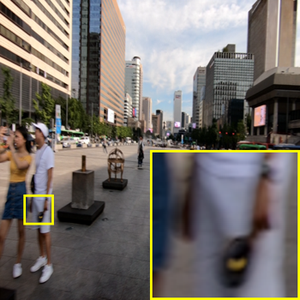} & 
    \includegraphics[width=.18\textwidth]{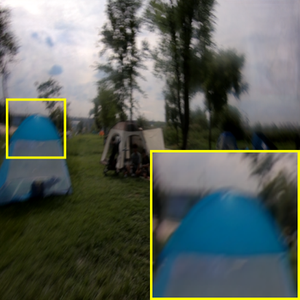} & 
    \includegraphics[width=.18\textwidth]{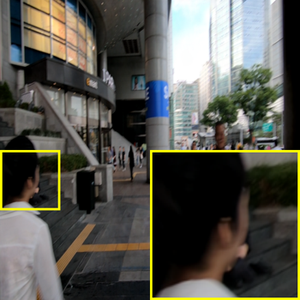} & 
    \includegraphics[width=.18\textwidth]{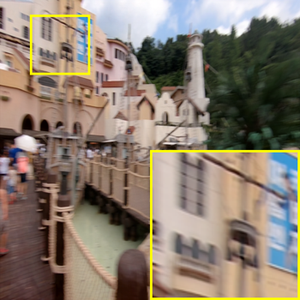} \\
 \multicolumn{5}{c}{\vspace{-6pt}} \\
  \end{tabular}
  \caption{Additional visual comparison on the \textcolor{yellow}{REDS} dataset.}
  \label{fig:appendix resultsREDS}
\end{figure*}

\begin{figure*}[!t]
  \centering
  \setlength{\tabcolsep}{5pt} 
  \begin{tabular}{ccccc}
    &   Sample 1 &   Sample 2 &   Sample 3 &   Sample 4 \\
    \raisebox{1.4\height}{\rotatebox[origin=c]{90}{ DiffBir}} &
    \includegraphics[width=.18\textwidth]{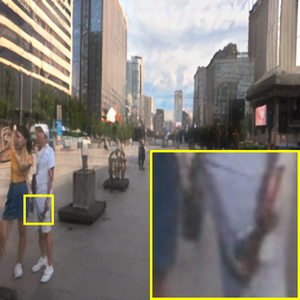} & 
    \includegraphics[width=.18\textwidth]{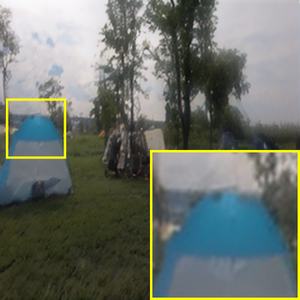} & 
    \includegraphics[width=.18\textwidth]{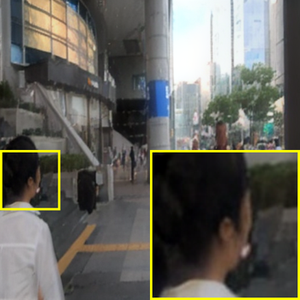} & 
    \includegraphics[width=.18\textwidth]{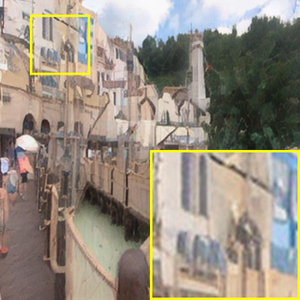} \\
   \multicolumn{5}{c}{\vspace{-6pt}} \\
    \raisebox{1.1\height}{\rotatebox[origin=c]{90}{ Diff-Plugin}} &
    \includegraphics[width=.18\textwidth]{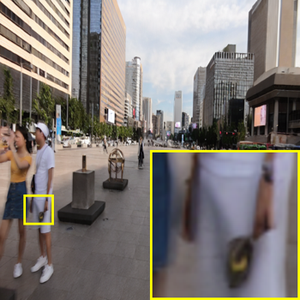} & 
    \includegraphics[width=.18\textwidth]{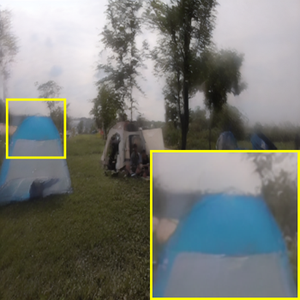} & 
    \includegraphics[width=.18\textwidth]{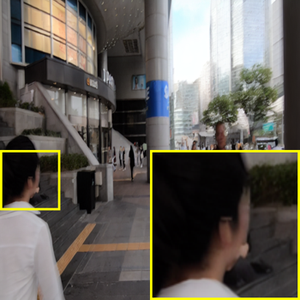} & 
    \includegraphics[width=.18\textwidth]{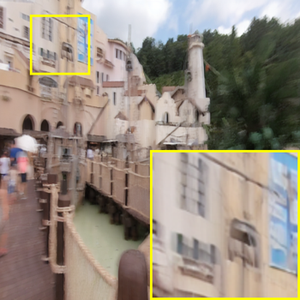}\\
   \multicolumn{5}{c}{\vspace{-6pt}} \\
    \raisebox{1.2\height}{\rotatebox[origin=c]{90}{ DA-CLIP}} &
    \includegraphics[width=.18\textwidth]{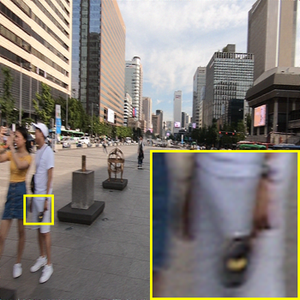} & 
    \includegraphics[width=.18\textwidth]{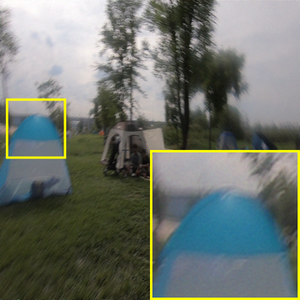} & 
    \includegraphics[width=.18\textwidth]{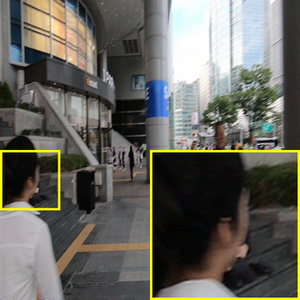} & 
    \includegraphics[width=.18\textwidth]{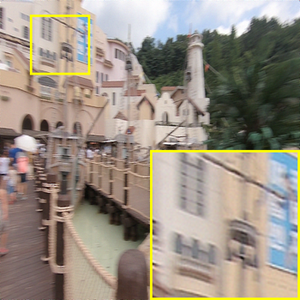}\\
   \multicolumn{5}{c}{\vspace{-6pt}} \\
    \raisebox{1\height}{\rotatebox[origin=c]{90}{ Ours w/o $Q_s$}} &
    \includegraphics[width=.18\textwidth]{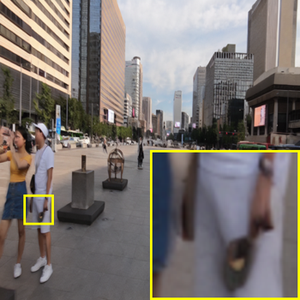} & 
    \includegraphics[width=.18\textwidth]{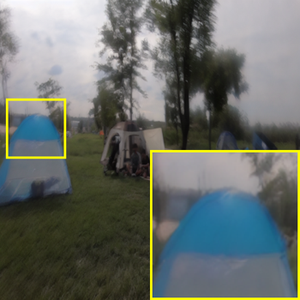} & 
    \includegraphics[width=.18\textwidth]{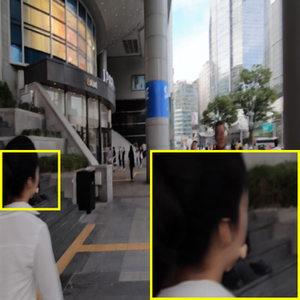} & 
    \includegraphics[width=.18\textwidth]{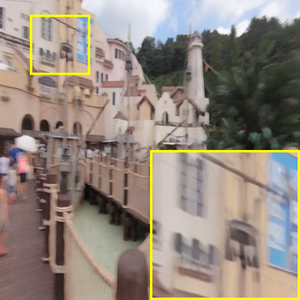} \\
   \multicolumn{5}{c}{\vspace{-6pt}} \\
    \raisebox{0.9\height}{\rotatebox[origin=c]{90}{ Ours w/o $T_{2}T_{3}$}} &
    \includegraphics[width=.18\textwidth]{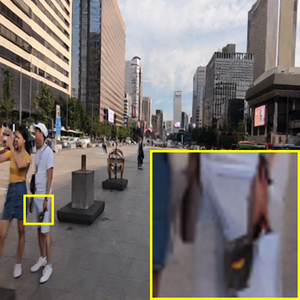} & 
    \includegraphics[width=.18\textwidth]{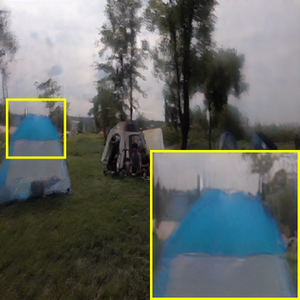} & 
    \includegraphics[width=.18\textwidth]{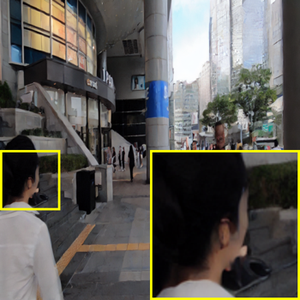} & 
    \includegraphics[width=.18\textwidth]{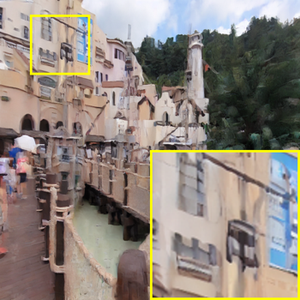}\\
   \multicolumn{5}{c}{\vspace{-6pt}} \\
    \raisebox{1.1\height}{\rotatebox[origin=c]{90}{ Ours w/o $T_{3}$}} &
    \includegraphics[width=.18\textwidth]{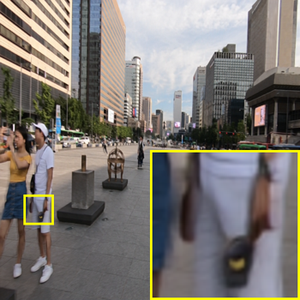} & 
    \includegraphics[width=.18\textwidth]{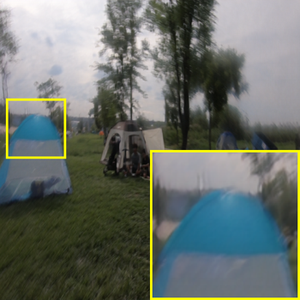} & 
    \includegraphics[width=.18\textwidth]{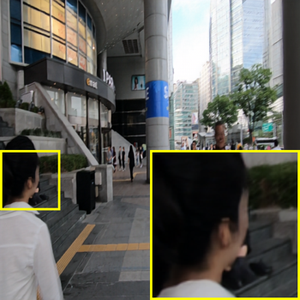} & 
    \includegraphics[width=.18\textwidth]{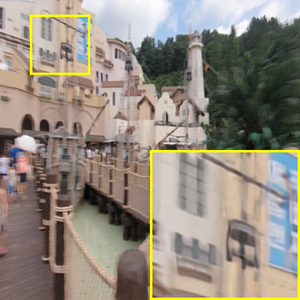} \\
   \multicolumn{5}{c}{\vspace{-6pt}} \\
    \raisebox{1.7\height}{\rotatebox[origin=c]{90}{ \textbf{Ours}}} &
    \includegraphics[width=.18\textwidth]{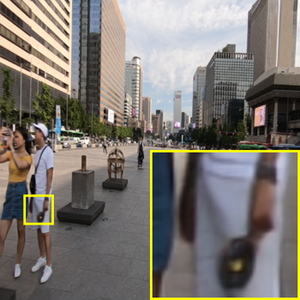} & 
    \includegraphics[width=.18\textwidth]{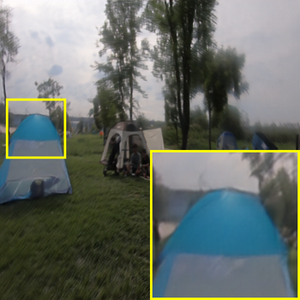} & 
    \includegraphics[width=.18\textwidth]{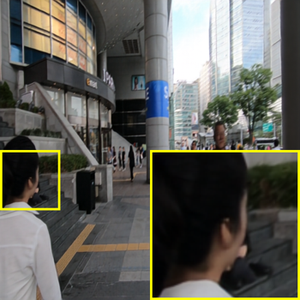} & 
    \includegraphics[width=.18\textwidth]{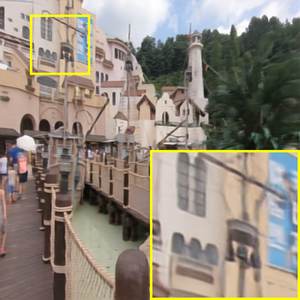} \\
   \multicolumn{5}{c}{\vspace{-6pt}} \\
    \raisebox{2.2\height}{\rotatebox[origin=c]{90}{ Blur}} &
    \includegraphics[width=.18\textwidth]{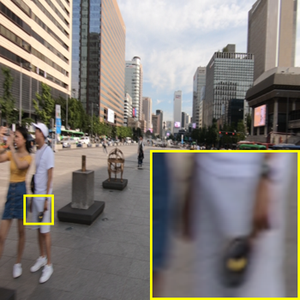} & 
    \includegraphics[width=.18\textwidth]{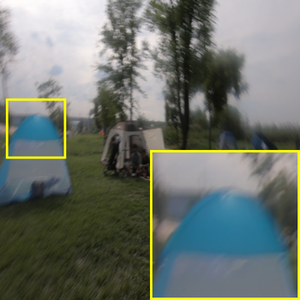} & 
    \includegraphics[width=.18\textwidth]{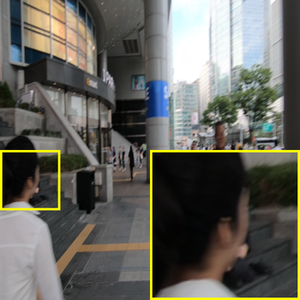} & 
    \includegraphics[width=.18\textwidth]{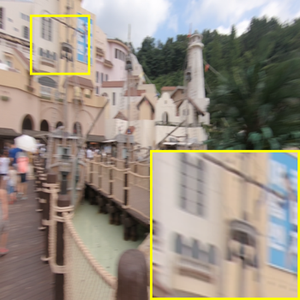} \\
   \multicolumn{5}{c}{\vspace{-6pt}} \\
  \end{tabular}
  \caption{Additional visual comparison on the \textcolor{yellow}{REDS} dataset (Continue).}
  \label{fig:appendix resultsREDS2}
\end{figure*}

\begin{figure*}[!t]
  \centering
  \setlength{\tabcolsep}{5pt} 
  \begin{tabular}{ccccc}
    &   Sample 1 &   Sample 2 &   Sample 3 &   Sample 4 \\
    \raisebox{0.9\height}{\rotatebox[origin=c]{90}{ Ground Truth}} &
    \includegraphics[width=.18\textwidth]{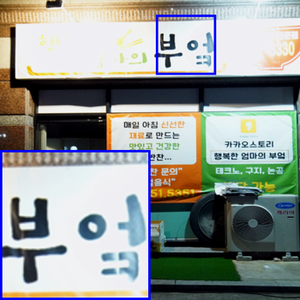} & 
    \includegraphics[width=.18\textwidth]{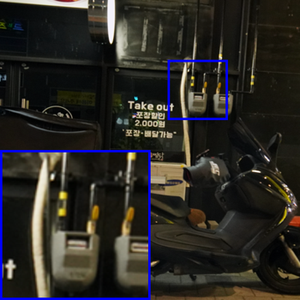} & 
    \includegraphics[width=.18\textwidth]{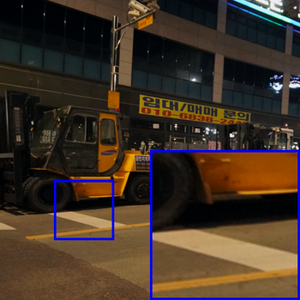} & 
    \includegraphics[width=.18\textwidth]{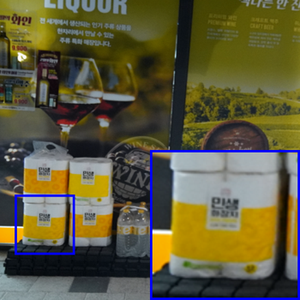}\\
   \multicolumn{5}{c}{\vspace{-6pt}} \\
    \raisebox{1.55\height}{\rotatebox[origin=c]{90}{ AirNet}} &
    \includegraphics[width=.18\textwidth]{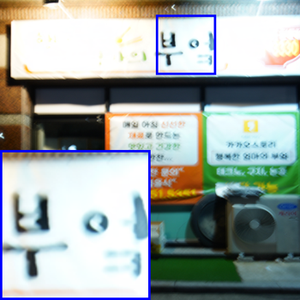} & 
    \includegraphics[width=.18\textwidth]{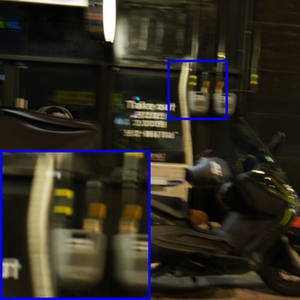} & 
    \includegraphics[width=.18\textwidth]{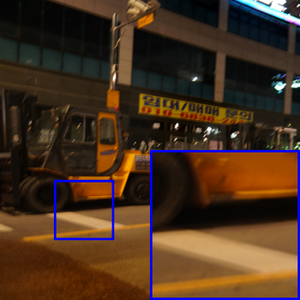} & 
    \includegraphics[width=.18\textwidth]{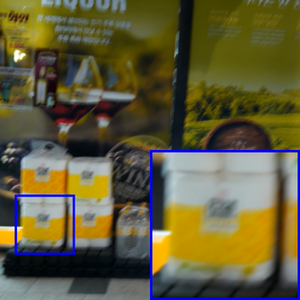}\\
   \multicolumn{5}{c}{\vspace{-6pt}} \\
    \raisebox{1.3\height}{\rotatebox[origin=c]{90}{ PromptIR}} &
    \includegraphics[width=.18\textwidth]{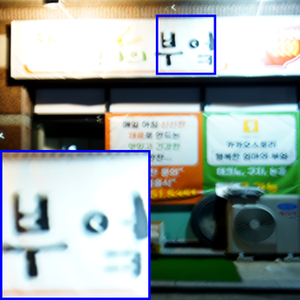} & 
    \includegraphics[width=.18\textwidth]{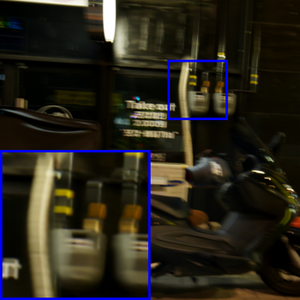} & 
    \includegraphics[width=.18\textwidth]{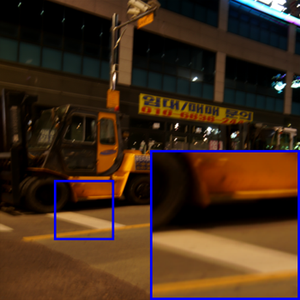} & 
    \includegraphics[width=.18\textwidth]{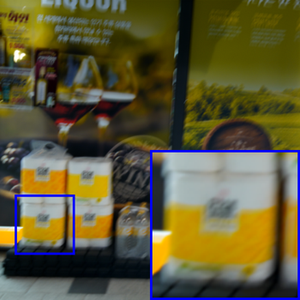}\\
   \multicolumn{5}{c}{\vspace{-6pt}} \\
    \raisebox{1.1\height}{\rotatebox[origin=c]{90}{ FFTformer}} &
    \includegraphics[width=.18\textwidth]{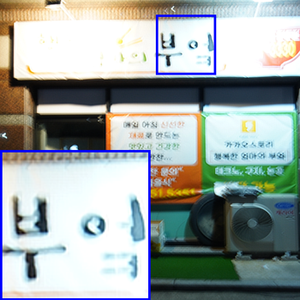} & 
    \includegraphics[width=.18\textwidth]{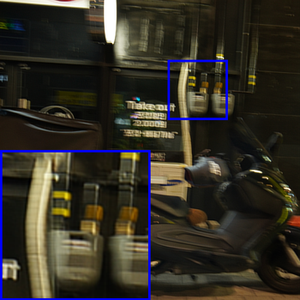} & 
    \includegraphics[width=.18\textwidth]{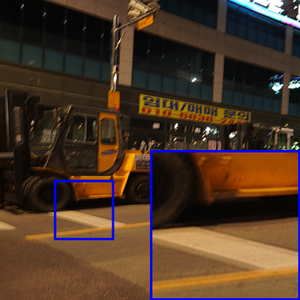} & 
    \includegraphics[width=.18\textwidth]{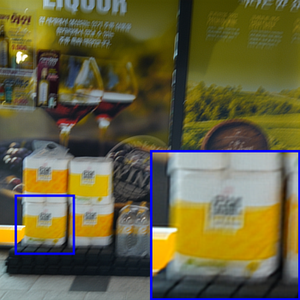}\\
   \multicolumn{5}{c}{\vspace{-6pt}} \\
    \raisebox{1.3\height}{\rotatebox[origin=c]{90}{ AdaRevD}} &
    \includegraphics[width=.18\textwidth]{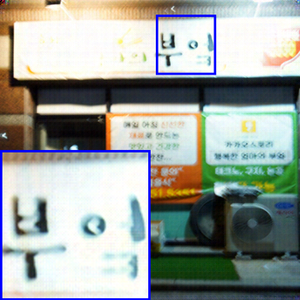} & 
    \includegraphics[width=.18\textwidth]{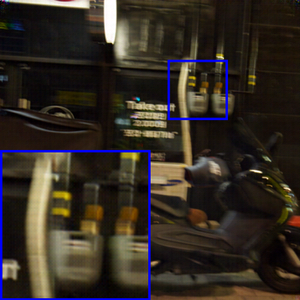} & 
    \includegraphics[width=.18\textwidth]{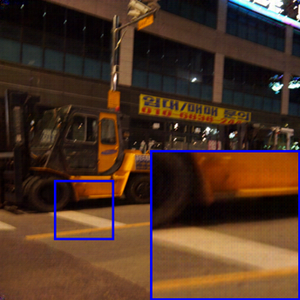} & 
    \includegraphics[width=.18\textwidth]{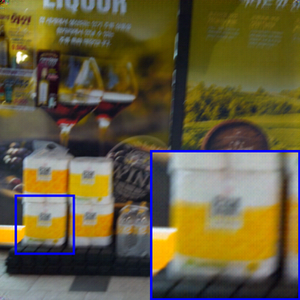}\\
   \multicolumn{5}{c}{\vspace{-6pt}} \\
    \raisebox{1.5\height}{\rotatebox[origin=c]{90}{ HI-diff}} &
    \includegraphics[width=.18\textwidth]{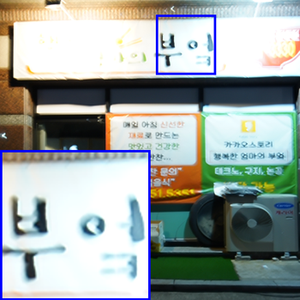} & 
    \includegraphics[width=.18\textwidth]{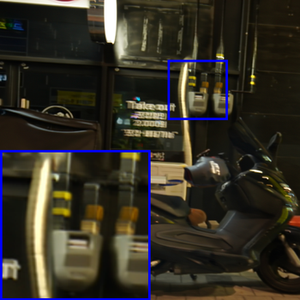} & 
    \includegraphics[width=.18\textwidth]{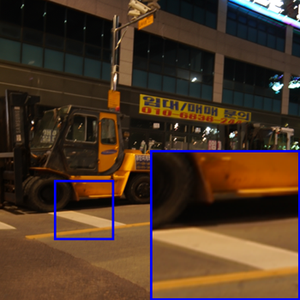} & 
    \includegraphics[width=.18\textwidth]{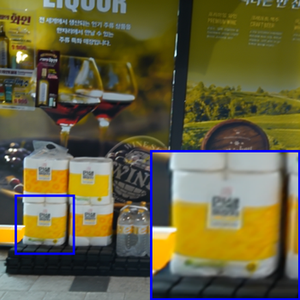}\\
   \multicolumn{5}{c}{\vspace{-6pt}} \\
    \raisebox{1.1\height}{\rotatebox[origin=c]{90}{ Blur2Blur}} &
    \includegraphics[width=.18\textwidth]{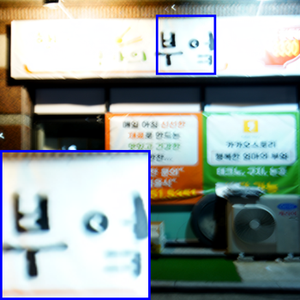} & 
    \includegraphics[width=.18\textwidth]{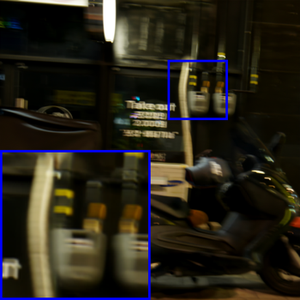} & 
    \includegraphics[width=.18\textwidth]{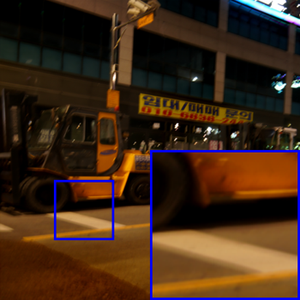} & 
    \includegraphics[width=.18\textwidth]{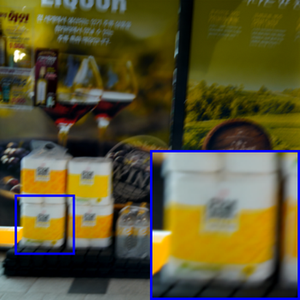} \\
 \multicolumn{5}{c}{\vspace{-6pt}} \\
  \end{tabular}
  \caption{Additional visual comparison on the \textcolor{blue}{RealBlur-J} dataset.}
  \label{fig:appendix resultsJ}
\end{figure*}

\begin{figure*}[!t]
  \centering
  \setlength{\tabcolsep}{5pt} 
  \begin{tabular}{ccccc}
    &   Sample 1 &   Sample 2 &   Sample 3 &   Sample 4 \\
    \raisebox{1.4\height}{\rotatebox[origin=c]{90}{ DiffBir}} &
    \includegraphics[width=.18\textwidth]{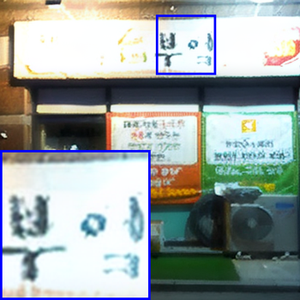} & 
    \includegraphics[width=.18\textwidth]{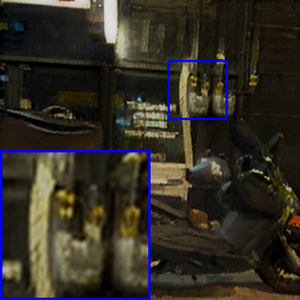} & 
    \includegraphics[width=.18\textwidth]{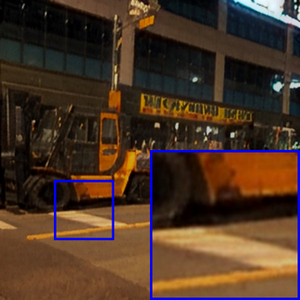} & 
    \includegraphics[width=.18\textwidth]{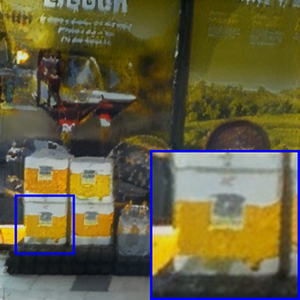} \\
   \multicolumn{5}{c}{\vspace{-6pt}} \\
    \raisebox{1.1\height}{\rotatebox[origin=c]{90}{ Diff-Plugin}} &
    \includegraphics[width=.18\textwidth]{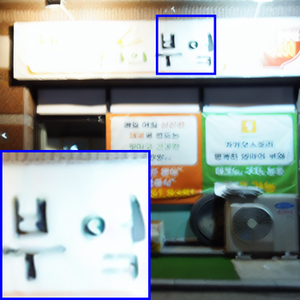} & 
    \includegraphics[width=.18\textwidth]{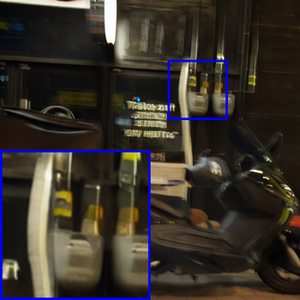} & 
    \includegraphics[width=.18\textwidth]{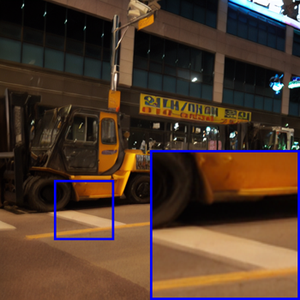} & 
    \includegraphics[width=.18\textwidth]{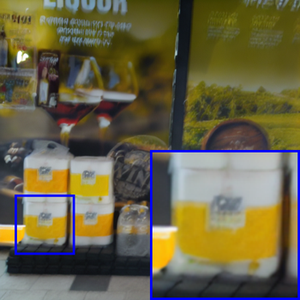}\\
   \multicolumn{5}{c}{\vspace{-6pt}} \\
    \raisebox{1.2\height}{\rotatebox[origin=c]{90}{ DA-CLIP}} &
    \includegraphics[width=.18\textwidth]{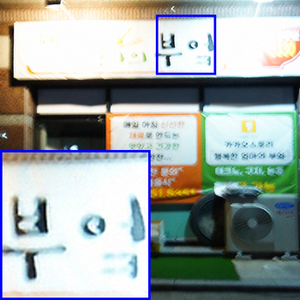} & 
    \includegraphics[width=.18\textwidth]{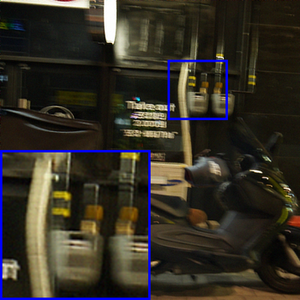} & 
    \includegraphics[width=.18\textwidth]{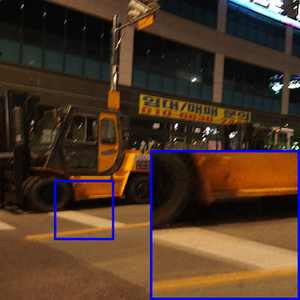} & 
    \includegraphics[width=.18\textwidth]{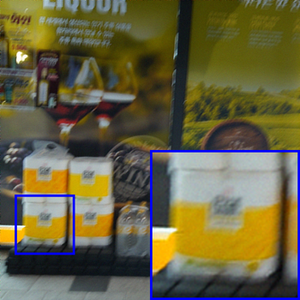}\\
   \multicolumn{5}{c}{\vspace{-6pt}} \\
    \raisebox{1\height}{\rotatebox[origin=c]{90}{ Ours w/o $Q_s$}} &
    \includegraphics[width=.18\textwidth]{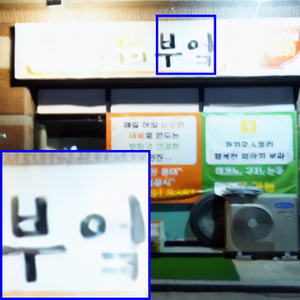} & 
    \includegraphics[width=.18\textwidth]{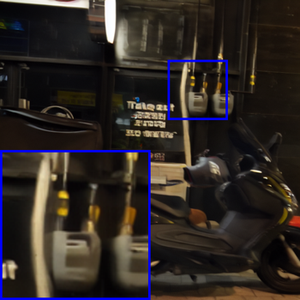} & 
    \includegraphics[width=.18\textwidth]{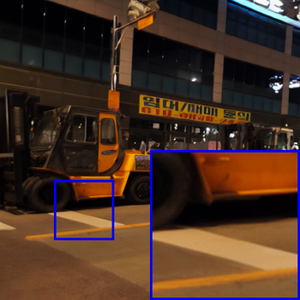} & 
    \includegraphics[width=.18\textwidth]{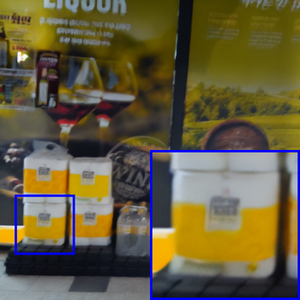} \\
   \multicolumn{5}{c}{\vspace{-6pt}} \\
    \raisebox{0.9\height}{\rotatebox[origin=c]{90}{ Ours w/o $T_{2}T_{3}$}} &
    \includegraphics[width=.18\textwidth]{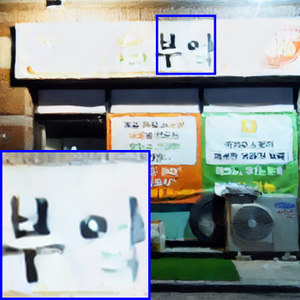} & 
    \includegraphics[width=.18\textwidth]{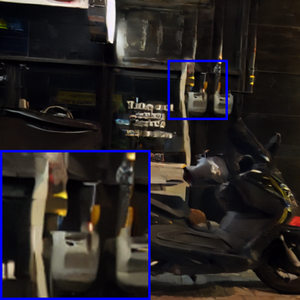} & 
    \includegraphics[width=.18\textwidth]{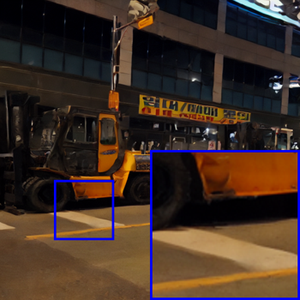} & 
    \includegraphics[width=.18\textwidth]{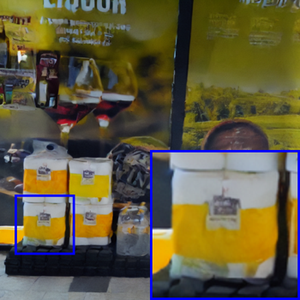}\\
   \multicolumn{5}{c}{\vspace{-6pt}} \\
    \raisebox{1.1\height}{\rotatebox[origin=c]{90}{ Ours w/o $T_{3}$}} &
    \includegraphics[width=.18\textwidth]{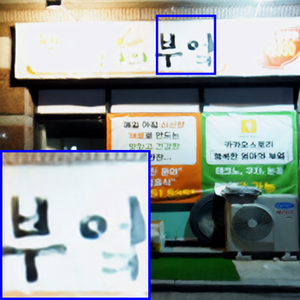} & 
    \includegraphics[width=.18\textwidth]{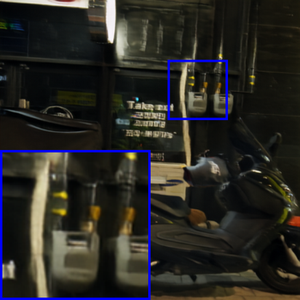} & 
    \includegraphics[width=.18\textwidth]{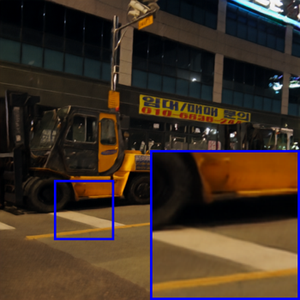} & 
    \includegraphics[width=.18\textwidth]{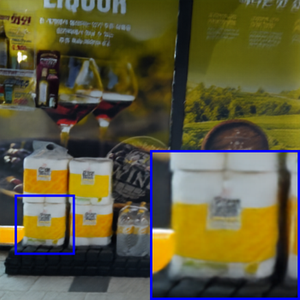} \\
   \multicolumn{5}{c}{\vspace{-6pt}} \\
    \raisebox{1.7\height}{\rotatebox[origin=c]{90}{ \textbf{Ours}}} &
    \includegraphics[width=.18\textwidth]{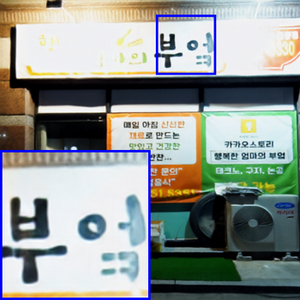} & 
    \includegraphics[width=.18\textwidth]{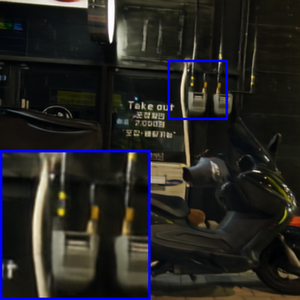} & 
    \includegraphics[width=.18\textwidth]{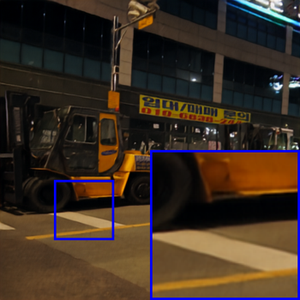} & 
    \includegraphics[width=.18\textwidth]{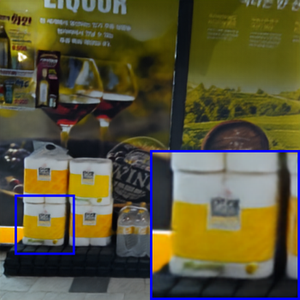} \\
   \multicolumn{5}{c}{\vspace{-6pt}} \\
    \raisebox{2.2\height}{\rotatebox[origin=c]{90}{ Blur}} &
    \includegraphics[width=.18\textwidth]{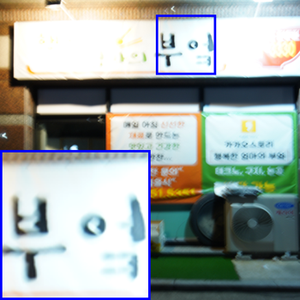} & 
    \includegraphics[width=.18\textwidth]{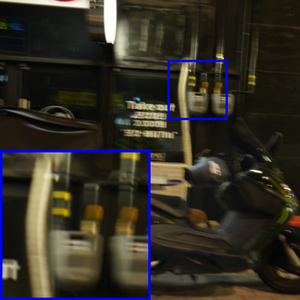} & 
    \includegraphics[width=.18\textwidth]{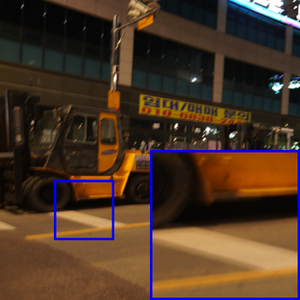} & 
    \includegraphics[width=.18\textwidth]{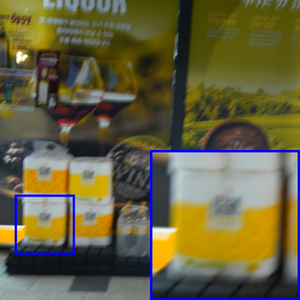} \\
   \multicolumn{5}{c}{\vspace{-6pt}} \\
  \end{tabular}
  \caption{Additional visual comparison on the \textcolor{blue}{RealBlur-J} dataset (Continue).}
  \label{fig:appendix resultsJ2}
\end{figure*}

\begin{figure*}[!t]
  \centering
  \setlength{\tabcolsep}{5pt} 
  \begin{tabular}{ccccc}
    &   Sample 1 &   Sample 2 &   Sample 3 &   Sample 4 \\
    \raisebox{0.9\height}{\rotatebox[origin=c]{90}{ Ground Truth}} &
    \includegraphics[width=.18\textwidth]{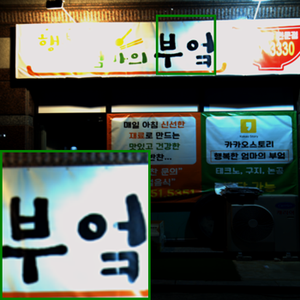} & 
    \includegraphics[width=.18\textwidth]{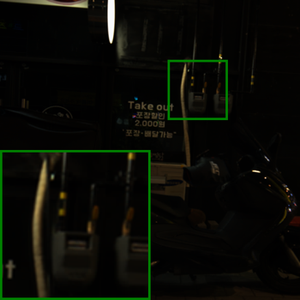} & 
    \includegraphics[width=.18\textwidth]{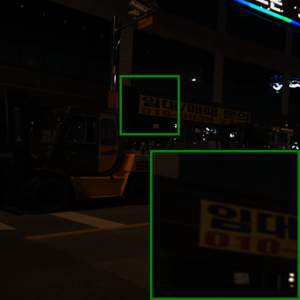} & 
    \includegraphics[width=.18\textwidth]{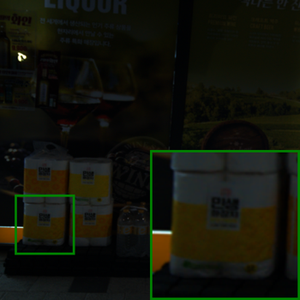}\\
   \multicolumn{5}{c}{\vspace{-6pt}} \\
    \raisebox{1.55\height}{\rotatebox[origin=c]{90}{ AirNet}} &
    \includegraphics[width=.18\textwidth]{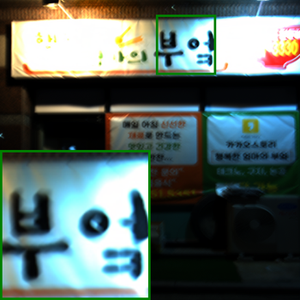} & 
    \includegraphics[width=.18\textwidth]{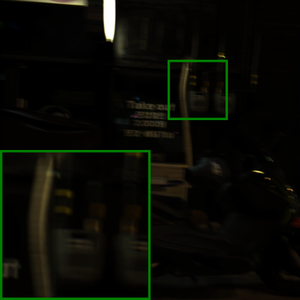} & 
    \includegraphics[width=.18\textwidth]{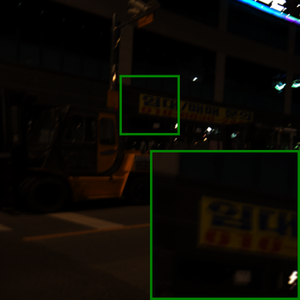} & 
    \includegraphics[width=.18\textwidth]{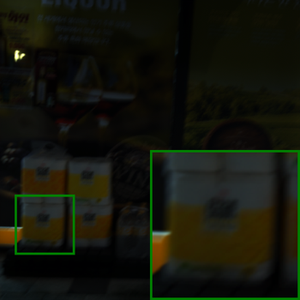}\\
   \multicolumn{5}{c}{\vspace{-6pt}} \\
    \raisebox{1.3\height}{\rotatebox[origin=c]{90}{ PromptIR}} &
    \includegraphics[width=.18\textwidth]{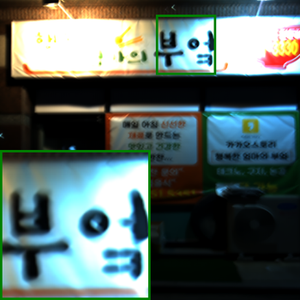} & 
    \includegraphics[width=.18\textwidth]{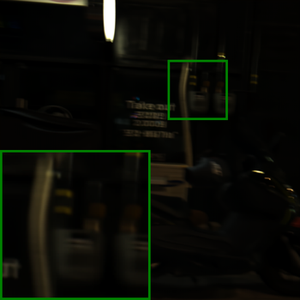} & 
    \includegraphics[width=.18\textwidth]{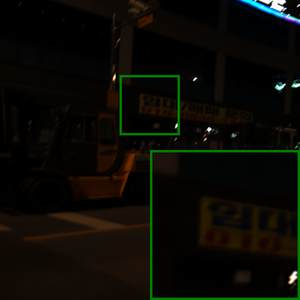} & 
    \includegraphics[width=.18\textwidth]{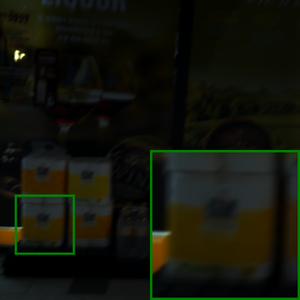}\\
   \multicolumn{5}{c}{\vspace{-6pt}} \\
    \raisebox{1.1\height}{\rotatebox[origin=c]{90}{ FFTformer}} &
    \includegraphics[width=.18\textwidth]{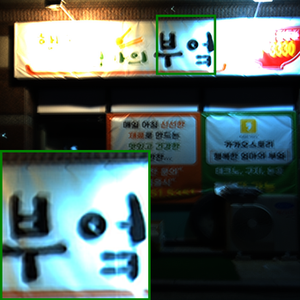} & 
    \includegraphics[width=.18\textwidth]{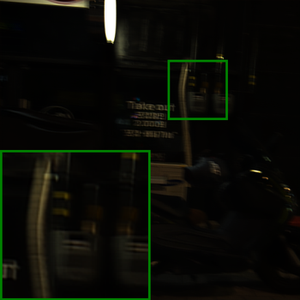} & 
    \includegraphics[width=.18\textwidth]{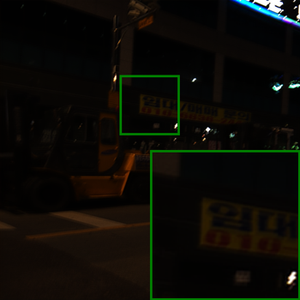} & 
    \includegraphics[width=.18\textwidth]{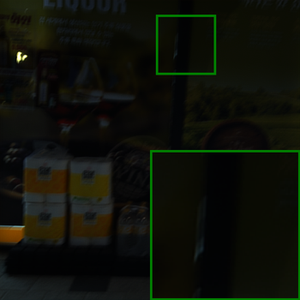}\\
   \multicolumn{5}{c}{\vspace{-6pt}} \\
    \raisebox{1.3\height}{\rotatebox[origin=c]{90}{ AdaRevD}} &
    \includegraphics[width=.18\textwidth]{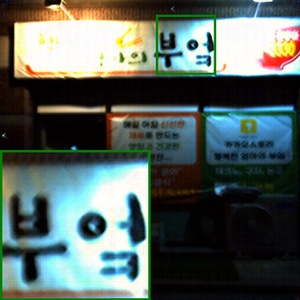} & 
    \includegraphics[width=.18\textwidth]{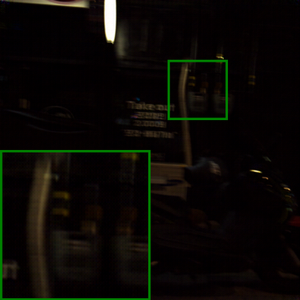} & 
    \includegraphics[width=.18\textwidth]{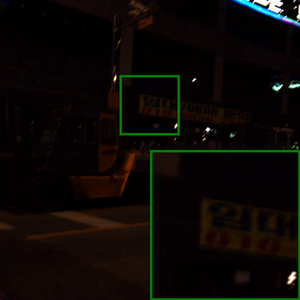} & 
    \includegraphics[width=.18\textwidth]{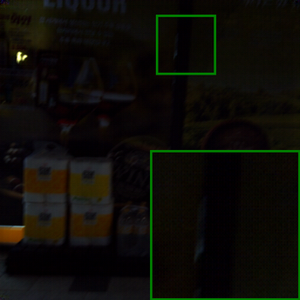}\\
   \multicolumn{5}{c}{\vspace{-6pt}} \\
    \raisebox{1.5\height}{\rotatebox[origin=c]{90}{ HI-diff}} &
    \includegraphics[width=.18\textwidth]{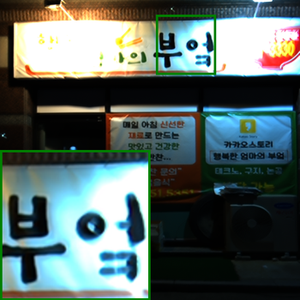} & 
    \includegraphics[width=.18\textwidth]{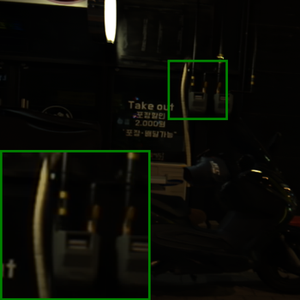} & 
    \includegraphics[width=.18\textwidth]{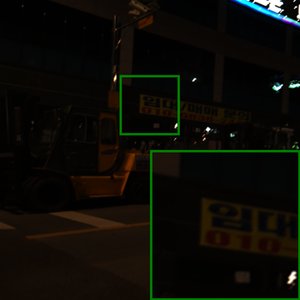} & 
    \includegraphics[width=.18\textwidth]{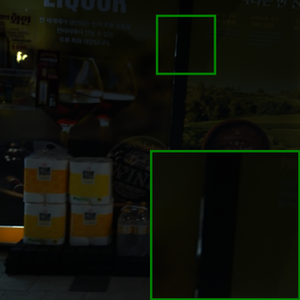}\\
   \multicolumn{5}{c}{\vspace{-6pt}} \\
    \raisebox{1.1\height}{\rotatebox[origin=c]{90}{ Blur2Blur}} &
    \includegraphics[width=.18\textwidth]{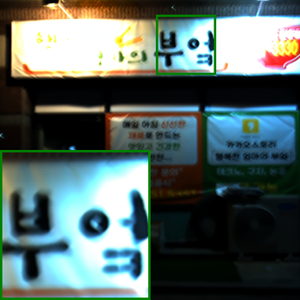} & 
    \includegraphics[width=.18\textwidth]{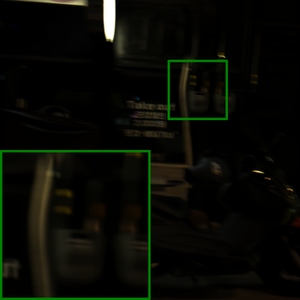} & 
    \includegraphics[width=.18\textwidth]{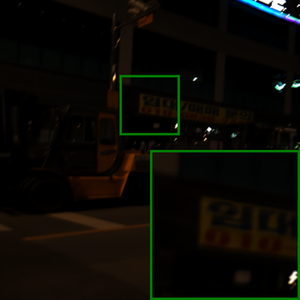} & 
    \includegraphics[width=.18\textwidth]{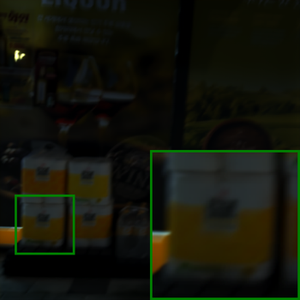} \\
 \multicolumn{5}{c}{\vspace{-6pt}} \\
  \end{tabular}
  \caption{Additional visual comparison on the \textcolor{green}{RealBlur-R} dataset.}
  \label{fig:appendix resultsR1}
\end{figure*}

\clearpage

\bibliographystyle{unsrt}
\bibliography{main}

\end{document}